\documentclass[times, review, 10pt]{elsarticle}

\usepackage[numbers]{natbib}
\usepackage{float}
\usepackage{graphicx}
\usepackage{subfigure}
\usepackage{xcolor}
\usepackage{soul} % For highlighting

\usepackage{comment}
\usepackage{makecell}

\usepackage{amsmath,amssymb,amsfonts}

\usepackage{hyperref}

\hypersetup{
    colorlinks=true,
    linkcolor=blue,
    filecolor=blue,      
    urlcolor=blue,
    citecolor=blue,
}

\def\,{$\mskip\thinmuskip$} \def\!{$\mskip-\thinmuskip$}

\usepackage{algorithm}
\usepackage[algo2e]{algorithm2e} 
\usepackage{algorithmicx}
\newcommand{\var}{\texttt}

\def\BibTeX{{\rm B\kern-.05em{\sc i\kern-.025em b}\kern-.08em
    T\kern-.1667em\lower.7ex\hbox{E}\kern-.125emX}}

\begin{document}

\title{Convex space learning for tabular synthetic data generation}
%\shorttitle{NextConvGeN}

% TODO: Update this
%\shortauthors{M. Mahendra et al.}
\author[1]{Manjunath Mahendra}
\author[1]{Chaithra Umesh}
\author[1]{Kristian Schultz}
\author[1,2,3]{Olaf Wolkenhauer}
\author[1,4]{Saptarshi Bej$^{*}$}
\affiliation[1]{Institute of Computer Science, University of Rostock; Germany}
\affiliation[2]{Leibniz-Institute for Food Systems Biology; Technical University of Munich, Freising; Germany}
\affiliation[3]{Stellenbosch Institute for Advanced Study, South Africa}
\affiliation[4]{School of Data Science, Indian Institute of Science Education and Research Thiruvananthapuram, India}
\cortext[ca1]{Corresponding author\\ \textit{E-mail addresses:} \url{manjunath.mahendra@uni-rostock.de} (M. Mahendra), \url{chaithra.umesh@uni-rostock.de} (C. Umesh), \url{kristian.schultz@uni-rostock.de} (K. Schultz),  \url{olaf.wolkenhauer@uni-rostock.de} (O. Wolkenhauer), \url{sbej7042@iisertvm.ac.in} (S. Bej)}
\begin{abstract}
While synthetic data for images and text has seen significant advancements, the structured nature of tabular data presents unique challenges in generating high-utility synthetic datasets while preserving privacy. A detailed study on how synthetic data generation models strike a balance between utility and privacy measures is missing. Moreover, the approach of Convex Space Learning (CSL) for synthetic data generation, which is popular in imbalanced classification, has not been explored for synthetic tabular data generation in general.

To address these two research gaps, we propose NextConvGeN, an extension of the ConvGeN framework that generalizes CSL for entire tabular data generation. NextConvGeN employs a generator-discriminator architecture that uses deep cooperative learning, refining synthetic data generation within local convex data neighborhoods. We then compare several state-of-the-art synthetic tabular data generation models to assess their performances qualitatively and quantitatively in the context of privacy-utility balance.

%Our results show that NextConvGeN achieves a balance between utility and privacy measures, making it a promising tool for privacy-preserving data sharing and analysis. This work contributes to advancing synthetic tabular data generation by expanding CSL beyond imbalanced classification, improving the theoretical grounding of synthetic data sampling, and providing a structured evaluation of privacy-utility trade-offs for tabular datasets, especially in the biomedical domain.
Our results show that NextConvGeN prioritizes utility preservation while incorporating privacy measures, making it a promising tool for generating high-utility synthetic data for analysis. This work advances synthetic tabular data generation by expanding convex space learning beyond imbalanced classification, strengthening the theoretical foundation of synthetic data sampling, and providing a structured evaluation of utility-driven tabular data generation, especially in the biomedical domain.
\end{abstract}

\maketitle
\section{Introduction}\label{sec:introduction}
In 2021, Bej \textit{et al.} introduced an approach called \textit{Convex Space Learning} (CSL) to address challenges in imbalanced classification \cite{bej_multi-schematic_2021}. The approach generates synthetic samples by interpolating data points within specific regions of the minority class's convex space, determined by the dataset's distribution \cite{bej_loras_2021} and using the synthetic samples to improve classifier performance on imbalanced datasets. The convex space, defined as the convex hull of the data, encompasses points sharing similar statistical properties within localized neighborhoods \cite{bej_loras_2021}. CSL adaptively identifies appropriate regions for sample generation based on task and dataset characteristics, unlike traditional oversampling methods that rely on predefined heuristics.

Schultz \textit{et al.} \cite{schultz_convgen_2024}, following the work of Bej \textit{et al.}, proposed ConvGeN, a deep neural network model leveraging CSL to address class imbalance in tabular data. This approach integrates CSL with deep generative modeling to synthesize samples that effectively balance target classes in tabular datasets. The oversampling performance of ConvGeN has demonstrated competitive results comparable to state-of-the-art methods for imbalanced classification. However, the scope of the models proposed so far is limited to class balancing for imbalanced classification problems and does not extend to entire tabular data generation. Here, we extend the ConvGeN framework to a broader context, adapting it to generate entire tabular datasets with diverse feature types, thereby expanding its applicability beyond imbalanced learning tasks.

In the realm of synthetic data generation, data modalities such as images and text have garnered significant attention. Substantial advancements in synthetic data generation for computer vision and Natural Language Processing (NLP) are creating promising opportunities, particularly in clinical research. Prominent models in computer vision include Variational AutoEncoders (VAEs) \cite{pesteie_adaptive_2019}, Generative Adversarial Networks (GANs) \cite{creswell_generative_2018}, and Denoising Diffusion Probabilistic Models (DDPMs) \cite{croitoru_diffusion_2023}.

More recently, the development of synthetic data generation techniques has extended to the domain of tabular data, a common format for organizing and analyzing structured information in the form of rows and columns. In tabular data, each column represents a specific feature or attribute, while rows correspond to individual samples \cite{borisov_deep_2022}. The structured and interpretable nature of clinical tabular data facilitates the development of more explainable models, further enhancing its applicability in the clinical domain.

In domains such as biomedicine, tabular data often contain sensitive and confidential information, such as medical records and personal identifiers, making data sharing challenging due to stringent privacy requirements \cite{kamthe_copula_2021}. Such data must be safeguarded to comply with privacy regulations, such as the General Data Protection Regulation (GDPR) in the European Union \cite{sirur_are_2018}. Furthermore, even when access to clinical data is granted, the scale and availability of such data may not always satisfy the demands of modern data-driven methodologies \cite{choi_generating_2017}.

Recent synthetic data generation algorithms have focused on challenges related to data privacy  \cite{borisov_deep_2022}. The strategy is to generate synthetic data replicating real-world clinical datasets' statistical and structural characteristics. By sharing synthetic data in place of real data, privacy risks can be mitigated, as synthetic datasets, in theory, do not reveal sensitive information. Further, synthetic data can be used for downstream analysis instead of real data. For this purpose, the synthetic data must mimic the performance of the real data on diverse downstream tasks to a high degree. In other words, the synthetic data must be high-utility.

However, achieving a balance between privacy and utility is a critical challenge \cite{bagdasaryan_differential_2019}. If the synthetic data too closely resembles the original data in terms of statistical and empirical properties, it may enable adversaries to infer sensitive information, thus undermining privacy protection. Conversely, if the synthetic data diverges significantly from the original data, models developed using the synthetic dataset may yield unreliable or inconsistent results, diminishing its practical utility.

An ideal synthetic data generation framework must effectively preserve both privacy and utility. However, a comprehensive, objective evaluation of how well current tabular synthetic data generation models balance these two critical aspects remains an open research question.

In this paper, we contribute to the ongoing research on synthetic tabular data generation at the following two points:
\renewcommand{\baselinestretch}{1.15}\normalsize
\begin{enumerate}
    \item We generalize the existing ConvGeN algorithm such that it can be used for synthetic tabular data generation in general, thereby proposing the NextConvGeN algorithm. The model comprises a generator and discriminator network, which cooperate among themselves (instead of competing like in adversarial learning) to learn regions in the convex data space of data neighborhoods that are more apt for sampling synthetic data points. We demonstrate the model's effectiveness in comparison with existing tabular synthetic data generation algorithms using publicly available benchmarking datasets from the clinical domain. Using CSL for the synthetic data generation ensures that the synthetic data satisfies certain theoretical aspects that we elaborate more in Section \ref{discussion}.
    \item We provide an objective comparison of major tabular synthetic data generation models in terms of how they balance between privacy and utility-based performance measures.
\end{enumerate}
\section{Related research}\label{literature survey} 
VAEs and GANs are two pioneering techniques in generative modeling that have significantly shaped research on synthetic data. Initially used for synthetic image generation, now they have found diverse applications across various domains \cite{sami_comparative_2019}. This section discusses the state-of-the-art synthetic tabular data generative models.\par

In 2019, Xu \textit{et al.} introduced CTGAN, a method for synthesizing tabular data that effectively models both continuous and categorical columns \cite{xu_modeling_2019}. CTGAN uses a unique normalization approach for continuous features and a strategy to handle imbalances in discrete data. Continuous features are processed using a variational Gaussian mixture model, while categorical columns are represented through one-hot encoding. The model incorporates a conditional vector to address imbalances and prevent mode collapse during training. CTGAN outperforms other generative models, such as CLBN, PrivBN, MedGAN, VeeGAN, and TableGAN, in machine learning efficacy on real datasets \cite{xu_modeling_2019}.\par

CTAB-GAN, an extension of CTGAN, was developed to address challenges with imbalanced data and skewed distributions, which are common in various datasets \cite{han_borderline-smote_2005}. CTAB-GAN combines conditional generation and sampling-based training techniques to enhance class representation, particularly for underrepresented classes. The method employs logarithmic transformation to compress and reduce the distance between tail and bulk data, aiding in more accurate synthesis. CTAB-GAN introduces information loss and classification loss into the generator's loss function to ensure the quality of synthetic data. It has been shown to produce synthetic data that is statistically similar to real data and useful for machine learning tasks \cite{zhao_ctab-gan_2021}.\par

Building on the advancements of CTGAN and CTAB-GAN, Zhao \textit{et al.} developed CTAB-GAN+ to address the limitations of GAN-based tabular data generators, particularly in handling variables that can be both numerical and categorical. CTAB-GAN+ incorporates the Wasserstein plus gradient penalty to stabilize GAN training and differential privacy in the encoder to protect individual data points while allowing meaningful aggregate analysis \cite{wang_generative_2017}. This model outperforms others in terms of machine learning utility and statistical similarity, making it a robust choice for synthetic data generation \cite{zhao_ctab-gan_2023}.

TabDDPM, introduced by Kotelnikov \textit{et al.}, adapts diffusion models, commonly used for image generation, to create synthetic tabular data \cite{kotelnikov_tabddpm_2023}. The model preprocesses continuous features using Gaussian quantile transformation and represents categorical features with one-hot encoding. TabDDPM employs multinomial diffusion for categorical features and Gaussian diffusion for continuous ones, with a multilayer perceptron modeling the reverse diffusion step. Although it performs well on benchmark datasets, generating high-quality synthetic samples, the model's reliance on one-hot encoding for categorical features can lead to data sparsity, especially in small sample sizes, and may not fully capture relationships between different feature types \cite{kotelnikov_tabddpm_2023}.\par

\section{Methodology}
The NextConvGeN model is a framework designed to generate synthetic tabular data using CSL. This section explains how the model works, including its architecture, key concepts, and step-by-step process.
\subsection{Overview of NextConvGeN}
NextConvGeN builds on the ConvGeN model by expanding its ability to create synthetic data without depending on class imbalances. It uses \textit{deep cooperative learning}, where a generator ($G$) and a discriminator ($D$) work together to improve the quality of generated data. Unlike ConvGeN, which focuses only on minority class neighborhoods, NextConvGeN works with neighborhoods across the entire dataset. The generator creates synthetic samples within the convex hull of the data neighborhoods, while the discriminator separates these synthetic samples from unrelated data points. The feedback between $G$ and $D$ helps improve the quality of the samples with each training step.
\begin{figure}[ht]
    \centering
    \includegraphics[clip, trim=0cm 2cm 0cm 1cm,width=\textwidth]{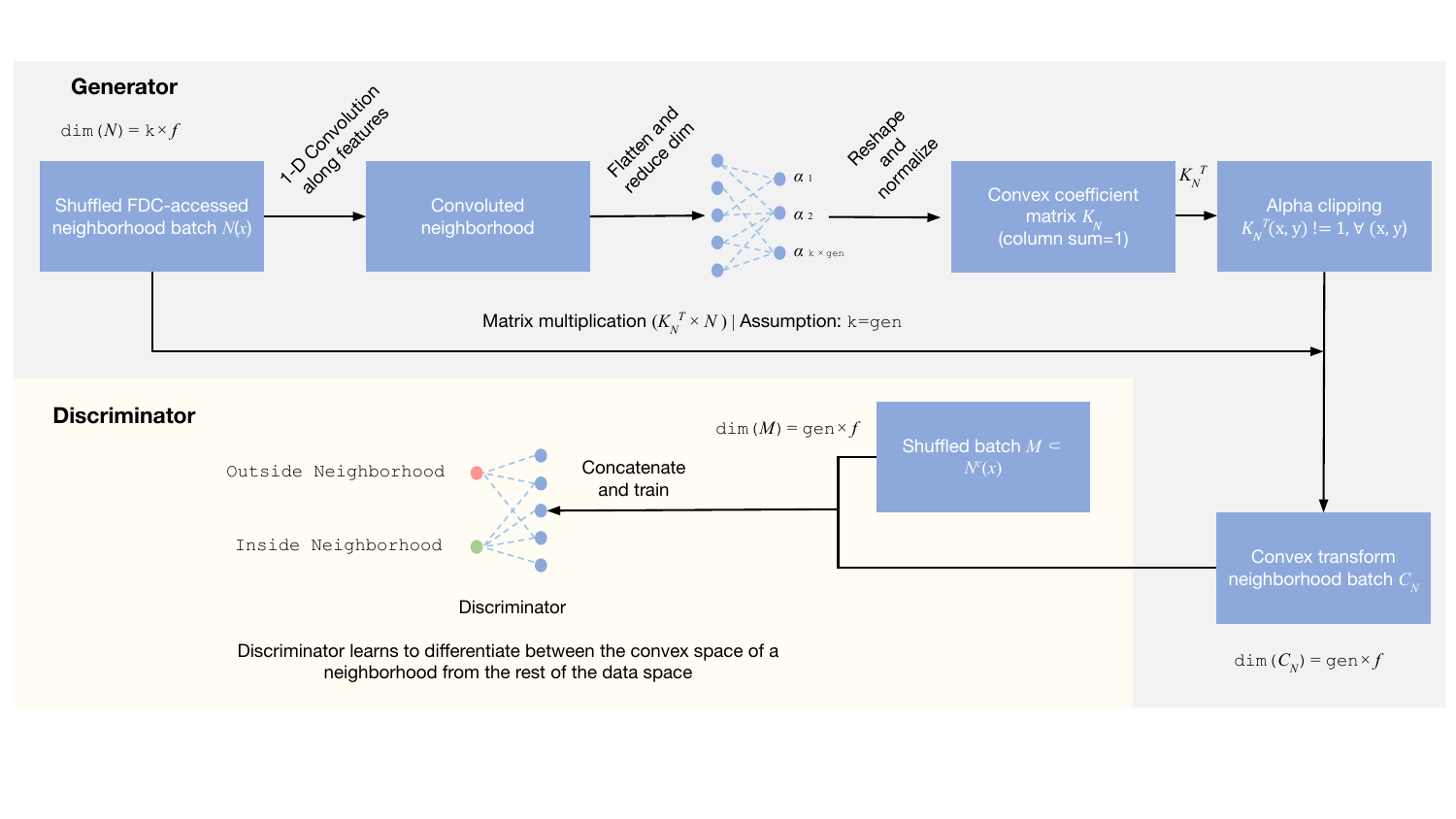}
    \caption[NextConvGeN architecture]{Architecture of the NextConvGeN model. The generator component takes shuffled data neighborhood batches as input and produces a convex combination of the samples from the neighborhood. The discriminator component is responsible for classifying these generated samples and comparing them against a randomly selected outside neighborhood batch of the same size as the synthetic inside neighborhood. The discriminator can be further trained and repurposed as the discriminator as a classifier after completing the NextConvGeN model training.}
    \label{fig_nextconvgen}
\end{figure}
\subsection{Formal framework}
Let $X$ denote the dataset, $x \in X$ a data instance, and $N(x) \subseteq X$ its $k$-neighborhood, derived through a nonlinear dimensional reduction method such as \textit{Feature-type Distributed Clustering (FDC)} \cite{bej_accounting_2023}. The generator maps the neighborhood $N(x)$ onto its convex hull, $\mathbb{C}_{N(x)}$ \[
G: N(x) \to C_N(x) \subset \mathbb{C}_{N(x)}
\] where $C_N(x)$ represents the convex combinations of elements within $N(x)$. The discriminator evaluates the separability of $C_N(x)$ from a disjoint batch $M \subset N^c(x)$, with $N^c(x)$ signifying the complement of $N(x)$ in $X$. This is formalized as \[
D: C_N(x) \oplus M \to \big\{(1,0), (0,1)\big\}
\] where $\oplus$ signifies batch concatenation, the optimization objective involves minimizing the binary cross-entropy loss function
\[
L_D = -\sum_{c \in C_N, y \in M} \log\Bigl(D(c)\Bigl) + \log\Bigl(1 - D(y)\Bigl)
\]
The generator undergoes indirect optimization through discriminator feedback, ensuring the synthetic samples retain the statistical integrity of the input neighborhood. The final objective incorporates a Mean-Squared Error (MSE) loss \[ L_{\text{Final}} = \text{MSE}(Y_{N \oplus M}, \hat{Y}_{N \oplus M})\] where $Y_{N \oplus M}$ and $\hat{Y}_{N \oplus M}$ correspond to the ground truth and predicted labels, respectively.

\subsection{Architectural Design}
\subsubsection{Generator}
The generator network ingests input neighborhoods $N(x) \in \mathbb{R}^{\texttt{k} \times f}$, where $\texttt{k}$ denotes the neighborhood size and $f$ the feature dimensionality. A 1-D convolutional layer reduces dimensionality, yielding a convex coefficient matrix $K_N \in \mathbb{R}^{\texttt{k} \times \texttt{k}}$. The matrix adheres to the following constraints:
\[
\sum_{j} K_{N}(i,j) = 1, \quad K_{N}(i,j) \geq 0 \; \forall \; i, j
\]

To mitigate excessive similarity between synthetic and original samples, an \textit{alpha-clipping} technique adjusts the extremal row values of $K_N^T$. This adjustment ensures that all coefficients remain within the convexity constraints while preventing the synthetic sample from being similar to any real sample. More information on alpha clipping is provided in the appendices. The convex transformation is subsequently applied:
\[
C_N = K_N^T N
\]
where each row of $C_N$ represents a synthetic sample derived from $N(x)$.

\subsubsection{Discriminator}
The discriminator processes concatenated batches $C_N \oplus M \in \mathbb{R}^{\texttt{gen} \times f}$, where $\texttt{gen} = \texttt{k}$. The architecture comprises three fully connected hidden layers with 250, 125, and 75 units, ending with a binary classification output layer with two nodes. The discriminator differentiates between synthetic samples ($C_N$) and real data points outside the neighborhood $M$.

\subsection{Training Procedure}
The training protocol follows the stages listed below:
\renewcommand{\baselinestretch}{1.15}\normalsize
\begin{enumerate}
    \item \textbf{Neighborhood sampling:} Derivation of input neighborhoods $N(x)$ via FDC, ensuring significant clustering across diverse feature types.
    \item \textbf{Synthetic sample generation:} The generator produces $C_N$ from $N(x)$ using learned convex coefficients.
    \item \textbf{Discriminator feedback:} The discriminator evaluates $C_N$ relative to $M$ and updates its parameters by minimizing $L_D$.
    \item \textbf{Generator refinement:} The generator parameters are indirectly adjusted through discriminator feedback to enhance synthetic sample fidelity.
\end{enumerate}

This iterative optimization facilitates convergence, with $G$ and $D$ mutually reinforcing their respective functionalities.

\begin{algorithm*}[!htbp]
\small
    \SetArgSty{textnormal}
    % \NoCaptionOfAlgo
    \caption{NextConvGeN}
    \label{algo:convGen:train}
    %%%%%%%%%%%%%%%%%%%%%%%%%%%%%%% 
    % Parameter section starts here
    %%%%%%%%%%%%%%%%%%%%%%%%%%%%%%%
    \SetKwInOut{In}{Inputs}
    \In{\newline
        \vspace{-0.5em}
        \begin{flushleft}
        \begin{tabular}{ l l }
            \var{data} & Data points to train on.
        \end{tabular}
        \end{flushleft}
    }
    \SetKwInOut{Parameter}{Parameters}
    \Parameter{%\newline
        \begin{flushleft}
        \vspace{-0.5em}
        \begin{tabular}{ l l p{8cm} }
            \var{disc\_train\_count} & $\geq 0$ & Extra training steps for the discriminator before updating  
            generator once. (used value: $5$)\\
            \var{k} & $\geq 2$ & Size of the neighborhood used to model convex data space. \\
            \var{gen} & $\geq \var{k}$ & Number of generated synthetic points by generator at a time. (used value: $\var{gen} = \var{k}$) \\
            \var{fdc} & ~ & A function that normalizes the data. \\
        \end{tabular}
        \end{flushleft}
    }
    %%%%%%%%%%%%%%%%%%%%%%%%%% 
    % Code section starts here
    %%%%%%%%%%%%%%%%%%%%%%%%%%
    \SetKwFor{For}{For}{do}{endfor}
    \SetKwIF{If}{ElseIf}{Else}{If}{then}{else If}{else}{endif}
    
    \vspace{1em}

    \textbf{Function} \var{train}(\var{data})
    
    \Begin{

        $\var{normalized\_data} \leftarrow$ fdc(\var{data})

        \var{neighbourhoods}  $\leftarrow$ Build neighbourhoods with \var{k} points size for all  \var{x} $\in \var{data}$ with closeness according to \var{normalized\_data}.

        $\var{labels} \leftarrow \big( \underbrace{(1, 0), \ldots ,(1, 0)}_{gen} , \underbrace{(0, 1), \ldots ,(0, 1)}_{gen} \big)$
        % \hfill{} (1, 0) / (0,1) are the label for minority / majority class.
        \vspace{3pt}

        \For{$1 \ldots \var{neb\_epochs}$}{
            \vspace{5pt}
            \For{$1 \ldots$ \var{disc\_train\_count}}{
                \For{\var{x} $\in$ data}{
                    %\var{batch\_x} $\leftarrow$ \var{k} nearest points to \var{x} in randomized order according to \var{neighborhoods}.
                    
                    \var{batch\_x} $\leftarrow$ neighborhood for \var{x} from \var{neighborhoods} in randomized order.
                    
                    \var{batch\_no\_x} $\leftarrow$ \var{gen} random points from \var{normalized\_data} without \var{batch\_x}.

                    \var{conv\_samples} $\leftarrow$ Generate \var{gen} points using the neighborhood of \var{x}.

                    \var{corrected\_samples} $\leftarrow \var{correct\_feature\_types}(\var{batch\_x}, \var{conv\_samples})$
                    
                    \var{concat\_sample} $\leftarrow (\var{corrected\_samples} \oplus \var{batch\_no\_x})$
                    
                    train discriminator with \var{concat\_sample} and \var{labels}
                }
            }
            \vspace{1em}
            
            \For{\var{x} $\in$ data}{
                %\var{batch\_x} $\leftarrow$ \var{k} nearest points to \var{x} in randomized order.
                \var{batch\_x} $\leftarrow$ neighborhood for \var{x} from \var{neighborhoods} in randomized order.

                \var{batch\_no\_x} $\leftarrow$ \var{gen} random points from \var{normalized\_data} without \var{batch\_x}.

                \var{concat\_sample} $\leftarrow (\var{generator}(\var{batch\_x}) ~\oplus~ \var{batch\_no\_x})$
                
                train generator by using
                $ \var{discriminator}( \var{concat\_sample} )$ and \var{labels}
            }
        }
    }
\end{algorithm*}

\subsection{Key Innovations}
\renewcommand{\baselinestretch}{1.15}\normalsize
\begin{enumerate}
    \item \textbf{Generalization beyond class imbalance:} NextConvGeN operates independently of class labels, extending its utility to diverse synthetic data generation scenarios.
    \item \textbf{Feature-Distributed Clustering:} FDC applies specialized similarity metrics tailored to continuous, ordinal, and nominal features, improving neighborhood relevance.
    \item \textbf{Alpha-clipping:} This mechanism prevents synthetic samples from closely resembling original data points, safeguarding data utility and privacy.
\end{enumerate}

\section{Datasets and evaluation measures}
\subsection{Datasets used and general protocol} 
In our benchmarking experiments, we selected ten publicly available datasets specifically from the biomedical domain. These datasets were chosen because biomedical data often exhibit characteristics such as small sample sizes, diverse feature types, and data imbalance, which pose challenges for synthetic data generation. A detailed overview of each dataset is provided in Table \ref{dataset_table}, highlighting key attributes that reflect their diversity and relevance to our study.

\begin{table*}[htbp]
\scriptsize
\caption{Description of the datasets used in our benchmarking studies. This table provides information on the dataset train and test sizes, the number of attributes, and the distribution of ordinal, nominal, and continuous features within each dataset.}\vspace{-5pt}
\begin{center}
 \begin{tabular}{c|c|c|c|c|c|c} \hline
  {\textbf{Dataset}}&  {\textbf{Train size}} & {\textbf{Test size}} & {\textbf{Features}} &  {\textbf{Ordinal}} & {\textbf{Nominal}} & {\textbf{Continuous}}\\
   \hline
   
   \href{https://www.kaggle.com/datasets/andrewmvd/heart-failure-clinical-data}{Heart failure} & $209$ & $90$ & $12$ & $5$ & $0$ & $7$ \\
   \hline

   \href{https://doi.org/10.24432/C52P4X}{Heart disease} & $211$ & $91$ & $13$ & $4$ & $4$ & $5$ \\
   \hline

   \href{https://www.kaggle.com/datasets/mysarahmadbhat/lung-cancer}{Lung cancer} & $216$ & $93$ & $15$ & $5$ & $9$ & $1$ \\
   \hline
   
   \href{https://codeocean.com/capsule/1269964/tree/v1/data/migraine.csv}{Migraine} & $263$ & $114$ & $23$ & $6$ & $16$ & $1$\\
  \hline
  
   \href{https://www.kaggle.com/datasets/fedesoriano/cirrhosis-prediction-dataset/code}{Liver cirrhosis} & $292$ & $126$ & $18$ & $4$ & $3$ & $11$ \\
   \hline

   \href{https://doi.org/10.24432/C5D02C}{Indian liver patients} & $396$ & $170$ & $10$ & $0$ & $1$ & $9$ \\
   \hline

   \href{https://www.kaggle.com/datasets/uciml/pima-indians-diabetes-database}{Pima Indian diabetes} & $537$ & $231$ & $8$ & $1$ & $0$ & $7$ \\
   \hline

   \href{https://doi.org/10.24432/C59W2D}{Contraceptive methods} & $997$ & $428$ & $9$ & $4$ & $4$ & $1$ \\
   \hline

   \href{https://doi.org/10.24432/C5H31Z}{Obesity} & $1460$ & $627$ & $16$ & $9$ & $4$ & $3$ \\
   \hline

   \href{https://www.kaggle.com/datasets/fedesoriano/stroke-prediction-dataset/code}{Stroke} & $3435$ & $1473$ & $10$ & $3$ & $4$ & $3$ \\
   \hline
\end{tabular}
\label{dataset_table}
\end{center}
\end{table*}
\vspace{-10pt}
We preprocessed the datasets to eliminate missing values and ensure all variables were numeric, as some generative models cannot handle non-numeric data. We specified feature types (continuous, ordinal, nominal) for the models and split each dataset into 70\% for training and 30\% for evaluation. To assess the NextConvGeN model, we compared it against five state-of-the-art generative models, generating synthetic samples five times the size of the training data. These were downsampled to match the target feature's cardinality in the real data, ensuring a fair comparison for utility and privacy evaluation.

\subsection{Evaluation measures}

Evaluating synthetic data quality is critical yet challenging due to a lack of standardized metrics for tabular data \cite{hernandez_synthetic_2022}. In this study, we employ utility and privacy measures for assessment. Utility metrics focus on ML task performance and fidelity, which measures statistical similarities between real and synthetic data \cite{jordon_synthetic_2022}. Privacy measures, on the other hand, quantify the risk of information leakage. Table \ref{performance measure} summarizes the evaluation metrics used in this study.

\begin{table*}[htbp] \scriptsize
\caption{Evaluation measures used for benchmarking analysis. This table summarizes various evaluation measures to compare generative models. The first column details the measure itself. the second column clarifies whether the measure focuses on model utility or privacy. the third column specifies the possible range of values for the measure. Finally, the last column indicates the optimal direction for the score (smaller or larger) to signify better performance.}\vspace{-5pt}
\begin{center} 
    \begin{tabular}{c|c|c|c} \hline {\textbf{Measure}}& {\textbf{Type of evaluation}} & {\textbf{Range}} & {\textbf{Optimal direction}} \\ \hline {Student's t test} & utility & $[0,1]$ & p-value $>0.05$\\ \hline {KL divergence} & utility & $[0,+\infty]$ & smaller\\ \hline {Cross validation} & utility & $[0,1]$ & smaller\\ \hline {Holdout data analysis} & utility & $[0,1]$ & smaller\\ \hline {Propensity score} & utility & $[0, 0.25]$ & smaller\\ \hline {Log-cluster metric} & utility & $[0,+\infty]$ & larger\\ \hline {Cross classification} & utility & $[0,1]$ & smaller\\ \hline {MIA} & privacy & $[0,1]$ & smaller\\ \hline {AIA on categorical features} & privacy & $[0,1]$ & smaller\\ \hline {AIA on continuous features} & privacy & $[0,+\infty]$ & smaller\\ \hline {Euclidean distance} & privacy & $[0,+\infty]$ & larger\\ \hline {Hausdorff distance } & privacy & $[0,+\infty]$ & larger\\ \hline {Cosine similarity} & privacy & $[0,1]$ & smaller\\ \hline \end{tabular} 
\label{performance measure} \end{center} \end{table*} 
\vspace{-15pt}
\subsection{Utility measures}
Utility measures assess the quality of synthetic data by comparing its statistical properties and predictive performance to real data. Common measures include:
\renewcommand{\baselinestretch}{1.15}\normalsize
\begin{itemize}
    \item \textbf{Student's t-test:} A statistical test used to evaluate if synthetic data preserves the mean of continuous features from real data. A p-value $>0.05$ indicates preservation \cite{hernandez_synthetic_2022}.
    \item \textbf{KL divergence:} This metric quantifies the difference between the probability distributions of categorical features in real and synthetic data. Smaller values suggest higher fidelity \cite{goncalves_generation_2020}.
    \item \textbf{Holdout data analysis:} Evaluates the predictive performance of classifiers trained on real and synthetic data when tested on unseen real data (holdout data). Smaller differences in F1-scores reflect higher utility. \cite{espinosa_quality_2023}.
    \item \textbf{Propensity score:} Assesses how distinguishable synthetic data is from real data using logistic regression. Scores closer to zero indicate higher similarity \cite{pathare_comparison_2023, espinosa_quality_2023}.
    \item \textbf{Log-cluster metric:} Evaluates clustering performance on combined real and synthetic data. A higher score indicates poor differentiation by the clustering algorithm \cite{goncalves_generation_2020}.
    \item \textbf{Cross-classification:} Compares statistical dependencies between features in real and synthetic data. Smaller differences in prediction accuracy indicate better preservation of dependencies \cite{dankar_multi-dimensional_2022}. 
\end{itemize}

\subsubsection{Privacy measures}
Privacy measures evaluate the extent to which synthetic data protects sensitive information:
\renewcommand{\baselinestretch}{1.15}\normalsize
\begin{itemize}
\item \textbf{Distance metrics:} Includes Euclidean, Hausdorff, and Cosine similarity measures. Smaller distances or higher similarity scores suggest higher privacy risks \cite{hernandez_synthetic_2022}.
\item \textbf{Membership Inference Attack (MIA):} Determines the likelihood of a specific record being part of the training data. Lower precision values ($< 0.5$) indicate strong privacy protection \cite{mendelevitch_fidelity_2021}.
\item \textbf{Attribute Inference Attack (AIA):} Assesses the risk of inferring sensitive attributes from synthetic data. Low accuracy or high RMSE values signify lower privacy risks \cite{hernandez_synthetic_2022}.
\end{itemize}

For a detailed description of the evaluation measures, please refer to the appendices.

\section{Results}\label{results}
This section uses tables and figures to present the evaluation results of synthetic data generated by tabular data generative models across ten publicly available clinical datasets.The results presented are the average values over five runs, each conducted with different random seeds, and the standard deviations for these results are provided in the GitHub repository. In the tables, if there is a cell filled with NaN, it indicates that we could not sample synthetic data generated for that dataset by that model to match the cardinality of the target column in the real data. Furthermore, we compared the performance as an average across the ten datasets for privacy assessment. TVAE and CTAB-GAN+ were not included in the privacy comparison because, for certain datasets, these models could not match the cardinality of the synthetic target column to it's corresponding real data.

\begin{table*}[htbp]\scriptsize\caption{Table displaying p-values for Student's t-test comparing real and synthetic data's continuous features across ten benchmarking datasets. p $> 0.05$ (marked in bold fonts in the table) suggests agreement with the null hypothesis, indicating similarity in feature means between real and synthetic data.}\label{t test result}\centering\tabularnewline\vspace{5pt}\begin{tabular}{@{\hskip4pt}c@{\hskip4pt}|@{\hskip4pt}c@{\hskip4pt}|@{\hskip4pt}c@{\hskip4pt}|@{\hskip4pt}c@{\hskip4pt}|@{\hskip4pt}c@{\hskip4pt}|@{\hskip4pt}c@{\hskip4pt}|@{\hskip4pt}c@{\hskip4pt}}\hline
\textbf{dataset}& \textbf{TVAE} & \textbf{CTGAN}  & \textbf{CTAB-GAN} & \textbf{CTAB-GAN+} & \textbf{TabDDPM} & \textbf{NextConvGeN}
\tabularnewline
\hline
\hline
Heart failure & \textbf{0.1402} & \textbf{0.0894} & 0.0390 & \textbf{0.1563} & \textbf{0.0958} & 0.0349 \tabularnewline

Heart disease & \textbf{0.1274} & 0.0012 & 0.0499 & \textbf{0.0979}& \textbf{0.4482} & 0.0190 \tabularnewline

Lung cancer & NaN & 0.0441 & 0.0029 & \textbf{0.1755}& \textbf{0.5898} & \textbf{0.4915} \tabularnewline

Migraine & NaN &  \textbf{0.1598} & \textbf{0.1393} & NaN &\textbf{0.4925} & \textbf{0.7074} \tabularnewline

Liver cirrhosis & NaN & \textbf{0.0657} & \textbf{0.0713} &\textbf{0.0918} & \textbf{0.1186} & 0.0375 \tabularnewline

Indian liver patients & \textbf{0.0728} & \textbf{0.0546} & 0.0377 & \textbf{0.1331} & \textbf{0.4163} & \textbf{0.7143} \tabularnewline

Pima Indian diabetes & \textbf{0.1994} & \textbf{0.0905} & \textbf{0.1263} & \textbf{0.1387} & 0.0280 & \textbf{0.6303} \tabularnewline

Contraceptive methods & \textbf{0.5098} & 0.0000 & \textbf{0.1584} & \textbf{0.1112} & \textbf{0.5991} & 0.0014 \tabularnewline

Obesity & \textbf{0.1630} & 0.0076 & 0.0001 & NaN & \textbf{0.6091} & \textbf{0.6555} \tabularnewline

Stroke & NaN & 0.0493 & 0.0000 & \textbf{0.0844} & \textbf{0.5409} & \textbf{0.3068} \tabularnewline
\hline\end{tabular}\end{table*}

\textbf{TabDDPM and NextConvGeN preserve distributions of both continuous and categorical features in synthetic data:} 
We evaluated the similarity of continuous feature distributions in real and synthetic data using the average p-values from Student's t-test (Table \ref{t test result}). A p-value above $0.05$ suggests a strong resemblance. The results indicate that TVAE and CTAB-GAN+ consistently yield p-values above this threshold, implying high similarity. However, mode collapse limited our ability to maintain target variable cardinality for some datasets. TabDDPM produced a p-value below $0.05$ for only one dataset, reinforcing its effectiveness. For NextConvGeN, CTGAN, and CTAB-GAN, we observed p-values above $0.05$ in six, five, and four datasets, respectively.

\begin{table*}[htbp]\scriptsize\caption{The table illustrates the KL divergence measure between real and synthetic data's categorical features across benchmark datasets. A lower value of the KL divergence score indicates a smaller difference between the real and synthetic data distributions. TabDDPM emerges as the top performer, consistently outperforming the other three models across most datasets.}\label{KL divergence}\centering\tabularnewline\vspace{5pt}\begin{tabular}{@{\hskip4pt}c@{\hskip4pt}|@{\hskip4pt}c@{\hskip4pt}|@{\hskip4pt}c@{\hskip4pt}|@{\hskip4pt}c@{\hskip4pt}|@{\hskip4pt}c@{\hskip4pt}|@{\hskip4pt}c@{\hskip4pt}|@{\hskip4pt}c@{\hskip4pt}}\hline
\textbf{dataset} & \textbf{TVAE} & \textbf{CTGAN}  & \textbf{CTAB-GAN} &\textbf{CTAB-GAN+}  & \textbf{TabDDPM} & \textbf{NextConvGeN}
\tabularnewline
\hline
\hline
Heart failure & 0.1726 & \textbf{0.0056} & 0.0307  & 0.0077 & 0.0228 & 0.0059 \tabularnewline
Heart disease & 0.0617 & 0.0106 & 0.2300 & 0.0148 & \textbf{0.0051} & 0.0133 \tabularnewline
Lung cancer & NaN & 0.0134 & 0.0099 & 0.0081 &  0.0023 & \textbf{0.0018} \tabularnewline
Migraine & NaN & 0.0128 & 0.3568 & NaN & \textbf{0.0024} & 0.0030 \tabularnewline
Liver cirrhosis & NaN & 0.0098 & 0.2264 & 0.0082 & 0.0809 & \textbf{0.0045} \tabularnewline
Indian liver patients & 0.3083 & 0.0272 & 0.0606 & 0.0082 & \textbf{0.0004} & \textbf{0.0004} \tabularnewline
Pima Indian diabetes & \textbf{0.0000} & 0.0342 & 0.1383 & 0.0179 & 0.0117 & 0.0057 \tabularnewline
Contraceptive Methods & 0.1026 & 0.0092 & 0.2400 & 0.0088 & \textbf{0.0014} & 0.0029 \tabularnewline
Obesity & 0.0590 & 0.0314 & 0.3701 & NaN & \textbf{0.0010} & 0.0023 \tabularnewline
Stroke & NaN & 0.0215 & 0.1561 & 0.0034 & \textbf{0.0005} & 0.0029 \tabularnewline
\hline\end{tabular}\end{table*}

Categorical feature distributions of real and synthetic data were assessed via KL divergence (Table \ref{KL divergence}), where lower scores indicate a closer resemblance to real data. TabDDPM outperformed all models, followed closely by NextConvGeN. In contrast, TVAE, CTGAN, and CTAB-GAN exhibited higher KL divergence scores, indicating poorer preservation of empirical distributions. CTAB-GAN+ achieved a lower KL divergence among GAN-based models, outperforming TVAE and CTGAN. Overall, TabDDPM and NextConvGeN generate synthetic continuous and categorical variables with distributions closely matching real data.

\begin{table}[htbp]\scriptsize\caption{Table illustrating the log-cluster metric across ten benchmarked datasets. A higher magnitude value indicates increased challenges in effectively clustering real from synthetic data. TabDDPM outperforms other models with a larger average magnitude across datasets.}\label{Log cluster metric}\centering\tabularnewline\vspace{5pt}\begin{tabular}{@{\hskip4pt}c@{\hskip4pt}|@{\hskip4pt}c@{\hskip4pt}|@{\hskip4pt}c@{\hskip4pt}|@{\hskip4pt}c@{\hskip4pt}|@{\hskip4pt}c@{\hskip4pt}|@{\hskip4pt}c@{\hskip4pt}|@{\hskip4pt}c@{\hskip4pt}}\hline
\textbf{dataset} & \textbf{TVAE}  & \textbf{CTGAN} & \textbf{CTAB-GAN} & \textbf{CTAB-GAN+}& \textbf{TabDDPM} & \textbf{NextConvGeN}
\tabularnewline
\hline
\hline
Heart failure & -9.5402 & -7.8421  & -6.2545  & \textbf{-11.3612} & -4.0491 & -2.0681 \tabularnewline
Heart disease & -8.7799 & -5.0021  & -2.3625 & -7.2651 & \textbf{-10.9247} & -2.0670 \tabularnewline
Lung cancer & NaN & -6.1234 & -6.9445 & -8.3901 & -9.0033 & \textbf{-9.3914} \tabularnewline
Migraine & NaN & -5.2308 & -1.6343 & NaN & \textbf{-21.4164} & \textbf{-21.4164} \tabularnewline
Liver Cirrhosis & NaN &  -3.0929 & -2.0673 & \textbf{-6.1925} & -2.7663 & -2.0703 \tabularnewline
Indian liver patients & -2.4319 & -4.9052 & -2.1909  & -5.8815 & -6.5057 & \textbf{-8.9846} \tabularnewline
Pima Indian diabetes & -6.1012 & -8.9490 & -6.3372 & -6.9917 & -2.0017 & \textbf{-12.7566} \tabularnewline
Contraceptive methods & -5.0534 & -6.5806 & -2.3289 & -7.1532 &  \textbf{-10.5784} & -9.2276 \tabularnewline
Obesity & -6.6065 & -7.7256 & -1.5535  & NaN & \textbf{-11.0229} & -10.9935 \tabularnewline
Stroke & NaN & -9.5835 & -2.6115  & -7.7486 &  \textbf{-14.9799} & -9.1026 \tabularnewline
\hline
Average & -6.4189 & -6.5035 &	-3.4285 & -7.6230 & \textbf{-9.3248} & -8.8078
\tabularnewline
\hline\end{tabular}\end{table}

\textbf{Synthetic data from TabDDPM preserves the distribution of clusters in real data:}
Table \ref{Log cluster metric} presents log-cluster values for various generative models across benchmark datasets, measuring the similarity of cluster distributions in real and synthetic data. A higher value indicates greater similarity. TabDDPM achieves the highest log-cluster values in five out of ten datasets, outperforming other models. NextConvGeN follows closely, with the second-highest average magnitude. These results suggest that TabDDPM and NextConvGeN best preserve cluster distributions in synthetic data.

\begin{table}[htbp]\scriptsize\caption{The table presents an absolute five-fold cross-validation performance difference between gradient-boosting classifiers trained on real and their corresponding synthetic data across benchmarking datasets. The lower the difference, the better the utility. NextConvGeN outperforms other models in most of the datasets. }\label{Cross validation}\centering\tabularnewline\vspace{5pt}\begin{tabular}{@{\hskip4pt}c@{\hskip4pt}|@{\hskip4pt}c@{\hskip4pt}|@{\hskip4pt}c@{\hskip4pt}|@{\hskip4pt}c@{\hskip4pt}|@{\hskip4pt}c@{\hskip4pt}|@{\hskip4pt}c@{\hskip4pt}|@{\hskip4pt}c@{\hskip4pt}}\hline
\textbf{dataset} & \textbf{TVAE} & \textbf{CTGAN} & \textbf{CTAB-GAN} & \textbf{CTAB-GAN+} & \textbf{TabDDPM} & \textbf{NextConvGeN}
\tabularnewline
\hline
\hline
Heart failure & \textbf{0.0154} & 0.1694 & 0.1770 & 0.1245 & 0.0755 & 0.0506 \tabularnewline
Heart disease & 0.0921 & 0.3373 & 0.3212 & 0.0483 & 0.0646 & \textbf{0.0367} \tabularnewline
Lung cancer & NaN & 0.0685 & 0.0601 & 0.0741 & 0.0258 & \textbf{0.0250} \tabularnewline
Migraine & NaN & 0.2928 & 0.2791 & NaN & \textbf{0.0597} & 0.0979 \tabularnewline
Liver cirrhosis & NaN & 0.1554 & 0.1258 & \textbf{0.0676} & 0.2071 & 0.1180 \tabularnewline
Indian liver patients & 0.0702 & 0.0227 & \textbf{0.0202}  & 0.0434 & 0.1305 & 0.1071 \tabularnewline
Pima Indian diabetes & 0.1118 & 0.1448 & 0.1418 & 0.1169 & \textbf{0.0555} & 0.0969 \tabularnewline
Contraceptive methods & 0.1655 & 0.1523  & 0.1374 & 0.1575 &  0.0517 & \textbf{0.0457} \tabularnewline
Obesity & 0.2181 & 0.6895  & 0.6936 & NaN & \textbf{0.0278} & 0.0348 \tabularnewline
Stroke & NaN & \textbf{0.0008} & 0.0013 & 0.0018 &  0.0026 & 0.0044 \tabularnewline
\hline\end{tabular}\end{table}

\textbf{Classifiers trained on real data and NextConvGeN, TabDDPM synthetic data shows similar performance:}
Table \ref{Cross validation} compares the absolute cross-validation difference between the classifiers trained on real and synthetic data. The cross-validation difference is expected to be smaller when the synthetic data is similar to real data. We observe that out of ten benchmarking datasets, the NextConvGeN and TabDDPM models outperform other synthetic data generation models for three datasets each, achieving the best overall performance.

\begin{table}[htbp]
    \scriptsize
    \caption{Table presenting the absolute difference in F1-score between classifiers trained on real and synthetic data with both models tested on the same holdout dataset across benchmarked datasets. This shows that the model trained on NextConvGeN synthetic data produces more consistent performance on unseen data. }
    \label{f1_score_validation}\centering\tabularnewline\vspace{5pt}\begin{tabular}{@{\hskip4pt}c@{\hskip4pt}|@{\hskip4pt}c@{\hskip4pt}|@{\hskip4pt}c@{\hskip4pt}|@{\hskip4pt}c@{\hskip4pt}|@{\hskip4pt}c@{\hskip4pt}|@{\hskip4pt}c@{\hskip4pt}|@{\hskip4pt}c@{\hskip4pt}}\hline
    \textbf{dataset} & \textbf{TVAE} & \textbf{CTGAN}  & \textbf{CTAB-GAN} & \textbf{CTAB-GAN+} & \textbf{TabDDPM} & \textbf{NextConvGeN}
    \tabularnewline
    \hline
    \hline
    Heart failure & 0.2337 & 0.3902 & 0.2252 & 0.2867 & 0.3181 & \textbf{0.0621} \tabularnewline
    
    Heart disease & \textbf{0.0398} & 0.0568 & 0.1554 & 0.1265 & 0.1824 & 0.2338 \tabularnewline
    
    Lung cancer & NaN & \textbf{0.0000} & \textbf{0.0000} & \textbf{0.0000} & \textbf{0.0000} & \textbf{0.0000} \tabularnewline
    
    Migraine & NaN & 0.2554 & \textbf{0.0374} & NaN & 0.0585 & 0.1365 \tabularnewline
    
    Liver cirrhosis & NaN & 0.1044 & 0.1132 & \textbf{0.0687} & 0.1107 & 0.0904 \tabularnewline
    
    Indian liver patients & \textbf{0.0000} & 0.2885 & 0.0968 & 0.0640 & \textbf{0.0000} & 0.0114 \tabularnewline
    
    Pima Indian diabetes & 0.0736 & 0.1250 & 0.1454 & 0.1384 & 0.0930 & \textbf{0.0194} \tabularnewline
    
    Contraceptive methods & 0.1135 & 0.1427 & 0.1192 & 0.1339 & \textbf{0.0754} & 0.0991 \tabularnewline
    
    Obesity & 0.0958 & \textbf{0.0246} & 0.0379 & NaN & 0.0292 & 0.0432 \tabularnewline
    
    Stroke & NaN & 0.2155 & \textbf{0.1432} & 0.2118 & 0.1794 & 0.2598 \tabularnewline
    
    \hline
    \end{tabular}
\end{table}

\textbf{Assessment of predictive consistency of classifiers trained on synthetic data on unseen data:} Table \ref{f1_score_validation} presents the absolute performance difference on holdout data between classifiers trained on real and synthetic data. A difference closer to zero indicates similar classifier performance, reflected by an equivalent F1 score. Results show that NextConvGeN, TabDDPM, and CTAB-GAN perform comparably and slightly better than other models. Notably, all models except TVAE achieve zero absolute difference in lung cancer data. TabDDPM and NextConvGeN exhibit an absolute difference close to zero in three out of ten datasets, making them better models for downstream applications such as classification tasks.

\begin{table*}[htbp]\scriptsize\caption{The table contains propensity scores for various generative models across benchmarking datasets. The closer the propensity score to zero, the higher the utility, indicating that it is difficult for machine learning classifiers to differentiate between real and synthetic samples. The results show that NextConvGeN can consistently generate synthetic data which is similar to  real data compared.}\label{Propensity score}\centering\tabularnewline\vspace{5pt}\begin{tabular}{@{\hskip4pt}c@{\hskip4pt}|@{\hskip4pt}c@{\hskip4pt}|@{\hskip4pt}c@{\hskip4pt}|@{\hskip4pt}c@{\hskip4pt}|@{\hskip4pt}c@{\hskip4pt}|@{\hskip4pt}c@{\hskip4pt}|@{\hskip4pt}c@{\hskip4pt}}\hline
\textbf{dataset} & \textbf{TVAE} & \textbf{CTGAN} & \textbf{CTAB-GAN} & \textbf{CTAB-GAN+} & \textbf{TabDDPM} & \textbf{NextConvGeN}
\tabularnewline
\hline
\hline
Heart failure & 0.0416 & 0.0586 &  0.0716 & 0.0330 & 0.1041 & \textbf{0.0033} \tabularnewline
Heart disease & 0.0949 & 0.1000 &  0.1618 & 0.1010 &  \textbf{0.0080} & 0.0321 \tabularnewline
Lung cancer & NaN & 0.0649  & 0.0535  & 0.0615 & 0.0057 & \textbf{0.0049} \tabularnewline
Migraine & NaN & 0.0451  & 0.2123 & NaN & \textbf{0.0045} & 0.0048 \tabularnewline
Liver cirrhosis & NaN & 0.1083 & 0.0723 & 0.0830 & 0.9022 & \textbf{0.0041} \tabularnewline
Indian liver patients & 0.0836 & 0.1087 & 0.1158 & 0.0385 & 0.0052 & \textbf{0.0017} \tabularnewline
Pima Indian diabetes & 0.0689 & 0.0754  & 0.1105 & 0.0497 & 0.1360 & \textbf{0.0014} \tabularnewline
Contraceptive Methods & 0.0471 & 0.0268  & 0.1342 & 0.0106 & \textbf{0.0012} & 0.0045 \tabularnewline
Obesity & 0.0383 & 0.0486 & 0.1958 & NaN & 0.0014 & \textbf{0.0013} \tabularnewline
Stroke & NaN & 0.0185 & 0.1166 & 0.0079 & \textbf{0.0007} & 0.0021 \tabularnewline
\hline\end{tabular}\end{table*}

\textbf{The ability of CSL to generate synthetic data with high utility:} In synthetic data generation, the propensity score is a widely used utility measure \cite{dankar_fake_2021, drechsler_synthetic_2011, raab_guidelines_2017}. This score ranges from 0 to 0.25, where values closer to zero indicate higher similarity between real and synthetic data, while values near 0.25 suggest more significant differences \cite{dankar_multi-dimensional_2022}. Table \ref{Propensity score} reports the propensity scores across benchmark datasets, showing that NextConvGeN achieves values close to zero for most of the datasets. This indicates that machine learning classifiers struggle to distinguish between real and NextConvGeN-generated data, demonstrating the effectiveness of CSL in producing high-utility synthetic samples for small tabular clinical datasets.

\begin{table*}[htbp]\scriptsize\caption{The table shows the absolute cross-classification scores difference between real and synthetic data for various models across the benchmarking datasets. A smaller difference signifies a stronger correlation between the features preserved in synthetic data. The findings indicate that NextConvGeN excels in maintaining inter-feature relationships compared to other models.}\label{Cross classification}\centering\tabularnewline\vspace{5pt}\begin{tabular}{@{\hskip4pt}c@{\hskip4pt}|@{\hskip4pt}c@{\hskip4pt}|@{\hskip4pt}c@{\hskip4pt}|@{\hskip4pt}c@{\hskip4pt}|@{\hskip4pt}c@{\hskip4pt}|@{\hskip4pt}c@{\hskip4pt}|@{\hskip4pt}c@{\hskip4pt}}\hline
\textbf{dataset} & \textbf{TVAE} & \textbf{CTGAN} & \textbf{CTAB-GAN} & \textbf{CTAB-GAN+} & \textbf{TabDDPM} & \textbf{NextConvGeN}
\tabularnewline
\hline
\hline
Heart failure & 0.0667 & 0.1918  & 0.1518 & 0.1239 & 0.1107 & \textbf{0.0558} \tabularnewline
Heart disease & 0.0730  & 0.2168  & 0.2632  & 0.1754 & 0.1420 & \textbf{0.0644} \tabularnewline
Lung cancer & NaN & 0.0947  & 0.0914 & 0.1073 & \textbf{0.0205} & 0.0209\tabularnewline
Migraine & NaN & 0.4165  & 0.4475  & NaN & \textbf{0.0241} & 0.0388 \tabularnewline
Liver cirrhosis & NaN & 0.1121 & 0.1288 & 0.0778 & 0.0884 & \textbf{0.0455} \tabularnewline
Indian liver patients & 0.0453 & 0.1095 & 0.0753 & 0.0777 & 0.0462 & \textbf{0.0247} \tabularnewline
Pima Indian diabetes & 0.0373 & 0.1188 & 0.1335 & 0.1314 & 0.0562 & \textbf{0.0364} \tabularnewline
Contraceptive methods & 0.0238 & 0.0906 & 0.1118 & 0.0835 & \textbf{0.0207} & 0.0261 \tabularnewline
Obesity & 0.3465 & 0.7695 & 0.7863  & NaN & \textbf{0.0301} & 0.0380 \tabularnewline
Stroke & NaN & \textbf{0.0059} & 0.0207  & 0.0075 & \textbf{0.0059} & 0.0063 \tabularnewline
\hline\end{tabular}\end{table*}

Table \ref{Cross classification} presents the absolute cross-classification difference for various generative models across ten benchmarking datasets. The cross-classification metric assesses statistical dependencies for a variable based on other variables using a classifier. Results show that NextConvGeN and TabDDPM achieve the best performance, outperforming other models in five out of ten datasets. Both models surpass GAN-based models and TVAE. Notably, NextConvGeN consistently yields small differences close to zero, likely due to its unique approach of treating categorical variable categories as real values, unlike other models that rely on one-hot encoding.

Overall, NextConvGeN and TabDDPM exhibit comparable performance in utility measures for small tabular clinical datasets. Notably, NextConvGeN outperformed other generative models in the commonly used evaluation metric, propensity score, indicating its ability to generate synthetic data that is highly similar to real data. The following section examines the privacy risks associated with different generative models. Since privacy is assessed across an average of ten publicly available clinical datasets, TVAE and CTAB-GAN+ are excluded from the comparison, as they failed to maintain the cardinality of the synthetic target column in some datasets.
\begin{figure}[htbp]
    \centering
    \includegraphics[width=0.3\textwidth]{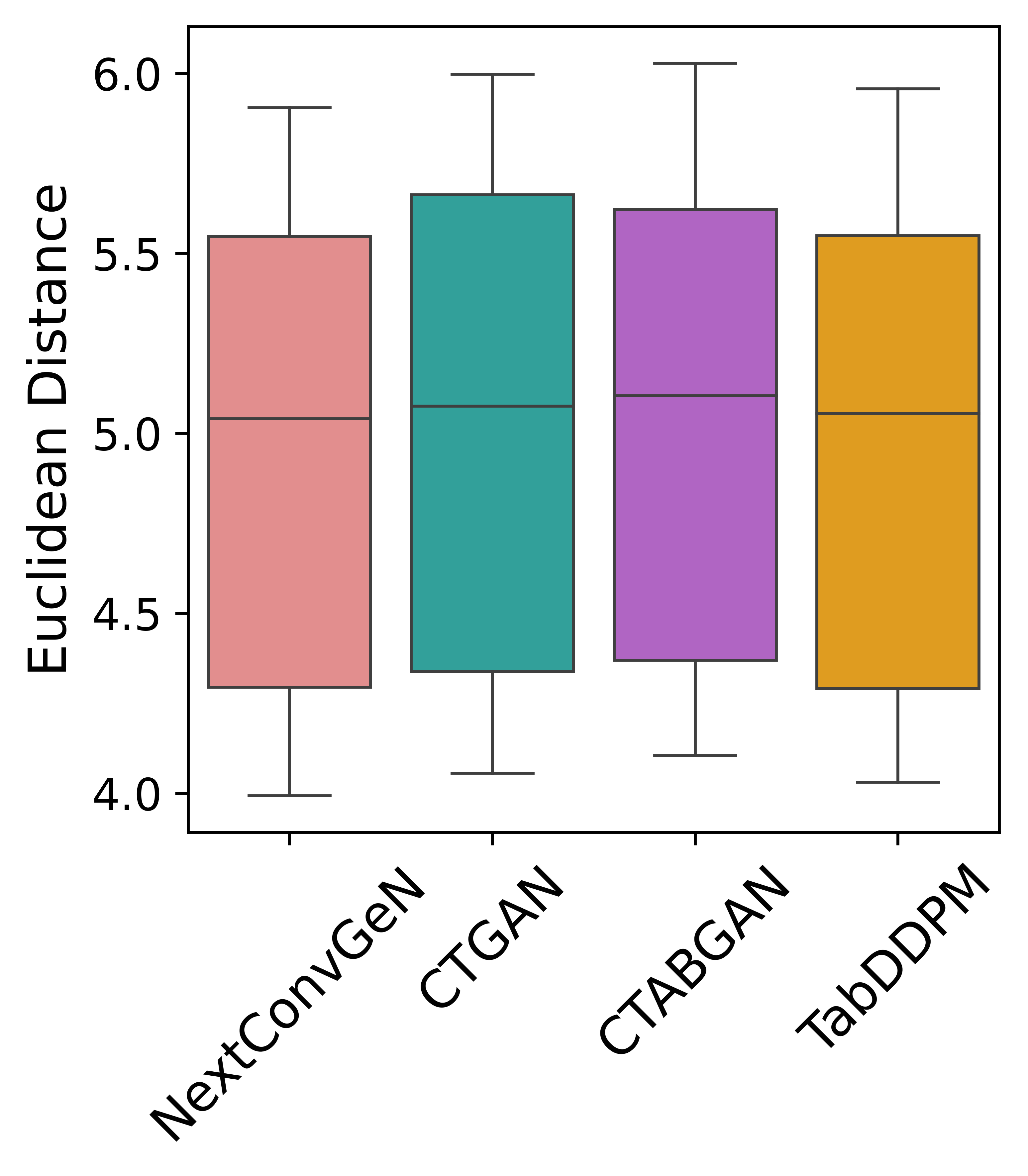}
    \includegraphics[width=0.3\textwidth]{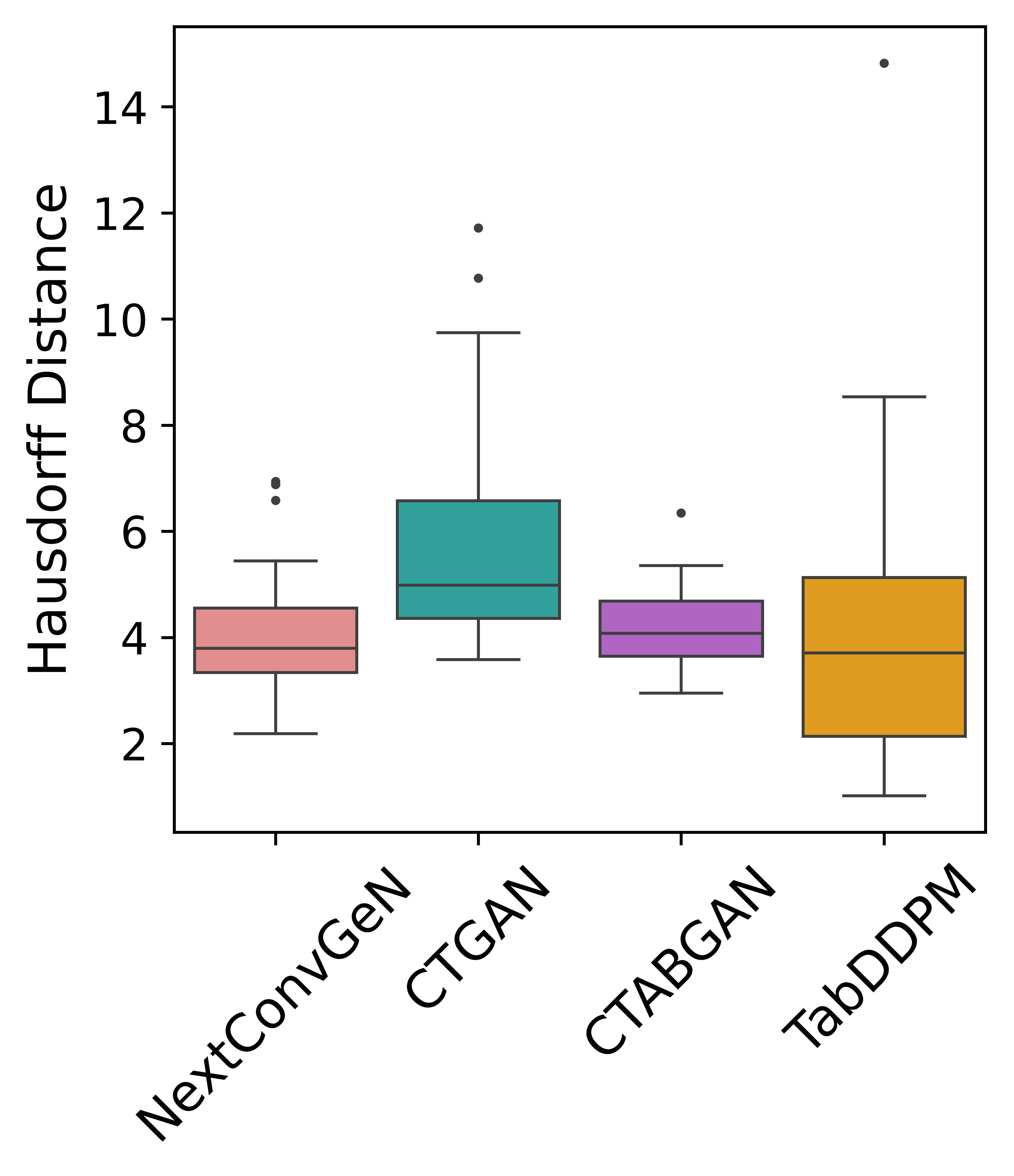}
    \includegraphics[width=0.3\textwidth]{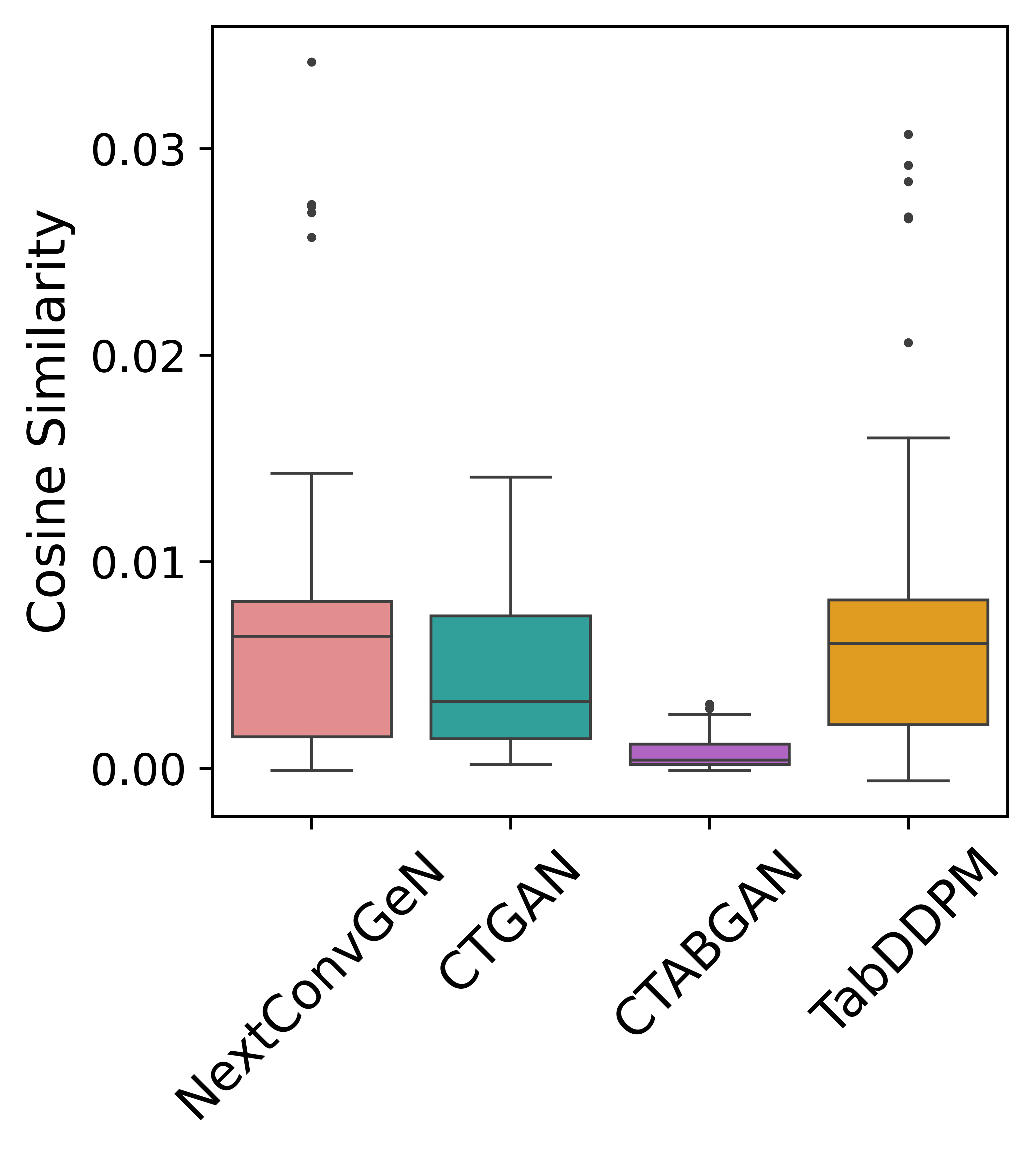}
    \vspace{-8pt}\caption{Left Plot (Euclidean Distance): Box plot illustrating the average Euclidean distance between real and synthetic datasets. Lower values indicate better privacy preservation. The similarity across models suggests comparable privacy risk.
    Centre Plot (Hausdorff Distance): Box plot depicting the Hausdorff distance between real and synthetic datasets. Higher values indicate better privacy preservation. CTGAN exhibits values in a higher range, implying lower privacy risk compared to other models.
    Right Plot (Cosine Similarity): Box plot displaying the cosine similarity between real and synthetic datasets. Lower values indicate better privacy preservation. CTAB-GAN consistently demonstrates lower cosine distances, implying lower privacy risk in terms of distance.}\label{distance measure plots}%
\end{figure}

\textbf{GAN-based models exhibit low risk of privacy breach in distance measure evaluation: }We assessed synthetic data privacy using similarity measures: Euclidean distance, Hausdorff distance, and Cosine similarity, computed between real and synthetic data across ten benchmarking datasets. Each generative model underwent five training rounds with different random seeds, yielding fifty distance values per model, visually represented through box plots (Figure \ref{distance measure plots}).

The Euclidean distance (higher is better) showed relatively uniform privacy risks across models. The Hausdorff distance (higher is better) indicated lower privacy risks for CTGAN due to greater real-to-synthetic sample distances, while other models exhibited smaller distances, raising privacy concerns. Regarding Cosine similarity (lower is better), CTAB-GAN demonstrated the lowest values, suggesting strong privacy protection, whereas NextConvGeN and TabDDPM displayed broader distributions with extreme values, implying potential privacy risks. Overall, GAN-based models demonstrate better privacy preservation. The reasons for this will be explored further in Section \ref{discussion}.
\begin{comment}
\begin{figure}[htbp]
    \centering
    \includegraphics[width=0.32\textwidth]{CTGAN.png}
    \includegraphics[width=0.32\textwidth]{CTABGAN.png}
    \includegraphics[width=0.32\textwidth]{TabDDPM.png}
    \includegraphics[width=0.32\textwidth]{NextConvGeN.png}
    \caption{The line plots above depict precision values for membership inference attacks against generative models. Each plot represents precision on the y-axis and the proportion of data available to the attacker on the x-axis, with different colored lines representing various thresholds. CTGAN, CTAB-GAN, and TabDDPM consistently exhibit precision values below $0.5$ across different access proportions and thresholds, indicating a low risk of reidentification. Conversely, NextConvGeN's plot shows precision values around $0.6$ across varying access levels and thresholds, suggesting an increased risk of reidentification.}
  \label{MIA}
\end{figure}
\end{comment}
\begin{figure}[htbp]
  \centering
  \subfigure{
    \includegraphics[width=0.32\textwidth]{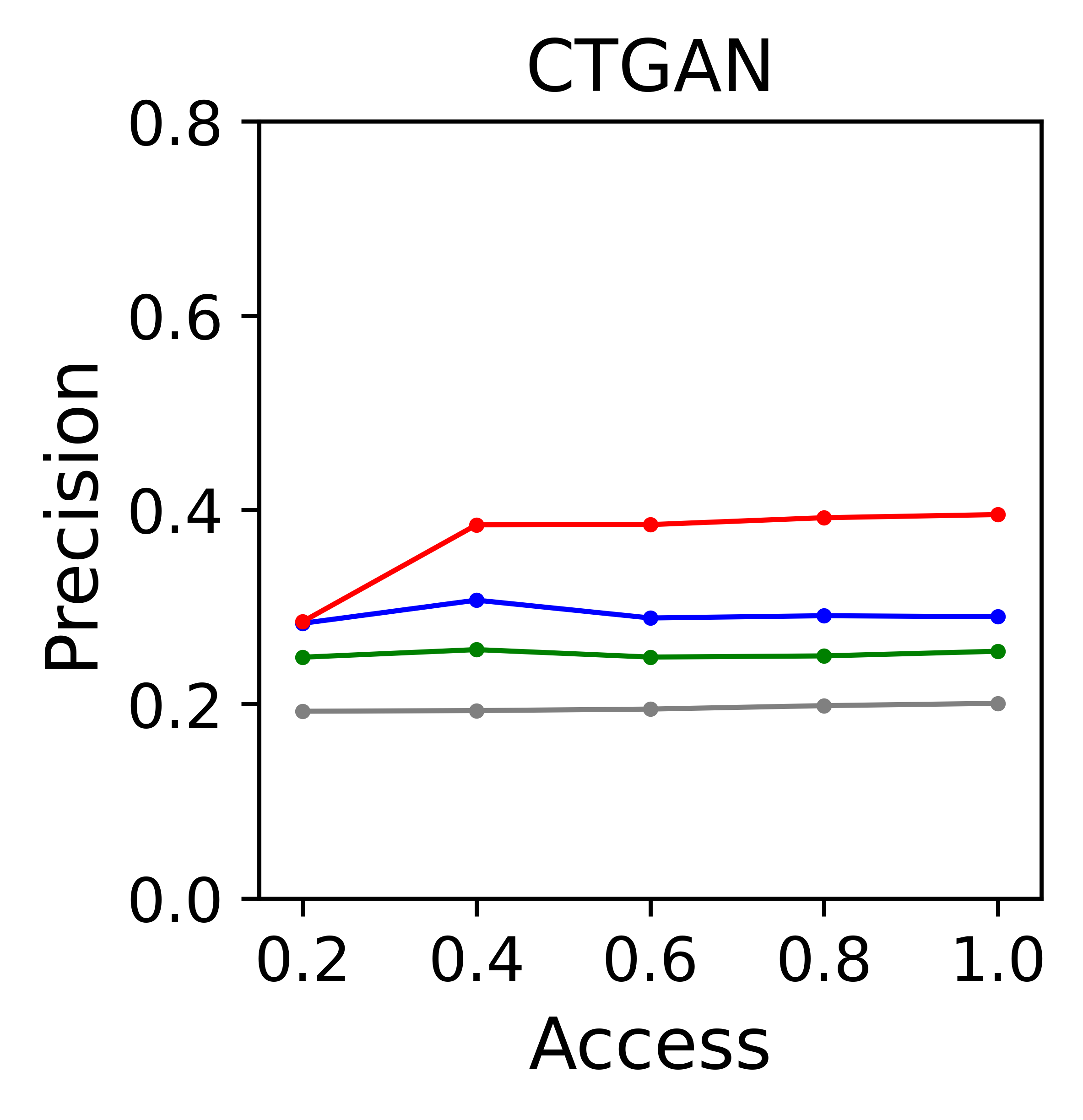}
  }
  \subfigure{
    \includegraphics[width=0.32\textwidth]{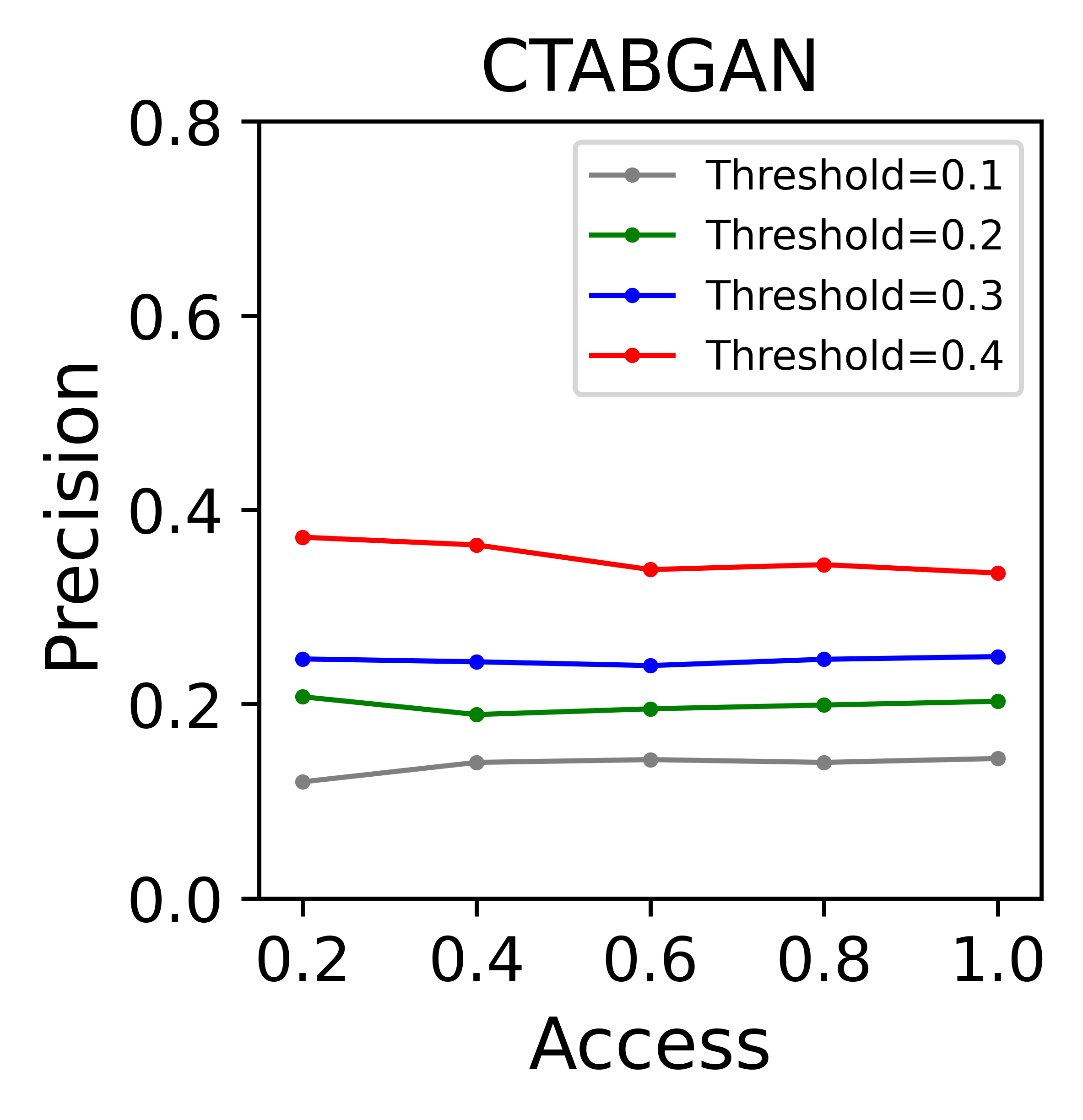}
  }
  \subfigure{
    \includegraphics[width=0.32\textwidth]{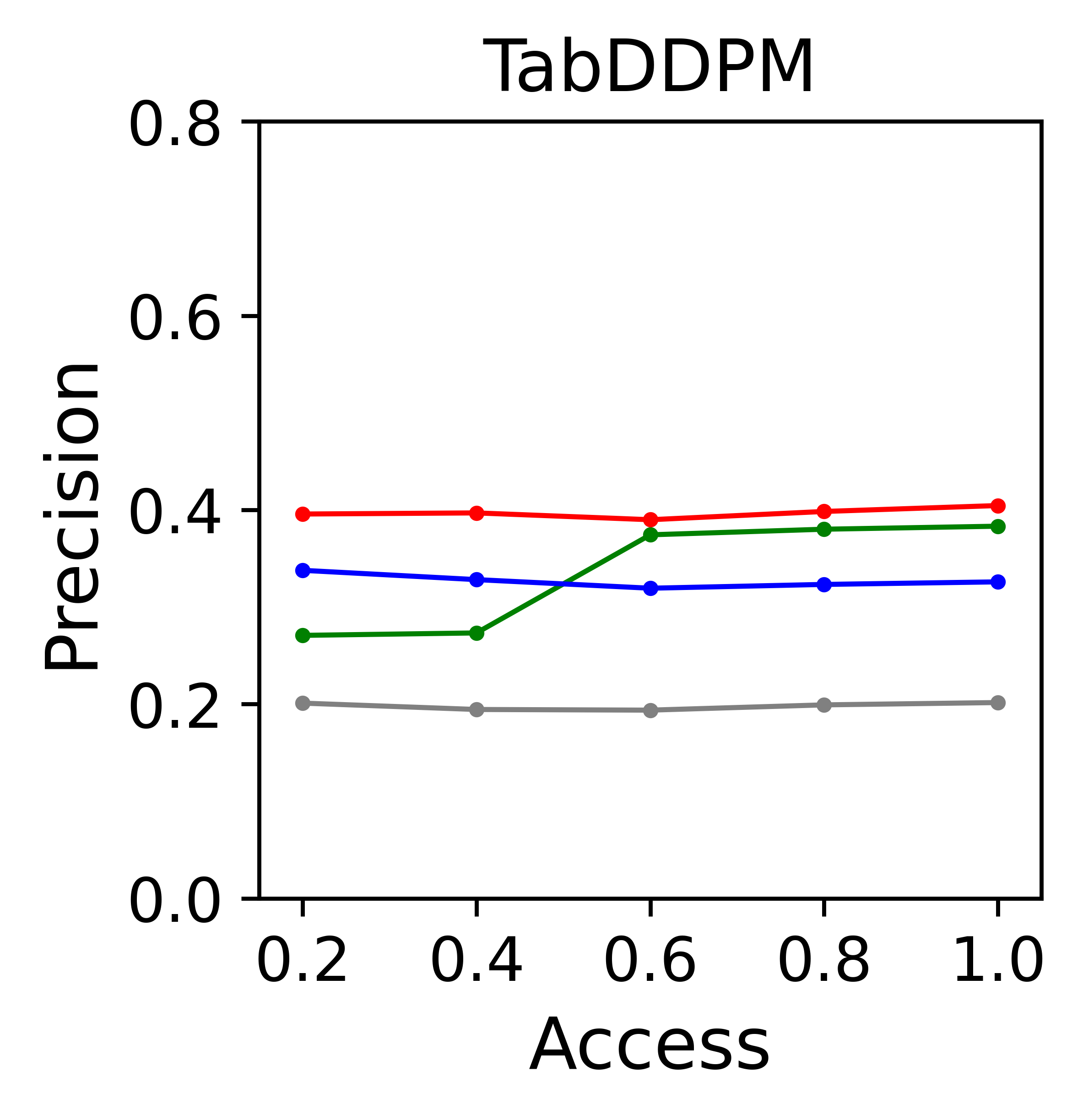}
  }
  \subfigure{
    \includegraphics[width=0.32\textwidth]{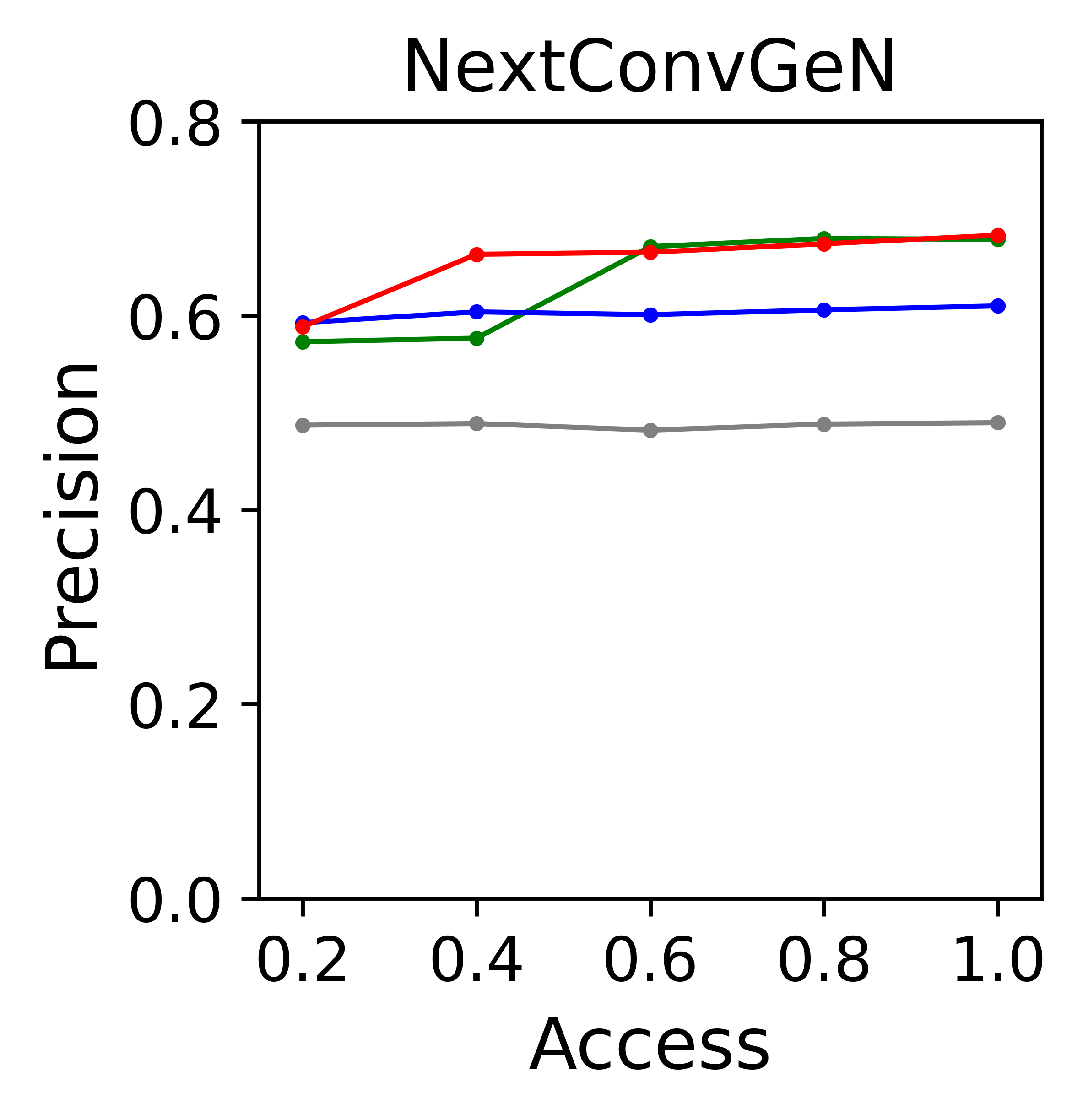}
  }
  \vspace{-8pt}\caption{The line plots above depict precision values for membership inference attacks against generative models. Each plot represents precision on the y-axis and the proportion of data available to the attacker on the x-axis, with different colored lines representing various thresholds. CTGAN, CTAB-GAN, and TabDDPM consistently exhibit precision values below $0.5$ across different access proportions and thresholds, indicating a low risk of reidentification. Conversely, NextConvGeN's plot shows precision values around $0.6$ across varying access levels and thresholds, suggesting an increased risk of reidentification.}
  \label{MIA}
\end{figure} 

\textbf{GAN-based models are less sensitive to MIA and AIA: }Figure \ref{MIA} illustrates the average precision score for MIA for the generative models in the benchmarking study. The lower the precision score, the lower the risk of privacy breach. While CTGAN, CTAB-GAN, and TabDDPM consistently exhibit precision scores below $0.5$ across different thresholds and access levels, NextConvGeN achieves a precision score of around $0.6$. Thus, NextConvGeN performs poorly in terms of MIA.

\begin{figure}[htbp]
  \centering
    \includegraphics[width=0.3\textwidth]{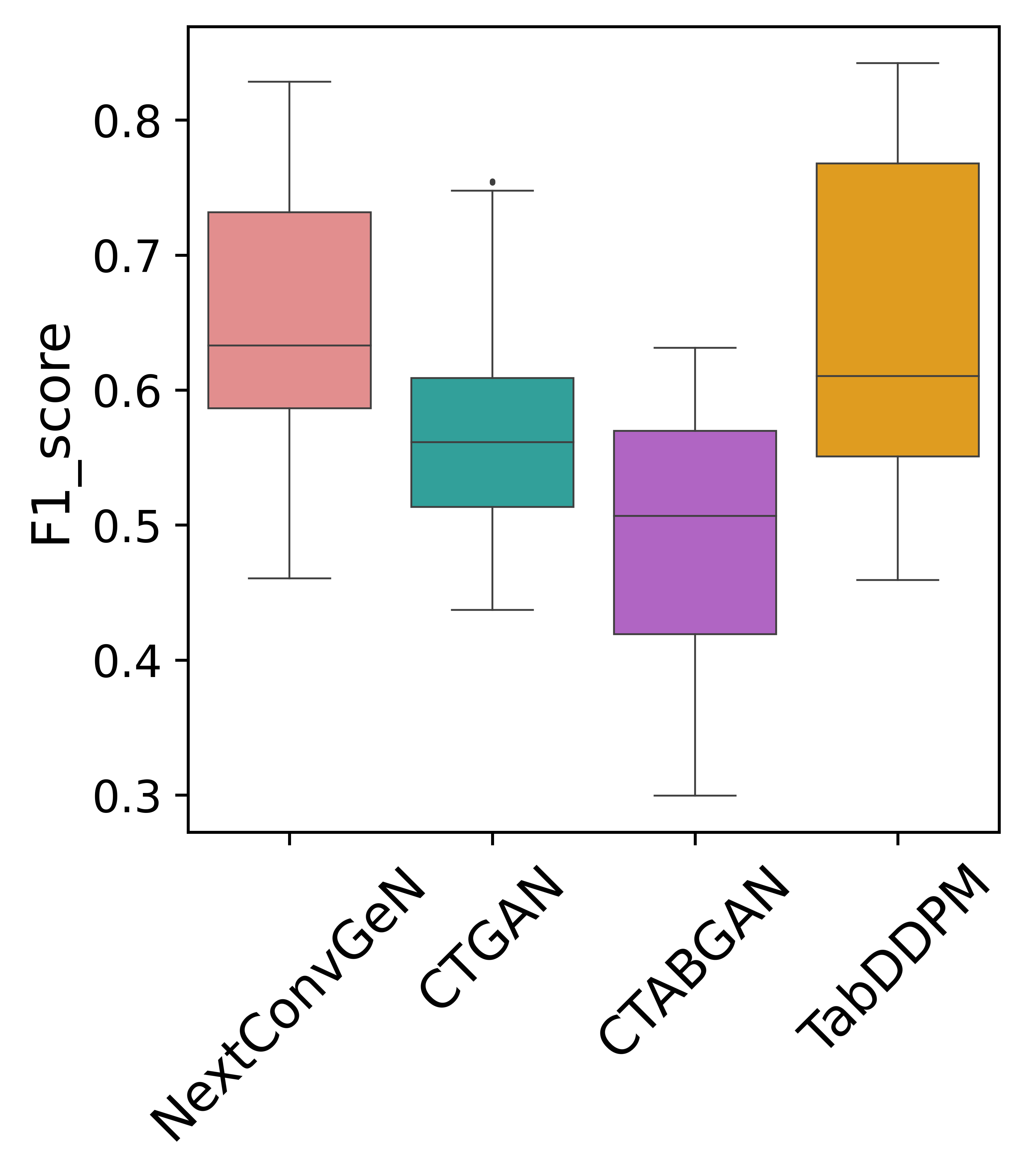}
    \includegraphics[width=0.3\textwidth]{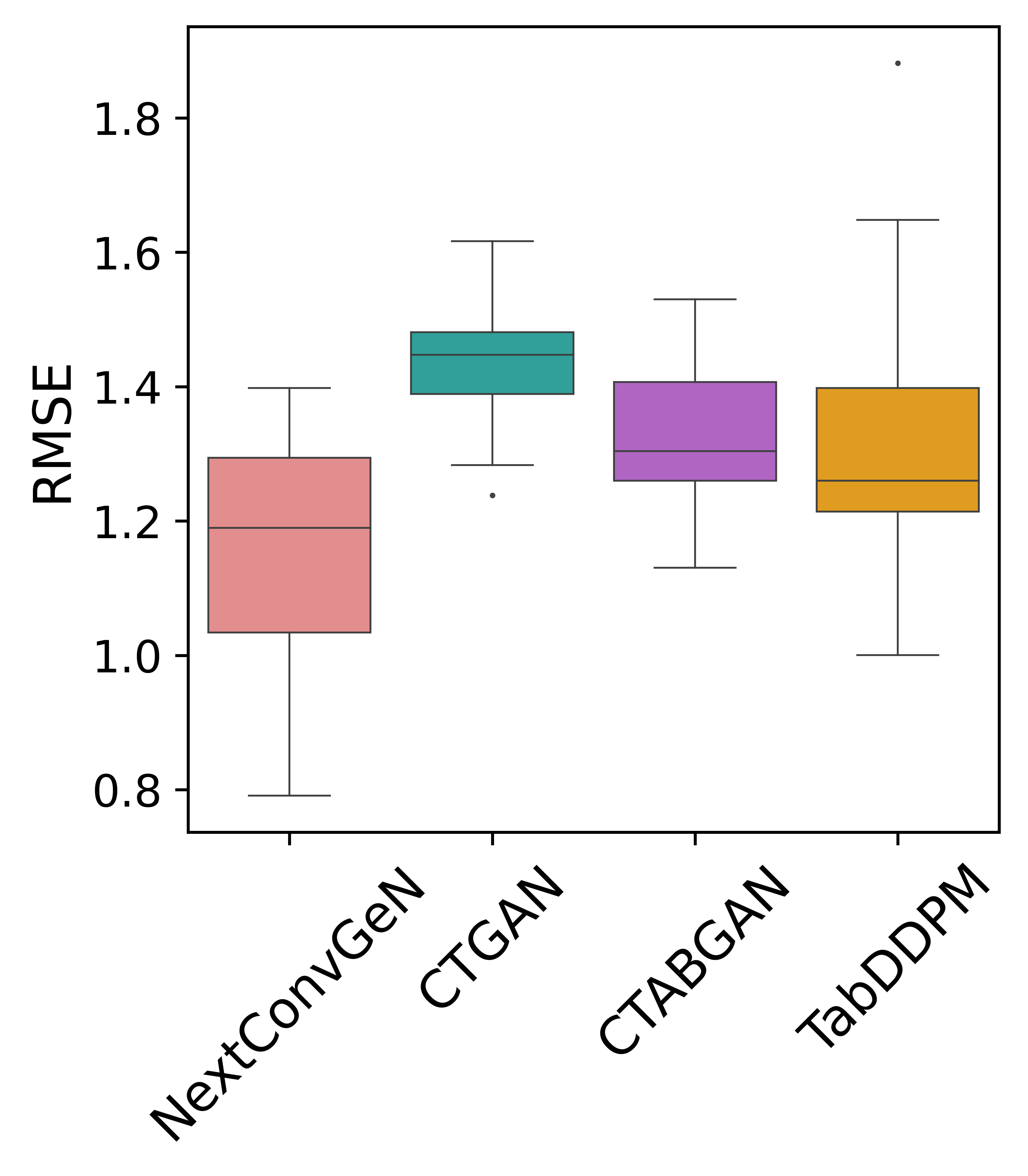}
    \vspace{-8pt}\caption{The box plots above showcase the results of the Attribute Inference Attack across different models. For categorical features, F1-scores were calculated, where lower values indicate a lower reidentification risk (left box plot). Root Mean Square (RMS) values were computed for continuous features, with larger values suggesting a lower reidentification risk (right box plot). The plots reveal that CTGAN and CTAB-GAN consistently exhibit lower F1-scores and higher RMS values distributions, indicating a lower reidentification risk compared to TabDDPM and NextConvGeN.}\label{F1_AIA}
\end{figure}

The results of the Attribute Inference Attack were analyzed using box plots to visualize the F1-score for predicting sensitive categorical features and the RMS value for predicting sensitive continuous features (Figure \ref{F1_AIA}). A lower F1-score indicates greater difficulty in predicting unknown categorical variables, while a higher RMS value suggests greater difficulty in estimating missing continuous sensitive variables. Analysis showed that CTGAN and CTAB-GAN struggled to predict sensitive features when trained on quasi-identifiers. This led to a higher RMS range for continuous features and a lower F1-score range for categorical features, indicating a lower risk of individual patient reidentification. In contrast, NextConvGeN and TabDDPM exhibited a higher risk of attribute inference.

Standard deviations for all evaluation measures across datasets over five runs are available in our GitHub repository. For brevity, we report standard deviations only for the Heart Disease dataset in Table \ref{std_dev_table}.

\begin{table*}[!ht]
\footnotesize
\caption{Standard deviations of evaluation measures for the Heart Disease dataset over five experimental runs. The result demonstrates the consistency of the results for key metrics across runs, highlighting the stability of the generated synthetic data. Detailed standard deviations for all other datasets are provided in the GitHub repository.}
\begin{center}
\begin{tabular}{c|c|c|c|c} 
\hline
{\textbf{Evaluation Measure}} & {\textbf{CTABGAN}} & {\textbf{CTGAN}} & {\textbf{NextConvGeN}} & {\textbf{TabDDPM}} \\ 
\hline
Student-T-test & 0.0479 & 0.0016 & 0.0270 & 0.2058 \\
KL divergence  & 0.0073 & 0.0057 & 0.0042 & 0.0020 \\
Propensity score & 0.0054 & 0.0392 & 0.0079 & 0.0056 \\
Log Cluster Metric & 0.4670 & 1.4701 & 0.0055 & 6.0218 \\
Cross-validation & 0.0234 & 0.0464 & 0.0150 & 0.0194 \\
F-1 score & 0.1851 & 0.0467 & 0.1309 & 0.1149 \\
Cross Classification & 0.0332 & 0.0364 & 0.0440 & 0.0060 \\
Euclidean distance & 0.0011 & 0.0053 & 0.0114 & 0.0049 \\
Hausdorff distance & 0.1201 & 0.6071 & 0.2686 & 0.7164 \\
\hline
\end{tabular}
\label{std_dev_table}
\end{center}
\end{table*}
\vspace{-5pt}
In addition to the evaluation measures discussed earlier, we further compared the models by extracting two principal components from the real data and the corresponding synthetic data generated by the tabular data generative models. We conducted a two-dimensional Kolmogorov-Smirnov (KS) test (also known as the Peacock test) to assess the statistical similarity between the distributions. We visualized the results with p-values displayed on each plot. These visualizations are provided in our GitHub repository. The results indicate that, for most datasets generated by NextConvGeN, the p-values exceeded $0.05$, demonstrating that synthetic data generated using CSL effectively preserves the data distribution in a low-dimensional space.

To sum up, we observe that the NextConvGeN and the TabDDPM models generate more realistic synthetic data. The synthetic data generated from such models can better mimic the real data regarding utility-based tasks. For example, if synthetic data from NextConvGeN is used for a classification task, then the classification performance would be relatively similar to the situation if the real data were used for the same classification task. Additionally, NextConvGeN preserves better fidelity of synthetic data and demonstrates better performance on downstream tasks such as ML classification. On the other hand, we notice that CTGAN and CTAB-GAN perform better in terms of privacy preservation.

\section{Discussion}\label{discussion}
\begin{figure}[htp]
\centering
\includegraphics[width=.3\textwidth]{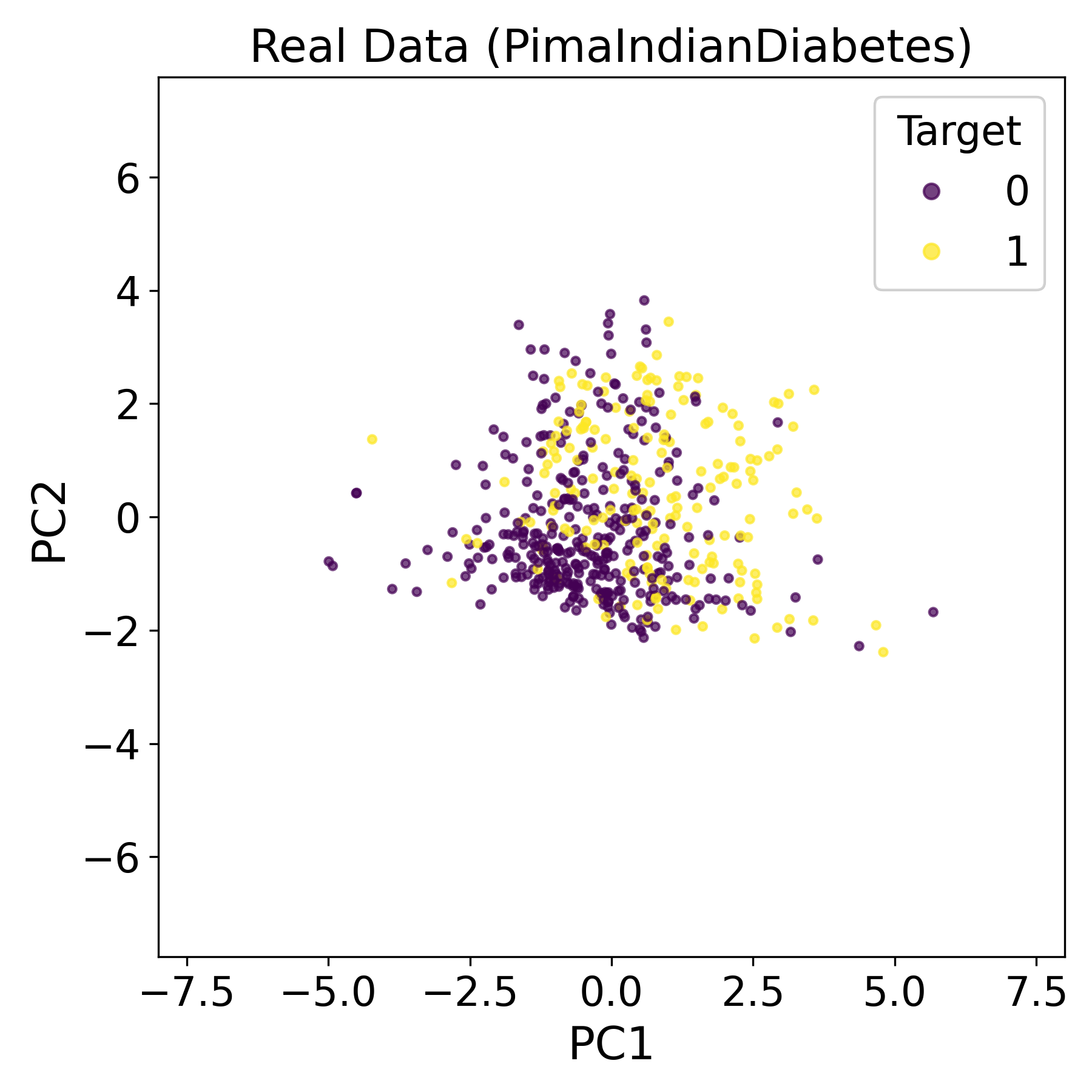}\quad
\includegraphics[width=.3\textwidth]{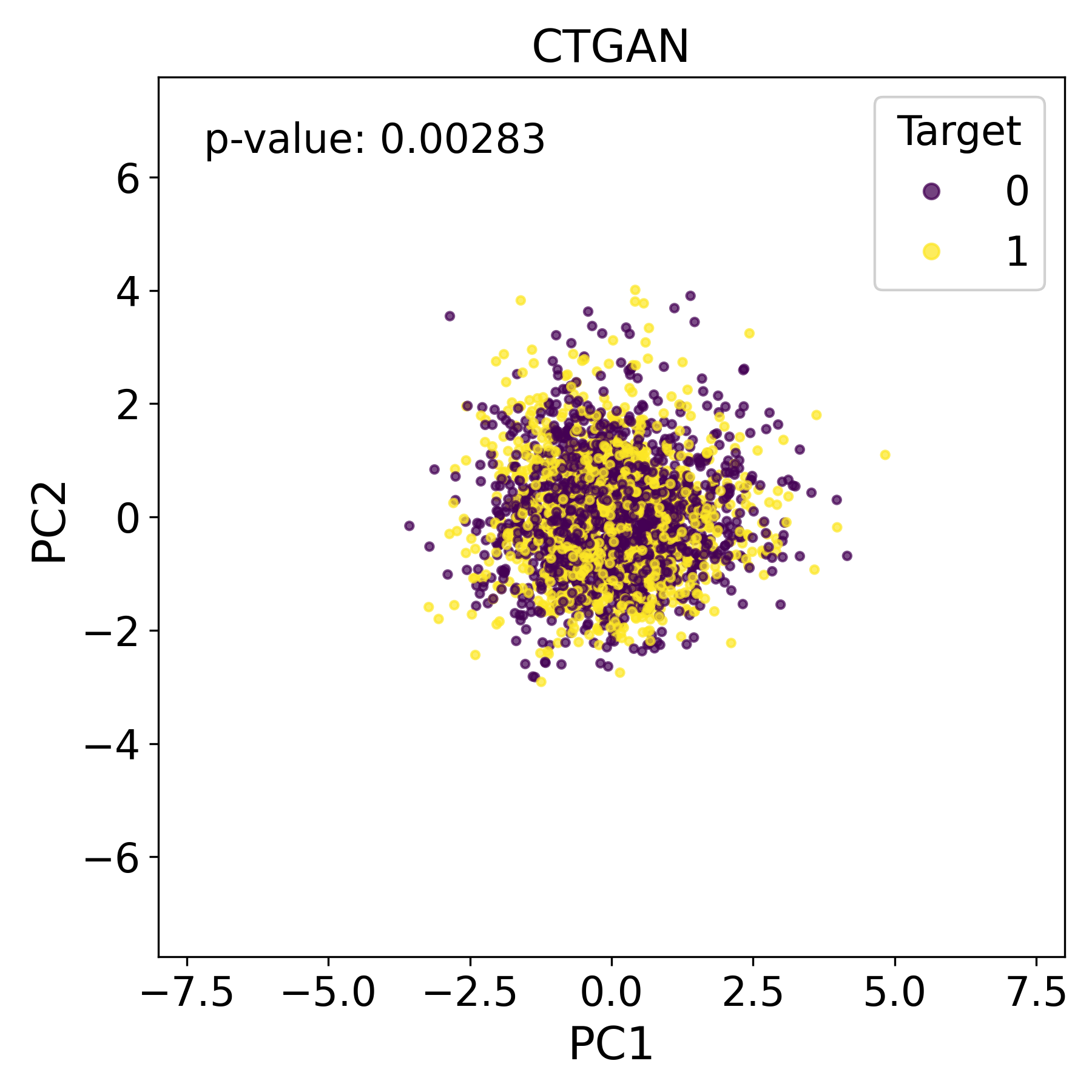}\quad
\includegraphics[width=.3\textwidth]{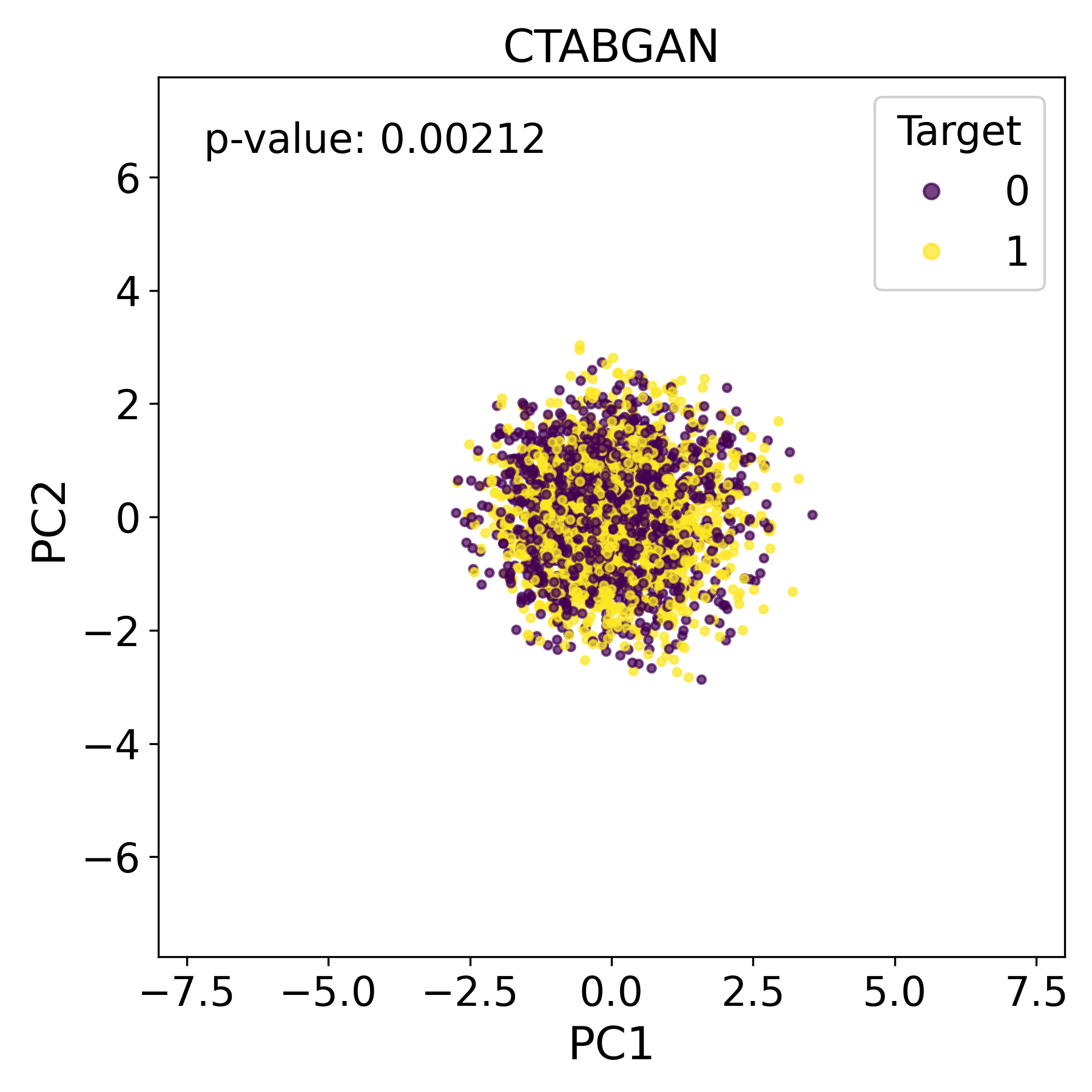}
\medskip
\includegraphics[width=.3\textwidth]{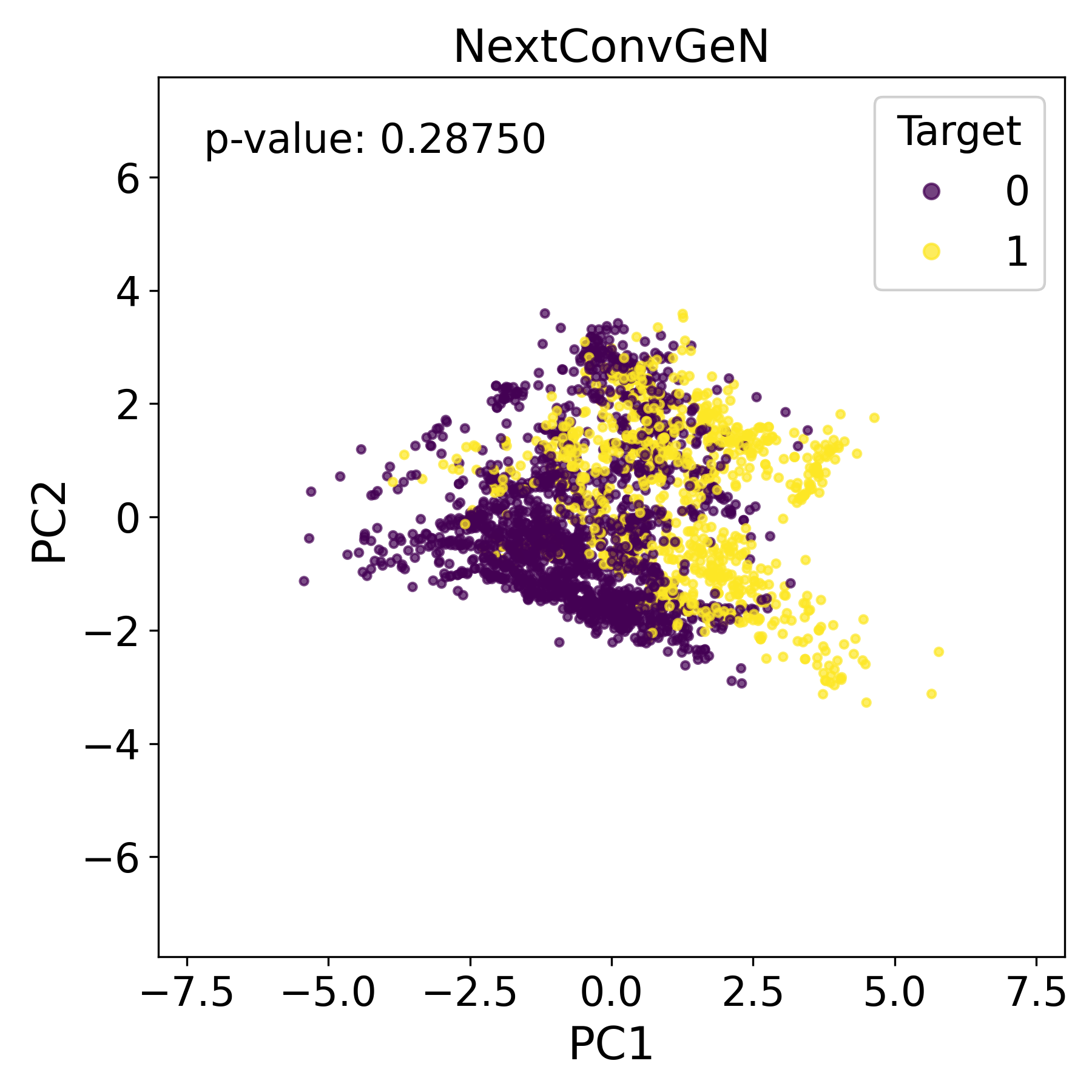}\quad
\includegraphics[width=.3\textwidth]{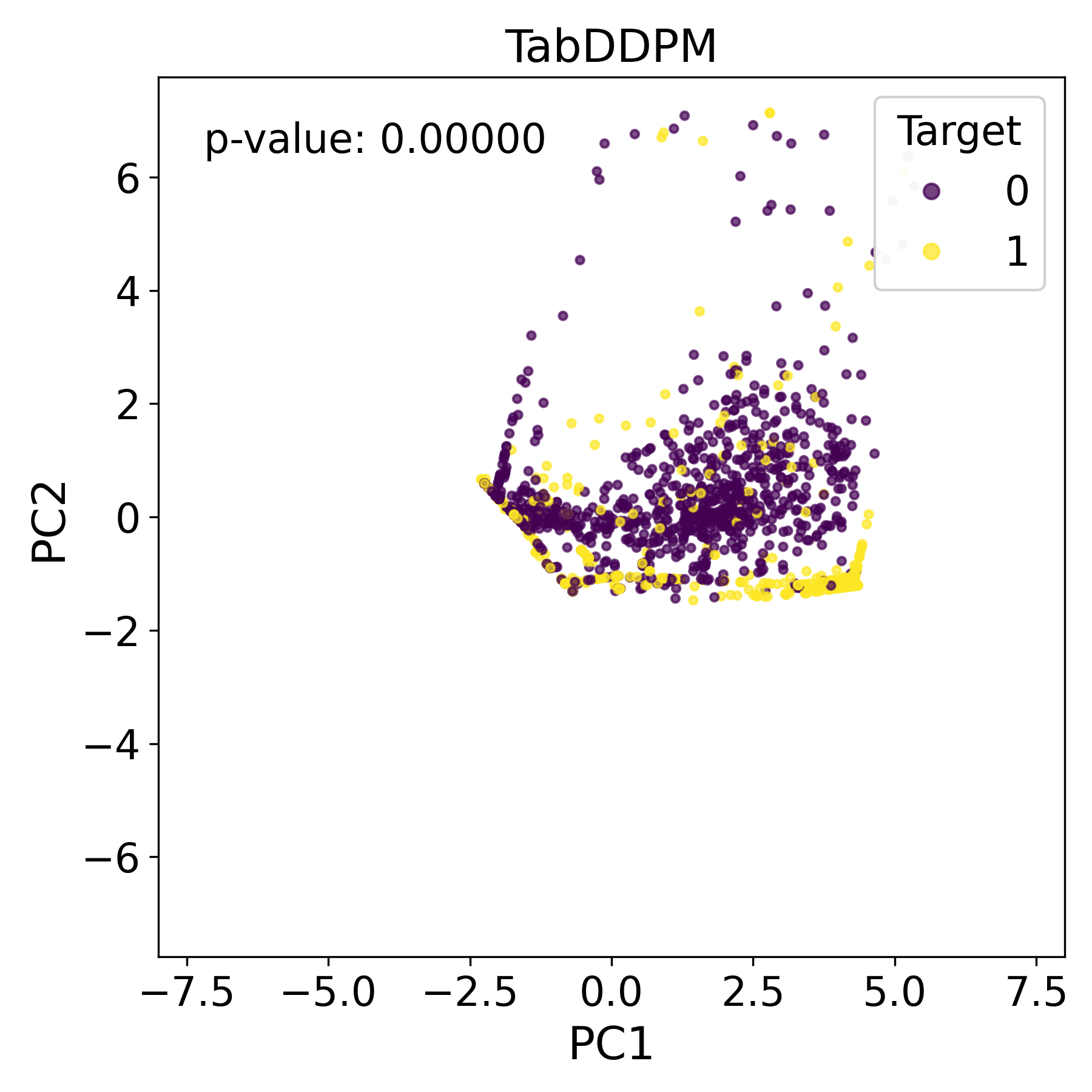}
\vspace{-8pt}\caption{PCA visualization of the first two principal components for Real data and synthetic data generated using CTGAN, CTABGAN, NextConvGeN, and TabDDPM models on the Pima Indian Diabetes dataset. The plot demonstrates that the synthetic data produced by the NextConvGeN model closely resembles the distribution of the real data, outperforming the other generative models in preserving the data structure.}
\label{pca plots}
\end{figure}

\textbf{Why use CSL for synthetic tabular data generation?:}
From our results in Section \ref{results}, it is clear that the ConvGeN model performs well in preserving utility measures across synthetic and real data. For propensity score, which is one of the most established utility metrics, NextConvGeN outperforms all other models. For measures like log cluster metric absolute cross-validation score differences and absolute F1-Score differences, NextConvGeN is also at par with the TabDDPM model. From the qualitative visualization of synthetic data generated by several models in Figure \ref{pca plots}, it is also clear that NextConvGeN preserves the distribution of the data quite well.

In addition to these, certain theoretical aspects make the model special. Let $x=(x_1,\dots,x_{f})$ be an arbitrary data point in a dataset $X$ with $f$ features. Let $N_k(x)$ be the set of the $k$-nearest neighbors of $x$ in a dataset $X$. If we define a synthetic sample as $s=\sum_{i=1}^k\alpha_k p_k$, where $p_k\in N_k(x)$ and $\sum_{i=1}^k\alpha_k=1$, if the random variables for corresponding column attributes $X_1,\dots,X_{f}$ follow some distribution with mean $\mu=(\mu_1,\dots,\mu_k)$ and standard deviation $\sigma=(\sigma_1,\dots,\sigma_k)$ in the neighborhood $N^k(x)$, then,
$$
E[s^j]=\sum_{i=1}^kE[\alpha_i]E[p^j_i]=\sum_{i=1}^k\frac{1}{k}\mu_j=\mu_j
$$
meaning that the $j$-th component of the synthetic samples generated in that neighborhood follows a distribution with the same mean as the distribution followed by the $j$-th component of a real sample in that neighborhood. Of course, the $\alpha$s are also considered as random variables here following a Dirichlet distribution.  Similarly,
\begin{equation}
\begin{split}
    \text{Var}(s^j) &= \text{Var}\bigg(\sum_{i=1}^k \alpha_ip^j_i\bigg)\\
    & = \sum_{i=1}^k \text{Var}(\alpha_ip^j_i) + \sum_{m=1}^{k}\sum_{l=1,l \neq m}^{k}
    \text{Cov}\bigg(\alpha_mp^j_l,\alpha_lp^m_j\bigg)\\
    & = \frac{2\times(\sigma^{j})^2}{k+1} 
\end{split}
\end{equation}
following the calculations from Bej \textit{et al.}\cite{bej_loras_2021}. This implies that synthetic samples generated by CSL have the same local mean as the original samples. The proposed NextConvGeN network can learn an appropriate data-distribution-specific variance. This ensures that the local distribution of the synthetic data closely follows that of the real data, which is also reflected in our qualitative visualization of the synthetic samples. This theoretical framework adds explainability to the synthetic samples generated, which is something special about the proposed model.

\textbf{Ablation study: }To evaluate the effectiveness of the proposed NextConvGeN model, we performed an ablation study focusing on the impact of the \var{fdc} and \var{alpha\_clip} parameters. By generating synthetic datasets with and without the \var{fdc} parameter and comparing their reduced-dimensional representations with the corresponding real data, we observed that including the \var{fdc} parameter significantly improves the alignment of synthetic and real data distributions, particularly for datasets with diverse feature types. These results were validated using the 2D Kolmogorov–Smirnov test and visual comparisons of the data distributions. Additional details, including figures and results, are provided in the appendices.

We conducted an ablation study to assess the effect of the \var{alpha\_clip} parameter in the NextConvGeN model, designed to minimize the generation of real samples as synthetic samples. The study compared the exact match scores (percentage of exact matches between real and synthetic samples) for various generative models, including NextConvGeN with and without the \var{alpha\_clip} parameter. A detailed explanation of the experiment, along with the results table, is provided in the appendices. In summary, the findings demonstrate that enabling the \var{alpha\_clip} parameter significantly reduces the exact match score, emphasizing its importance in controlling the resemblance between synthetic and real data.

\textbf{Utility-privacy trade-off in synthetic data generation:} Our results highlight the inherent dilemma of designing synthetic data generation strategies that balance privacy preservation and utility. High utility measures generally indicate that the statistical distributions of real and synthetic data are similar (Figure \ref{pca plots}), which in turn implies comparable performance across real and synthetic data in downstream machine learning tasks like clustering or classification. However, achieving high utility often compromises privacy. When synthetic data closely resembles real data distributions, the risk of reidentification increases as the synthetic data becomes easier to reverse-engineer back to the original data.
In Table \ref{Cross validation}, the absolute five-fold cross-validation performance difference between gradient boosting classifiers trained on real and synthetic data is notable. For CTGAN and CTAB-GAN, this difference exceeds 10\% for more than half of the datasets, reflecting suboptimal performance in practical scenarios. This can be attributed to the fact that GAN-based models outperform other approaches in privacy evaluation, as they do not preserve qualitative plots (refer to Figure \ref{pca plots}). The distribution of synthetic data generated by these models deviates significantly from the corresponding real data, which enhances privacy performance but compromises utility. Conversely, models like TabDDPM and NextConvGeN exhibit higher utility but underperform in privacy measures. This outcome underscores the utility-privacy trade-off and emphasizes the need for models that strike a balanced compromise.

\textbf{Adversarial learning vs. direct data utilization:} GAN-based models such as CTGAN and CTAB-GAN rely on adversarial learning, where the generator learns to map a random noise distribution to the real data distribution. Hence, the lower-dimensional plots of synthetic data generated by these models are sparse in Figure \ref{pca plots}, deviating significantly from their corresponding real distribution. A trained adversarial model of such nature does not require real data during synthetic sample generation is advantageous for privacy preservation on the one hand but requires large training datasets to achieve satisfactory utility on the other hand. 

In contrast, models like TabDDPM and NextConvGeN utilize real data points directly.  For instance, NextConvGeN generates synthetic samples by leveraging neighborhoods sampled from real data, allowing it to learn the distribution of the data space. As a result, the synthetic data distribution more closely resembles its corresponding real distribution, as demonstrated in Figure \ref{pca plots}. This enables higher utility but increases privacy risks. These differences highlight that GAN models are less suited for small, domain-specific datasets, such as clinical data from single health facilities, due to their reliance on extensive training data.

\textbf{Mode collapse and imbalanced datasets:} The evaluation of TVAE and CTAB-GAN+ highlights the persistent issue of mode collapse, where generative models fail to capture the diversity of the data. Despite architectural enhancements in CTAB-GAN+, such as training by sampling and packing, performance on small clinical datasets remains suboptimal. Limited training data exacerbates mode collapse, particularly for minority class samples. In TVAE, latent space compression during encoding may impair the decoder's ability to reconstruct all target classes accurately. Conversely, NextConvGeN mitigates mode collapse by training generators within minority sample neighborhoods, ensuring a more balanced representation. These findings suggest that TVAE and CTAB-GAN+ are ill-suited for small or imbalanced datasets.

Below, we summarize the following observations from our study that hint towards some research gaps:
\renewcommand{\baselinestretch}{1.15}\normalsize
\begin{enumerate}
    \item \textbf{Objective performance assessment:}
    There is a lack of clarity on what constitutes acceptable utility for synthetic data generation models. Objective benchmarks and thresholds must be established, especially for evaluating the utility-privacy trade-off. Developing standardized metrics for assessing the balance between utility and privacy is essential for fair comparison across models.
   \item \textbf{Task-specific synthetic data generation:}
        Synthetic data serves diverse purposes, including mimicking classification or clustering tasks, circumventing privacy issues, or maintaining logical dependencies. A single model is unlikely to excel across all tasks. Task-specific synthetic data generation models tailored to particular use cases should be explored.
   \item \textbf{Hybrid approaches:}
        Current models employ different philosophies, such as adversarial learning, diffusion models, and CSL. For example, diffusion and convex space models perform well for utility, whereas adversarial approaches prioritize privacy. A hybrid approach combining these philosophies could offer a balanced performance in terms of utility and privacy.
\end{enumerate}

\section{Conclusion}
The most important aspect of our work is that we introduced a completely new philosophy of synthetic data generation apart from known approaches like adversarial learning, diffusion models, autoencoders, Bayesian approaches, etc. Qualitatively and quantitatively we observe that our model is a high-utility model, comparable to the state-of-the-art TabDDPM model. As discussed in Section \ref{discussion}, it has some interesting theoretical aspects as well since it locally preserves the statistical properties of the original data.

Despite these advancements, several challenges remain. Measuring and balancing the trade-off between utility and privacy remains an open problem. Future research can also prioritize the development of standardized metrics to objectively evaluate generative model performance and define acceptable utility thresholds. Exploring task-specific generative approaches tailored to applications such as classification, clustering, or logical dependency preservation will address diverse user needs. 

\section*{CRediT authorship contribution statement}
\textbf{Manjunath Mahendra:} Writing – original draft,
Methodology, Validation. \textbf{Chaithra Umesh:} Writing – review \& editing, Visualisation. \textbf{Saptarshi Bej:} Writing – review \& editing, Conceptualization, Supervision. \textbf{Kristian Schultz:} Writing – review \& editing, Algorithm optimization, \textbf{Olaf Wolkenhauer:} Writing – review \& editing, Supervision.
\section*{Conflict of Interest}
The authors have no conflict of interest.
\section*{Availability of code and results}
To support transparency, re-usability, and reproducibility, we provided detailed Jupyter notebooks from our experiments in \href{https://github.com/manjunath-mahendra/NextConvGeN}{GitHub}.
\section*{Acknowledgment}
This work has been supported by the German Research Foundation (DFG), FK 515800538, obtained for `Learning convex data spaces for generating synthetic clinical tabular data'.

\bibliographystyle{elsarticle-num}
\renewcommand{\baselinestretch}{1.15}\normalsize
\bibliography{NextConvGeN}
\newpage    
\renewcommand{\thetable}{\Alph{table}} % Change table numbering to A, B, C
\setcounter{table}{0} % Reset table counter
\renewcommand{\thefigure}{\Alph{figure}} % Change figure numbering to A, B, C
\setcounter{figure}{0} % Reset figure counter
\setcounter{section}{0}  % Reset section counter
\renewcommand{\thesection}{\Alph{section}} % Change section to A, B, C
\renewcommand{\thesubsection}{\thesection.\arabic{subsection}} % Subsections: A.1, A.2, etc.
\section*{Appendices}
\section{Feature-type Distributed Clustering (FDC)}
In NextConvGeN model the neighborhoods are identified using a unique nonlinear dimensional reduction approach called Feature-type Distributed Clustering (FDC) introduced by Bej \textit{et al.} \cite{bej_identification_2022}. This approach provides a more effective framework for generating informative low-dimensional embedding from high-dimensional tabular data than conventional Uniform Manifold Approximation and Projection (UMAP), a commonly used nonlinear dimension reduction technique with a choice of similarity measures. FDC applies the UMAP technique independently to each feature type (continuous, ordinal, and nominal) using specific similarity measures \cite{bej_accounting_2023}. The resulting lower dimensional feature space is more meaningful, considering the diverse feature types present in the data. Hence, we explore the neighborhoods based on this lower dimensional feature space.

\section{Alpha clipping}
\vspace{-20pt}
\begin{figure}[h!]
    \centering
    \includegraphics[width=\textwidth]{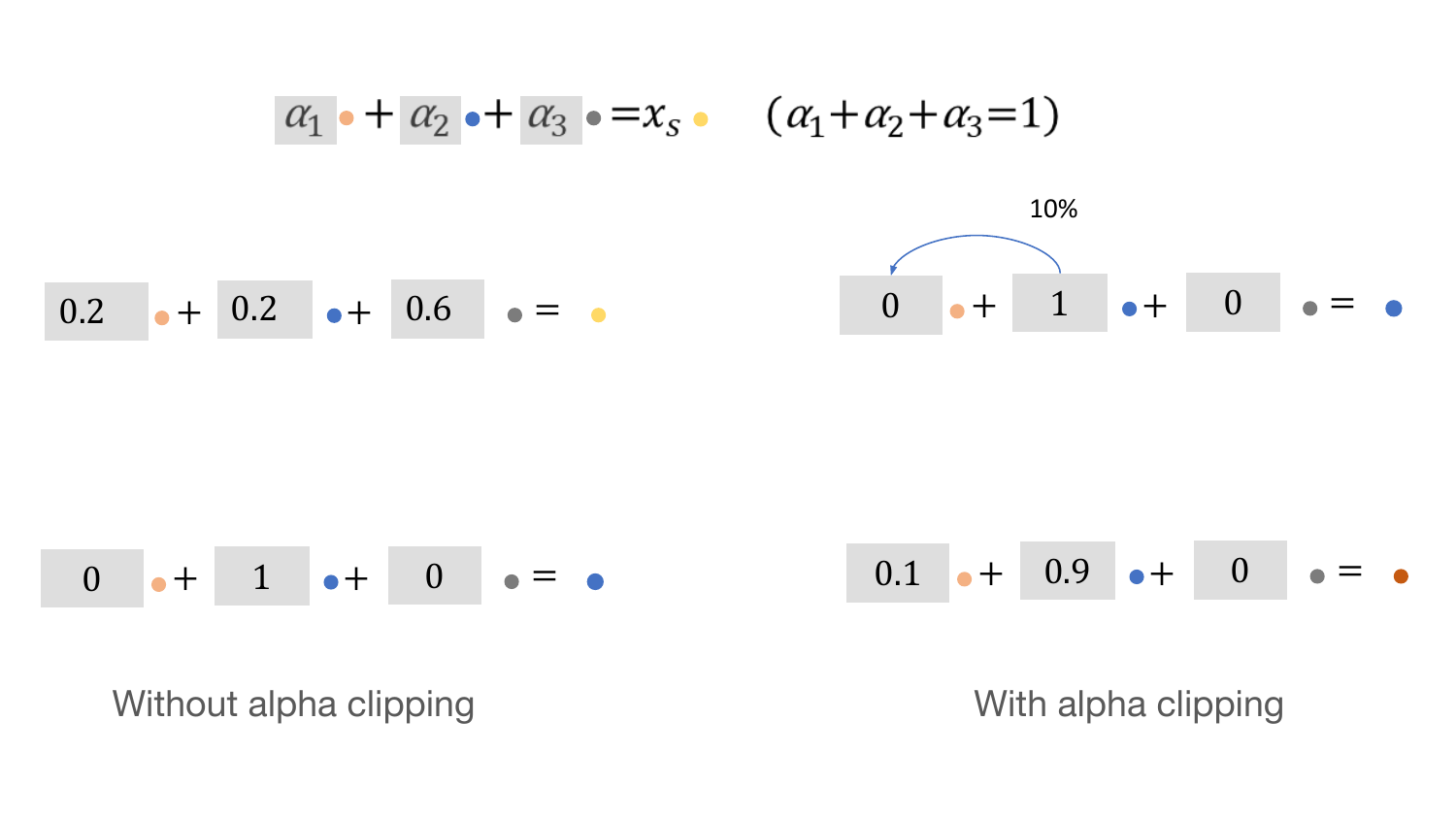}
    \vspace{-1.5cm}
    \caption{Illustration of the alpha clipping process. The left side demonstrates synthetic sample generation without alpha clipping, where a coefficient can reach 1, leading to a synthetic sample identical to a real sample. The right side shows the effect of alpha clipping, where the maximum coefficient is clipped by 10\% (i.e., \var{alpha\_clip} = 0.1)and redistributed, ensuring the synthetic sample remains distinct while maintaining convexity constraints.}
    \label{fig:alpha_clipping}
\end{figure}

The generator produces synthetic samples constrained within the convex hull of the input neighborhood. However, in some cases, one of the convex coefficients $\alpha_j$ in the convex combination $K_N$ may reach one, causing the synthetic sample to replicate a real sample. To mitigate this issue, an \textit{alpha clipping} technique is applied. Let $K_N \in \mathbb{R}^{\texttt{k} \times \texttt{k}}$ denote the convex coefficient matrix, where \[
\sum_{j} K_N(i, j) = 1, \quad K_N(i, j) \geq 0 \; \forall \; i, j.
\]The maximum coefficient $\alpha_{\max} = \max K_N(i, j)$ $\forall i$ is identified, and a fraction $\delta \in [0, 1]$ of its value is clipped \[
\alpha_{\text{clipped}} = \alpha_{\max} - \delta \alpha_{\max}
\] with the clipped value redistributed to the minimum coefficient $\alpha_{\min} = \min K_N(i, j)$ as \[
\alpha_{\text{adjusted}} = \alpha_{\min} + \delta \alpha_{\max}.
\]
See Figure \ref{fig:alpha_clipping} for an example illustrating the mechanism of the alpha clipping technique.

\section{Evaluation measures}
Evaluation of synthetic data is crucial but lacks standardized metrics, especially for tabular data \cite{hernandez_synthetic_2022}. We assess synthetic data quality using Utility and Privacy measures. Utility encompasses task performance and Fidelity, the statistical similarity to real data \cite{jordon_synthetic_2022}. Privacy quantifies information leakage. Table 1 in the manuscript summarizes our evaluation metrics. 
\subsection{Utility measures}
We employed some common Utility measures to evaluate the quality of synthetic data generated by generative models. In this section, we are going to provide a brief description of the utility measures used for synthetic data evaluation.
\renewcommand{\baselinestretch}{1.15}\normalsize
\begin{itemize}
    \item \textbf{Student's t-test: } A t-test, a univariate statistical test, is used to compare the means of two groups to determine if there is a significant difference, particularly when data is continuous and normally distributed. Assuming normal distribution, a t-test is performed on the continuous features of real and synthetic data, with a significance level of $0.05$. The null hypothesis ($h_0$) assumes equal means for real and synthetic features. If the $p$-value exceeds 0.05, $h_0$ is accepted, indicating that the synthetic data preserves the properties of the real data \cite{hernandez_synthetic_2022}.
    
    \item \textbf{Kullback-Leibler (KL) divergence: } KL divergence is used to measure the difference between two probability distributions \cite{hernandez_synthetic_2022}. For our evaluation, we, therefore, adapted this measure to quantify the similarity between the real and synthetic data's probability distributions of each categorical feature. A zero value indicates identical distributions, while larger values signify a greater discrepancy. It is important to note that, like the student's t-test, KL divergence is computed independently for each feature. This means that it does not capture potential dependencies between features in the data \cite{goncalves_generation_2020}.
    
    \item \textbf{Machine learning efficacy: } To calculate this performance measure, first, two Gradient-boosting classifiers are trained on real training data and its corresponding synthetic data separately. Gradient boosting classifier is employed due to its ability to combine weak models into a powerful learner that effectively handles numerical and categorical features. After training, both the models were tested on the same holdout data, which was not seen by either the classifiers or the generative models during training \cite{espinosa_quality_2023}. This is known as holdout data analysis and is equivalent to the Train-on-Synthetic and Test-on-Real (TSTR) and Train-on-Real and Test-on-Real (TSTR) evaluation metric. The absolute difference in the performance (in terms of the F1-score) was calculated. It is important to note that a five-fold cross-validation was performed before training, and the absolute performance difference between the models is calculated. The smaller the differences, the better the machine learning efficiency of synthetic data.\par
    
    \item\textbf{Propensity score:} The feasibility of training a machine learning classification model to differentiate between real and synthetic data is measured by propensity score \cite{pathare_comparison_2023, espinosa_quality_2023, dankar_multi-dimensional_2022}. Also known as the distinguishability metric, the propensity score is a well-known utility measure \cite{dankar_fake_2021, drechsler_synthetic_2011, raab_guidelines_2017}. To calculate the propensity score, each sample is assigned a binary indicator/label, with $1$ for real data and $0$ for synthetic data. A logistic regression classifier is trained to predict the likelihood of a given sample being real. The closer the predicted probability is to $1$, the more likely the sample is a real record, while values closer to $0$ indicate a synthetic record. The following equation defines the propensity score:

    \begin{equation}
        propensity = \frac{1}{n} \sum_{i=1}^{n}\left(p_i - 0.5\right)^2
    \end{equation}
    where $n$ is the total number of sample and $p_i$ is the probability of the $i$\textit{-th} sample being real by the classifier.
    
    A probability of $0.5$ means that the classifier cannot differentiate between the real and synthetic data. Therefore, the probability mean square difference would range between $0$ and $0.25$ if the two datasets were identical and exceed $0.25$ if they were different. \cite{hernandez_synthetic_2022}.\par

    \item \textbf{Log-cluster metric:} The log-cluster metric assesses whether a clustering algorithm can differentiate between real and synthetic data in an unsupervised manner. At first, real and synthetic data were labeled and concatenated. The $k$-Means clustering algorithm is then applied to this concatenated data to extract clusters. The following formula calculates the log cluster metric score $M(X_R, X_S)$, indicating how well the clustering algorithm can distinguish between real and synthetic data \cite{goncalves_generation_2020}.
    \begin{equation}
        M(X_R, X_S) = \log \left( \frac{1}{k} \sum_{i=1}^{k} \left[\frac{n^R_i}{n_i} - c \right]^2 \right)
    \end{equation}
    Where:
    \begin{align*}
        & k \text{ is the number of clusters, in our case, we extract two clusters,} \\
        & n^R_i \text{ is the number of points in the } i\text{-th cluster from the real data,} \\
        & n_i \text{ is the total number of points in the } i\text{-th cluster,} \\
        & c \text{ is the proportion of real tuples in the merged dataset,} \\
        & \quad \text{calculated as } c = \frac{n^R}{n^R + n^S}, \text{ where } n^S \text{ is the number of points from the synthetic data.}
    \end{align*}
    
    A larger score in terms of magnitude indicates that a clustering algorithm cannot differentiate between real and synthetic data.\par

    \item\textbf{Cross classification:} The cross-classification is a metric that measures how well the synthetic data captures the statistical dependency structure of real data  \cite{dankar_multi-dimensional_2022}. This measures the dependence via prediction accuracy (F1-score) of one feature based on the features in the data. To calculate this metric, two classifiers are trained on synthetic and real data separately, with each categorical feature considered as a target/label \cite{dankar_multi-dimensional_2022}. The performance of both models is then evaluated on holdout data, and the absolute difference between their performance is calculated. The average absolute difference over the features is reported in Section \ref{results}. If this difference is close to zero, it is expected that the synthetic data is similar to real data in preserving the statistical dependencies between the features.

    \end{itemize}

\subsection{Privacy measures}
Benchmarking studies often evaluate how well synthetic data generation models preserve privacy, though only a few privacy measures exist, and no standard measure has been established. In our experiments, we used three privacy measures: Distance metrics, Membership Inference Attack (MIA), and Attribute Inference Attack (AIA). A small average distance between real and synthetic data points suggests higher privacy risks. MIA involves determining if a specific record was used in training, while AIA occurs when an attacker, knowing some attributes of the real data, uses synthetic data to infer unknown attributes. Further details on these measures are provided below.
\renewcommand{\baselinestretch}{1.15}\normalsize
\begin{itemize}
    \item \textbf{Distance metrics:} Hernadez \textit{et al.} suggested three similarity measures to evaluate how private synthetic data is compared to real data. The metrics employed include Euclidean distance, Hausdorff distance, and cosine similarity. The pairwise similarity value is calculated for each pair of records between real and synthetic, and the mean values of those pairwise similarity values are analyzed. Euclidean distance captures the difference between two data points by calculating the straight-line distance in a multidimensional space. A smaller Euclidean distance indicates a high risk of privacy breach. Hausdorff distance calculates the maximum pairwise shortest distance between real and synthetic data. A smaller Hausdorff distance is associated with a higher privacy risk. Conversely, cosine similarity measures the directional similarity between data points, reflecting how closely their angles align in the multidimensional space. The higher the value of cosine similarity, the higher the privacy risk. 
    \item \textbf{MIA:} This measure evaluates the re-identification risk in synthetic patient data through a simulated attack scenario. The attacker has access to the complete synthetic data and a random subset of real patient data (special case where only 50\% used for generative model training) ranging in availability from 20\% to 100\% (It might contain samples that have been used for model training). Following research by Goncalves \textit{et al.}, we compare real data samples to the synthetic data samples using Hamming distance, a metric indicating attribute differences \cite{goncalves_generation_2020}. If the distance falls below a predefined threshold ($0.4$, $0.3$, $0.2$, or $0.1$), the attacker suspects a match, suggesting a patient record from the real data was used to generate the corresponding synthetic records. Precision, the proportion of correctly identified matches, is calculated to assess the effectiveness of the attack. According to Hernadez \textit{et al.}  and Mendelevitch \textit{et al.}, lower precision (below $0.5$) indicates good privacy preservation, while higher values suggest an increased risk of re-identification  \cite{mendelevitch_fidelity_2021}. By plotting precision against the percentage of accessible real data and the chosen threshold, researchers can analyze the impact of attacker ability and similarity thresholds on the re-identification risk of the synthetic data, which we have used later in Figure \ref{MIA}. 
    \item \textbf{AIA:} This measure investigates how attackers can use synthetic data to predict missing sensitive attributes in real data. The attackers will have access to complete synthetic data and quasi-identifiers for some real records. Quasi-identifiers are attributes or features that are not unique identifiers, such as age or gender. A decision tree classifier or regressor is trained on quasi-identifiers of synthetic data to learn the sensitive feature or attribute depending upon the target type \cite{hernandez_synthetic_2022}. These trained models are used to predict the sensitive information of the real data with the quasi-identifier. The model is evaluated by calculating accuracy for categorical features and root-mean-squared error (RMSE) for numerical features. High accuracy or low RMSE indicates the attacker successfully predicted missing attributes, posing a high disclosure risk. Conversely, lower accuracy and higher RMSE suggest a low risk of attackers inferring sensitive information using this approach \cite{mendelevitch_fidelity_2021}.
  
\end{itemize}

\begin{figure}[htp]
\centering
\includegraphics[width=.3\textwidth]{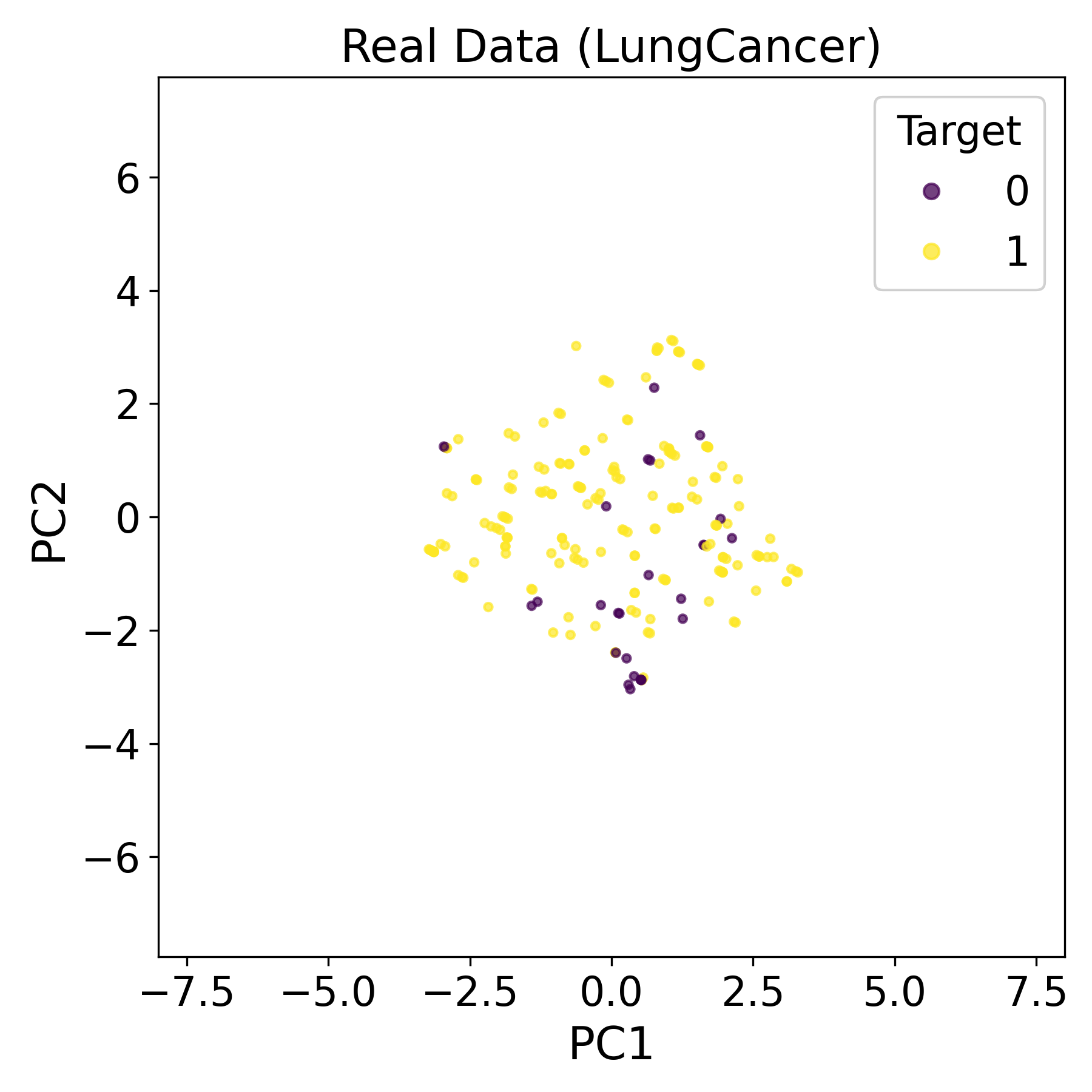}\quad
\includegraphics[width=.3\textwidth]{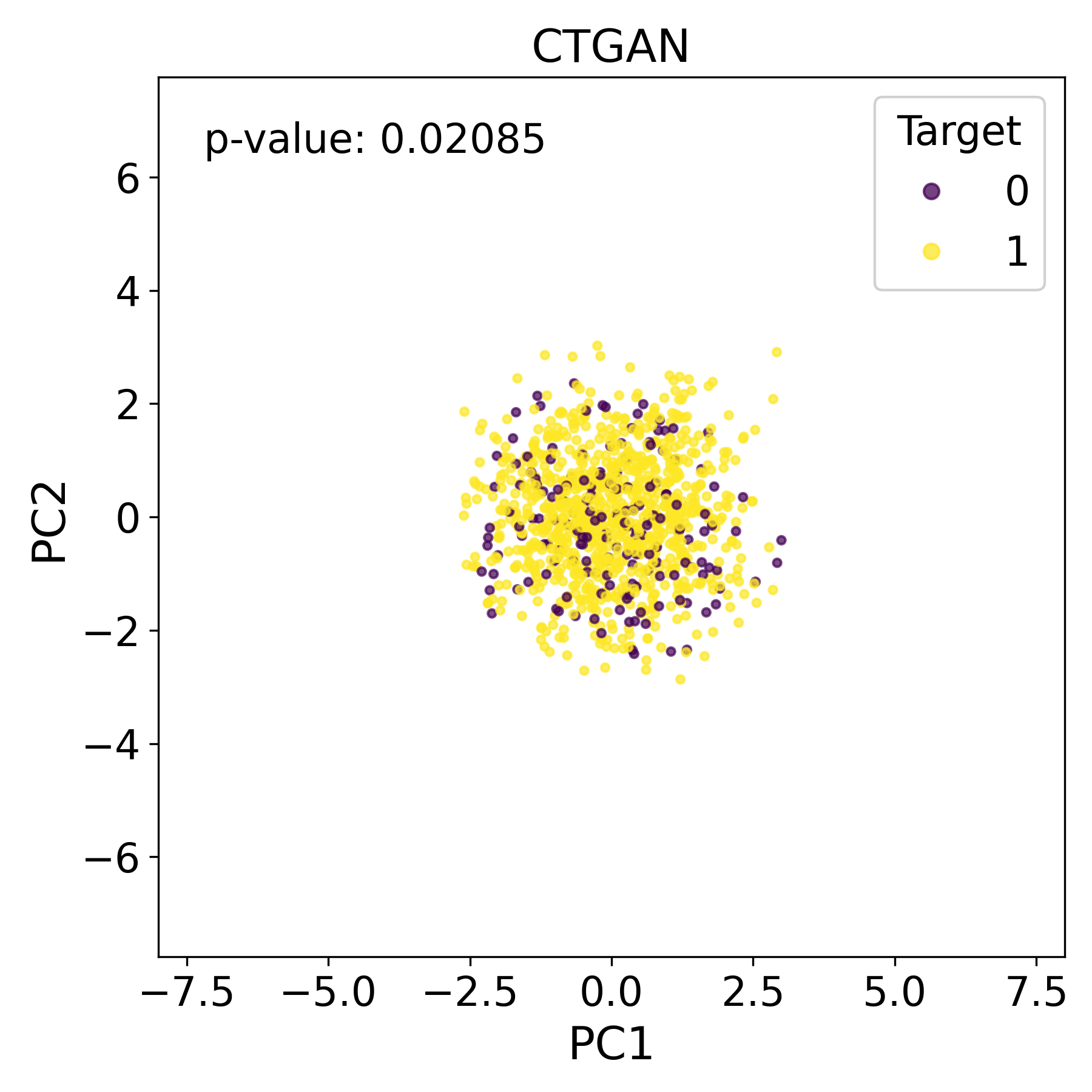}\quad
\includegraphics[width=.3\textwidth]{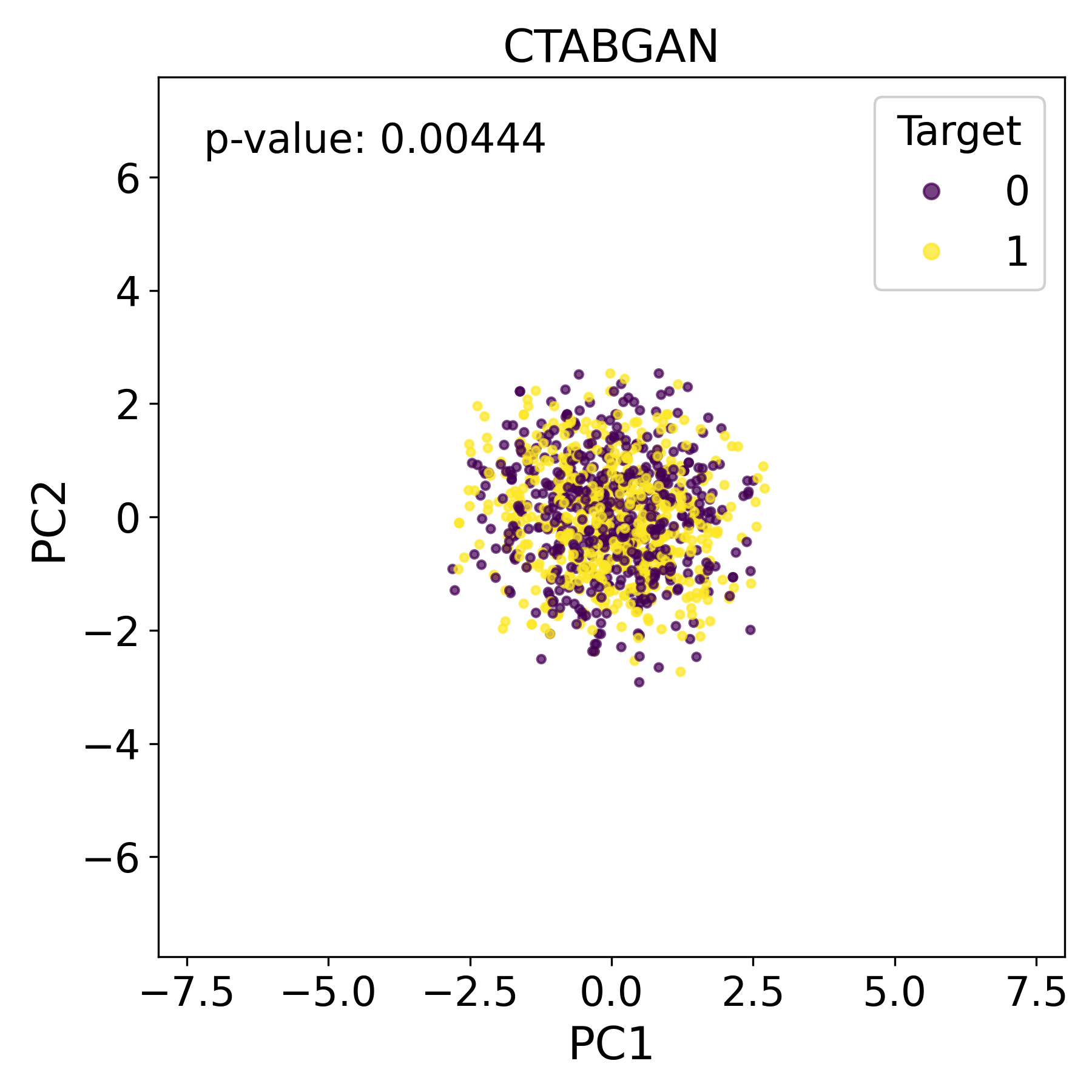}

\medskip

\includegraphics[width=.3\textwidth]{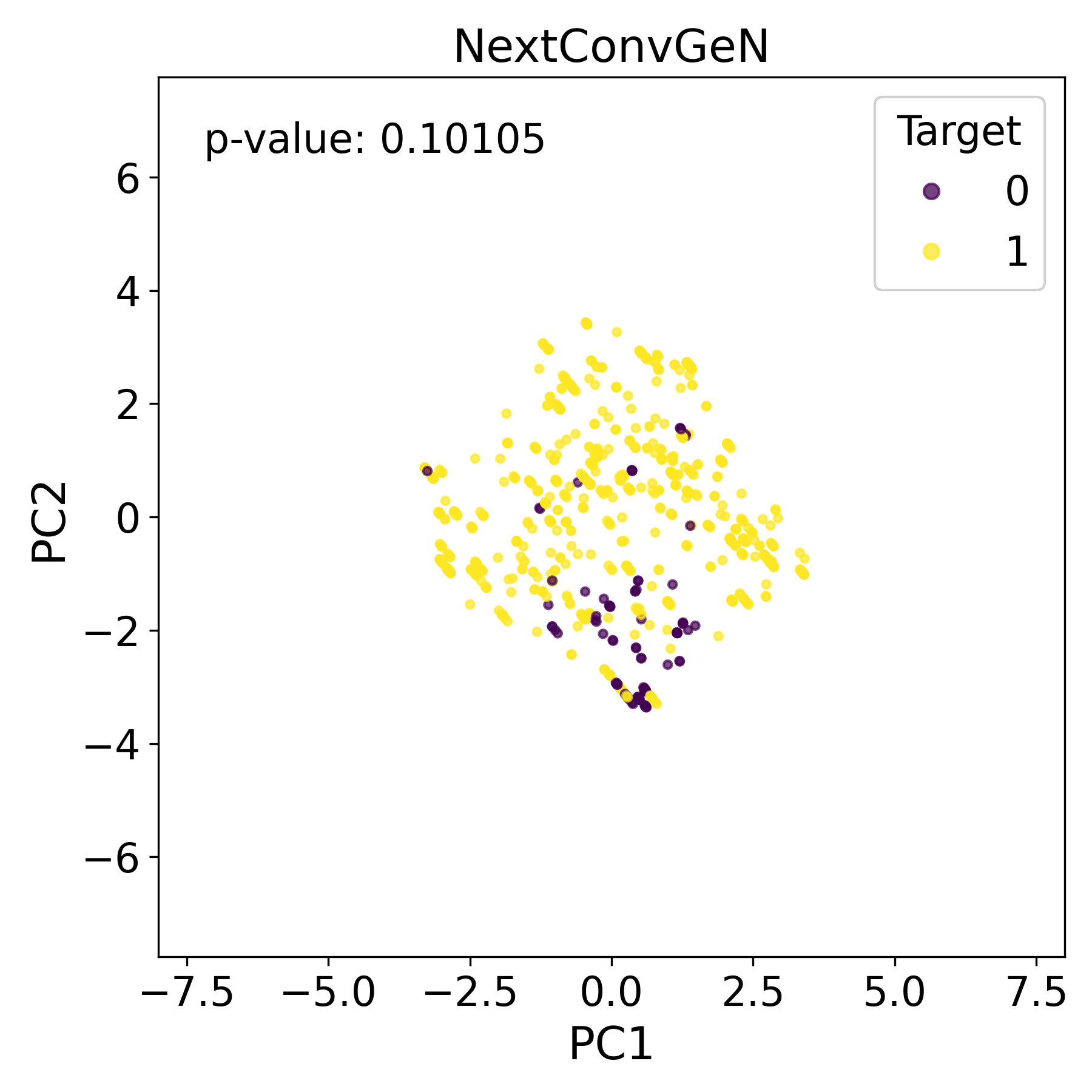}\quad
\includegraphics[width=.3\textwidth]{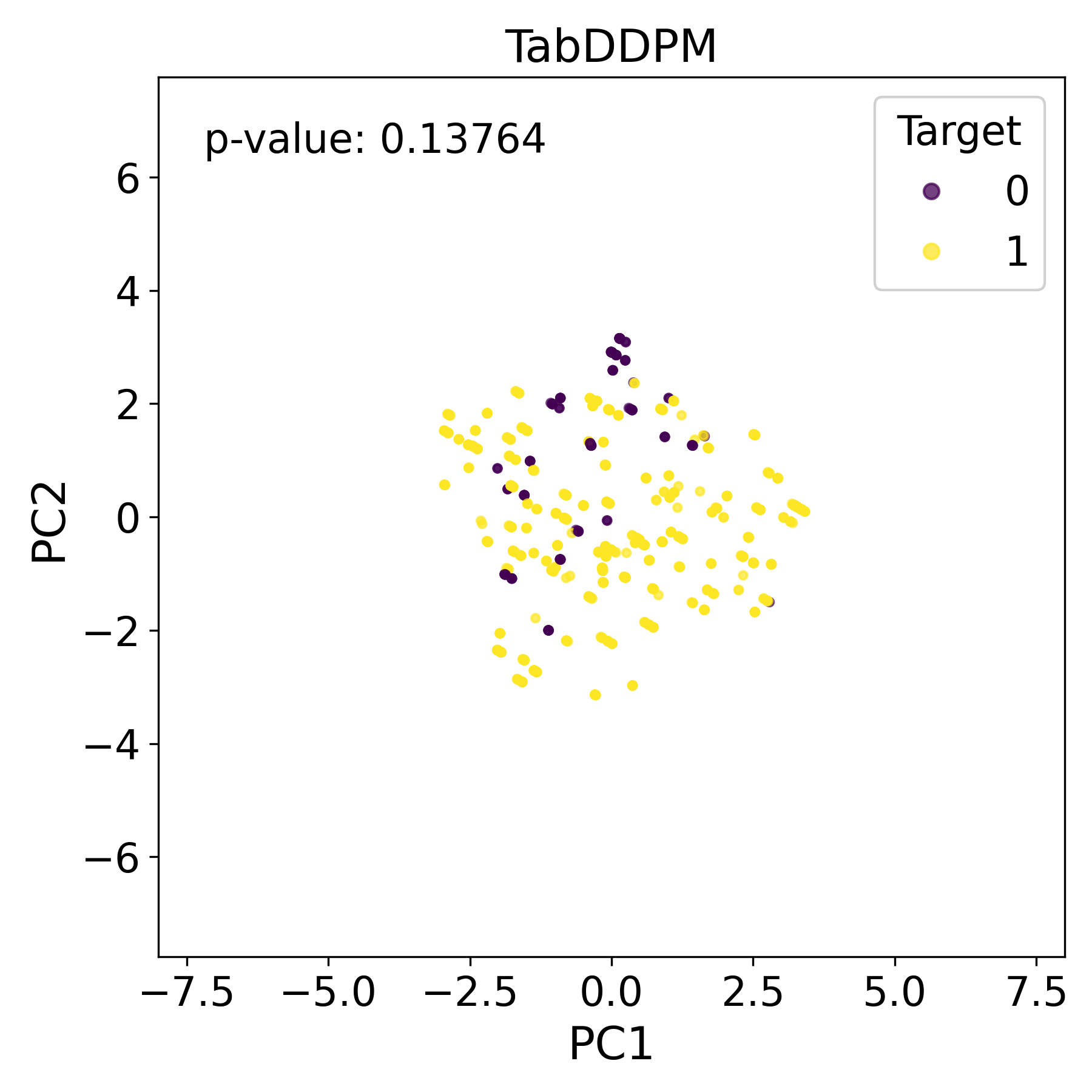}

\caption{PCA visualization of the first two principal components for Real data and synthetic data generated using CTGAN, CTABGAN, NextConvGeN, and TabDDPM models on the Lung dataset. The plot demonstrates that the synthetic data produced by the NextConvGeN and the TabDDPM models closely resembles the distribution of the real data, outperforming the other generative models in preserving the data structure.}
\label{lung cancer pca plots}
\end{figure}

\begin{figure}[htp]
\centering
\includegraphics[width=.3\textwidth]{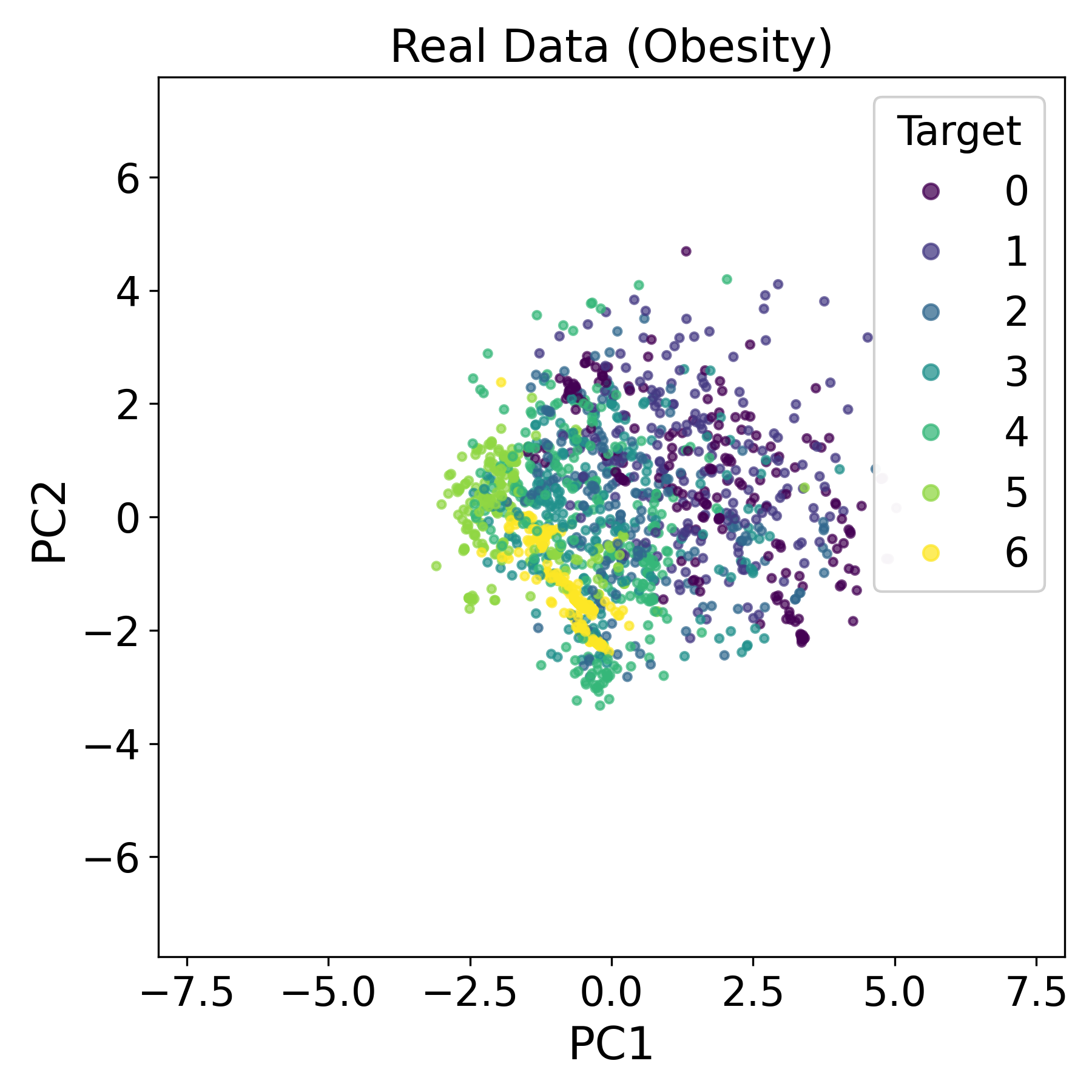}\quad
\includegraphics[width=.3\textwidth]{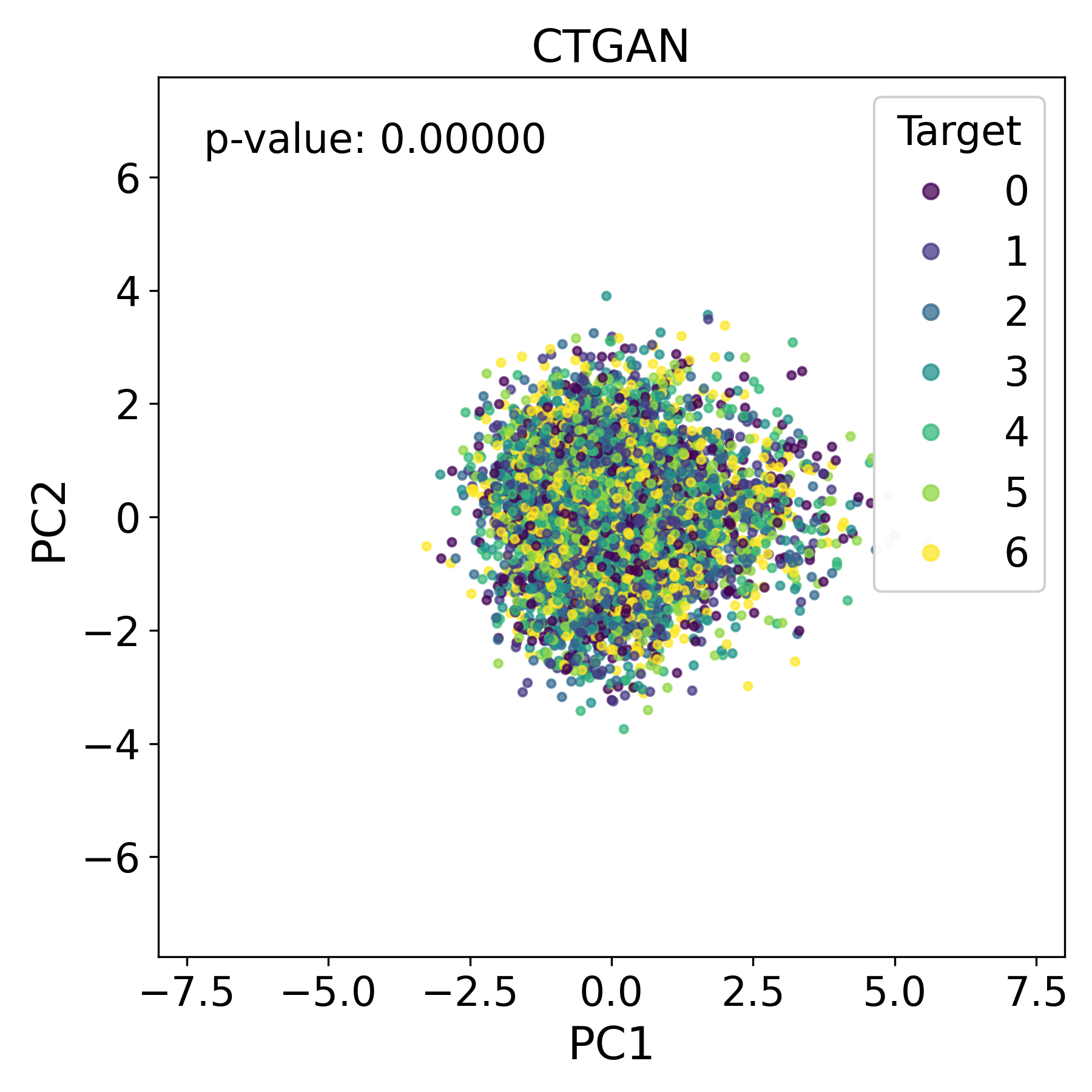}\quad
\includegraphics[width=.3\textwidth]{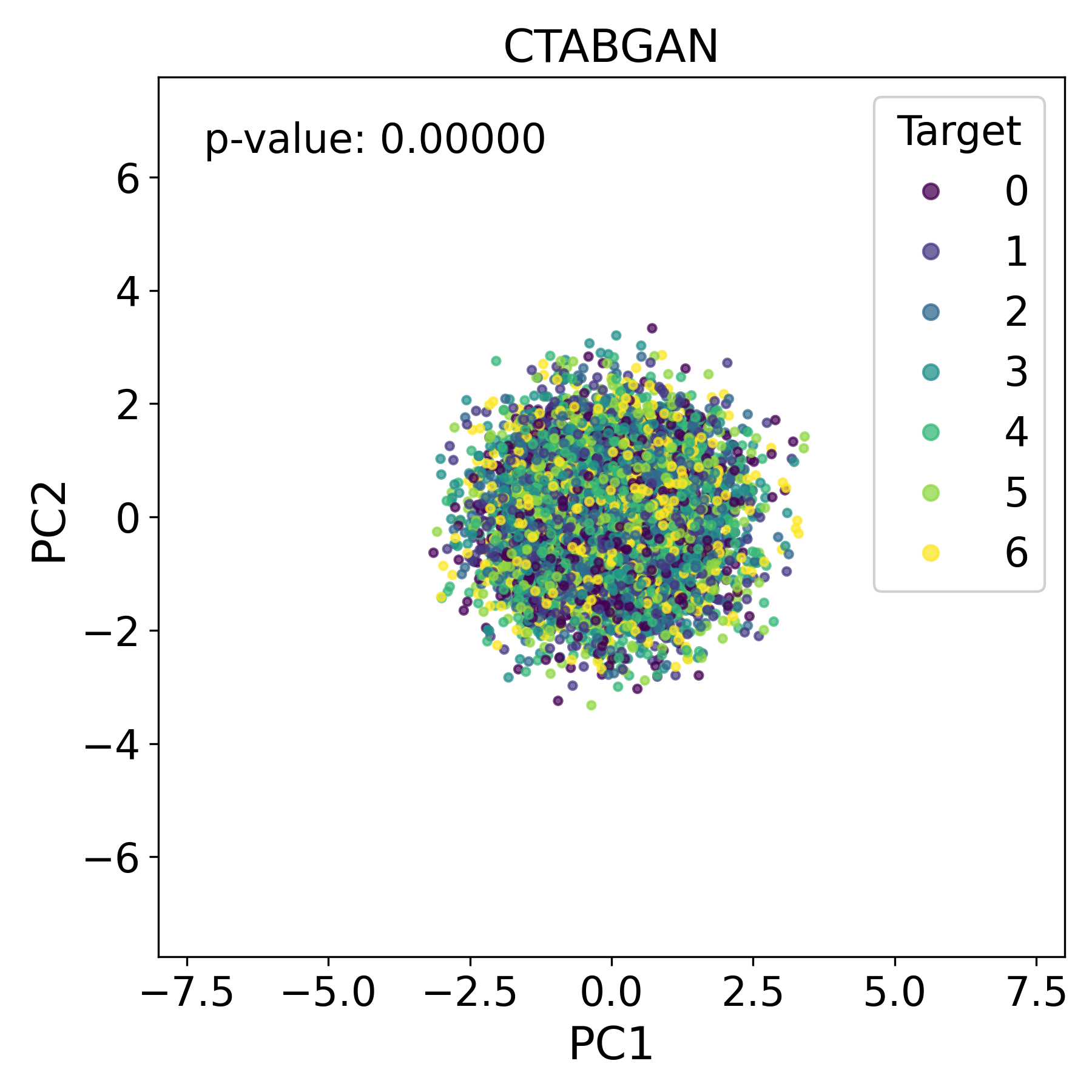}

\medskip

\includegraphics[width=.3\textwidth]{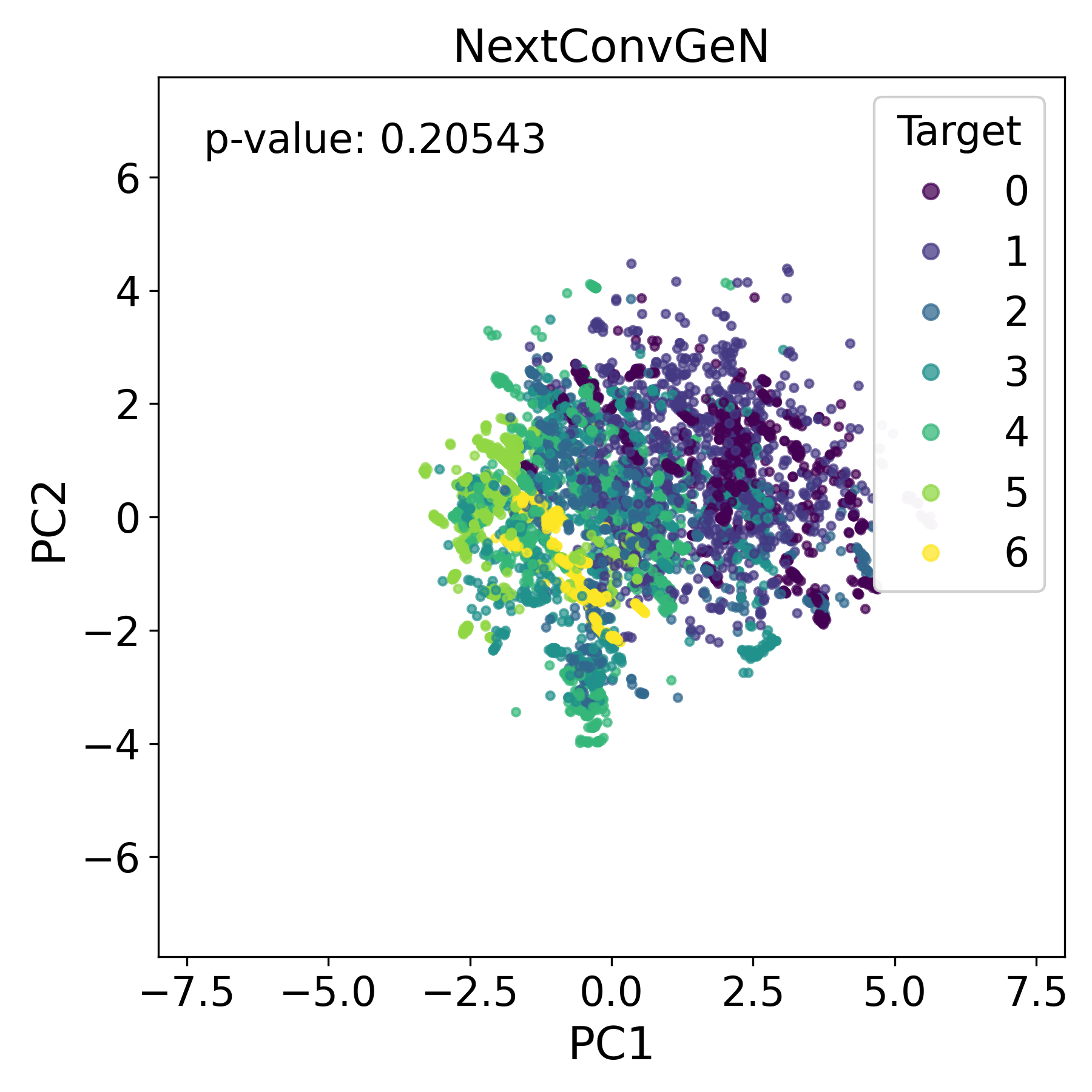}\quad
\includegraphics[width=.3\textwidth]{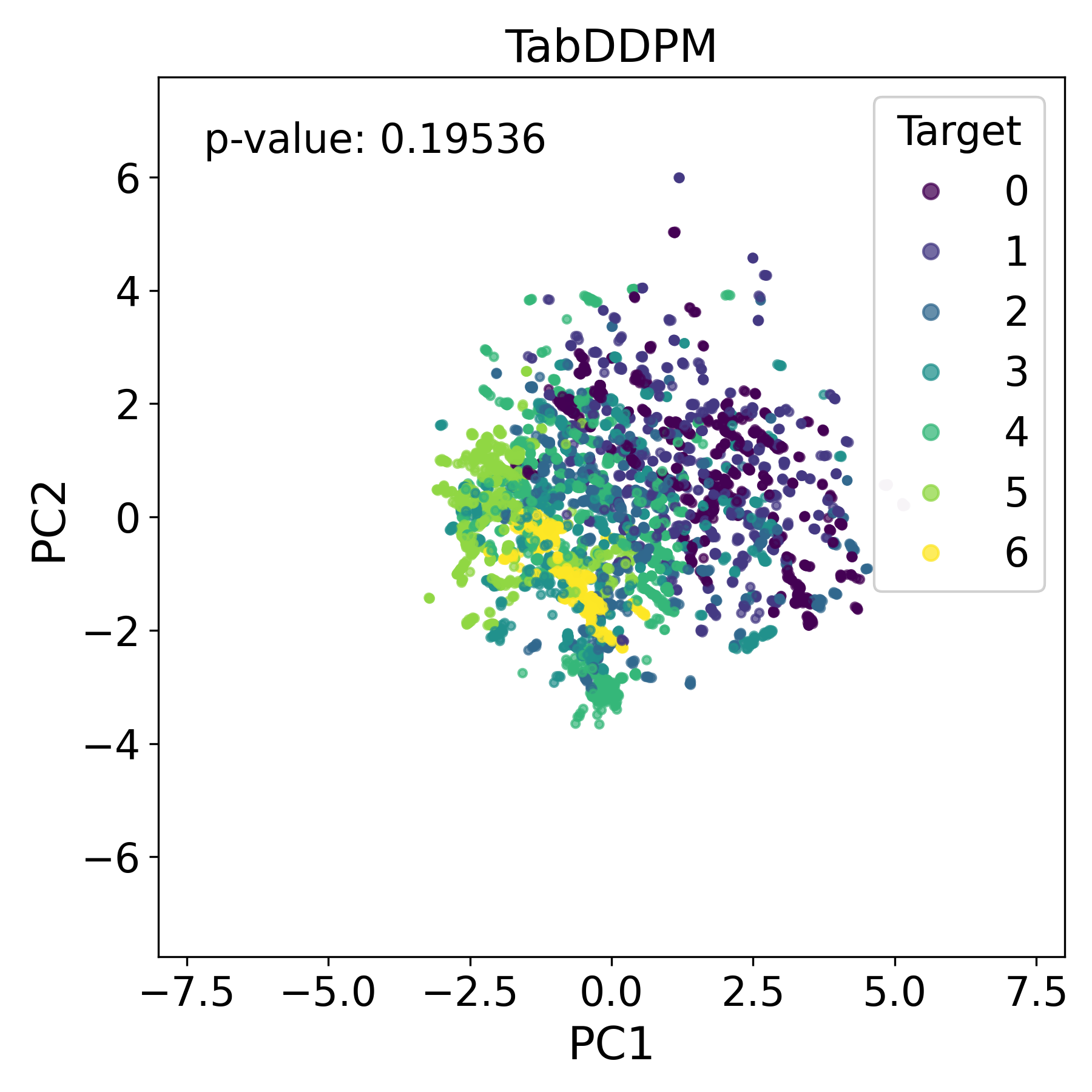}

\caption{PCA visualization of the first two principal components for Real data and synthetic data generated using CTGAN, CTABGAN, NextConvGeN, and TabDDPM models on the Obesity dataset. The plot demonstrates that the synthetic data produced by the NextConvGeN and the TabDDPM model closely resembles the distribution of the real data, outperforming the other generative models in preserving the data structure.}
\label{Obesity pca plots}
\end{figure}

\begin{figure}[htp]
\centering
\includegraphics[width=.3\textwidth]{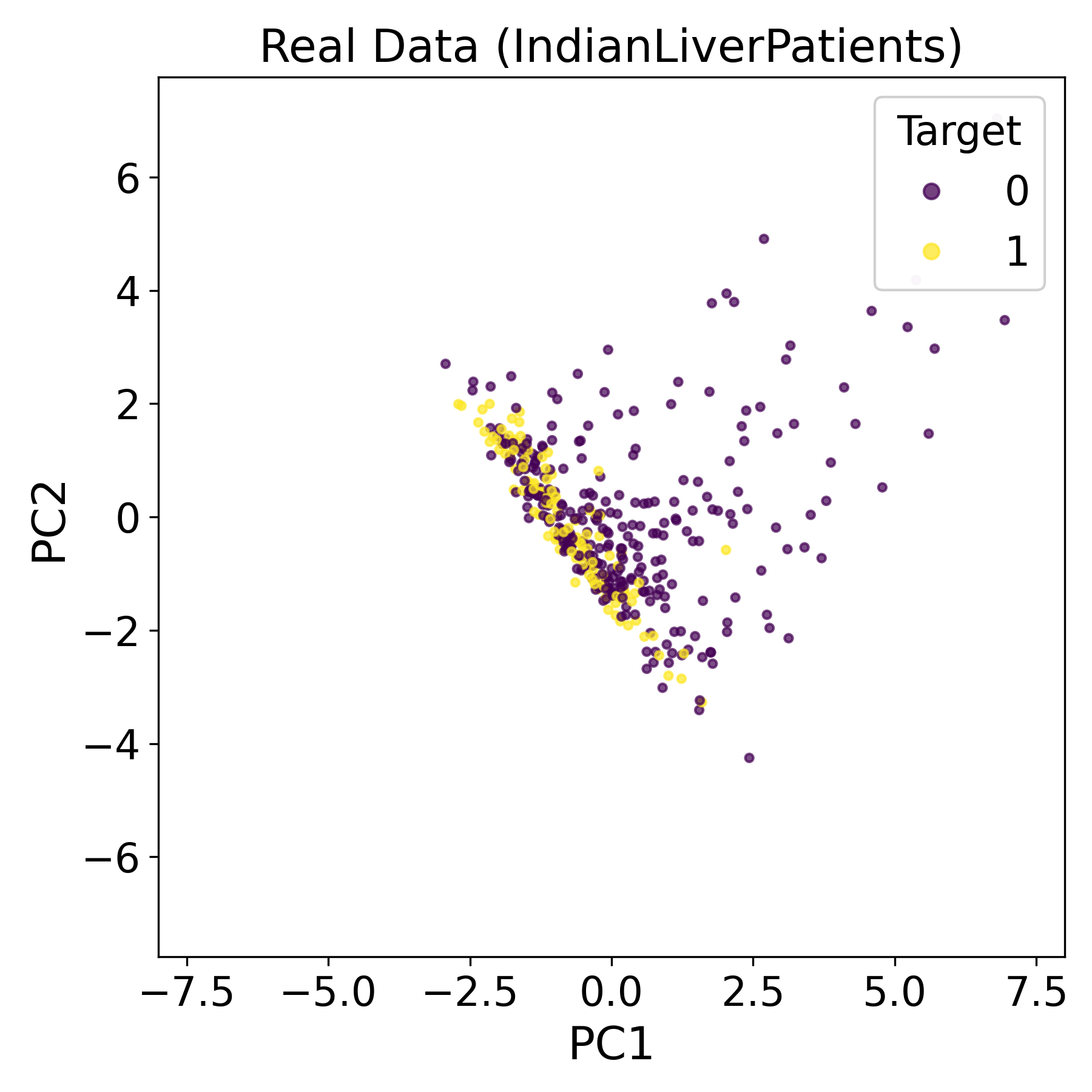}\quad
\includegraphics[width=.3\textwidth]{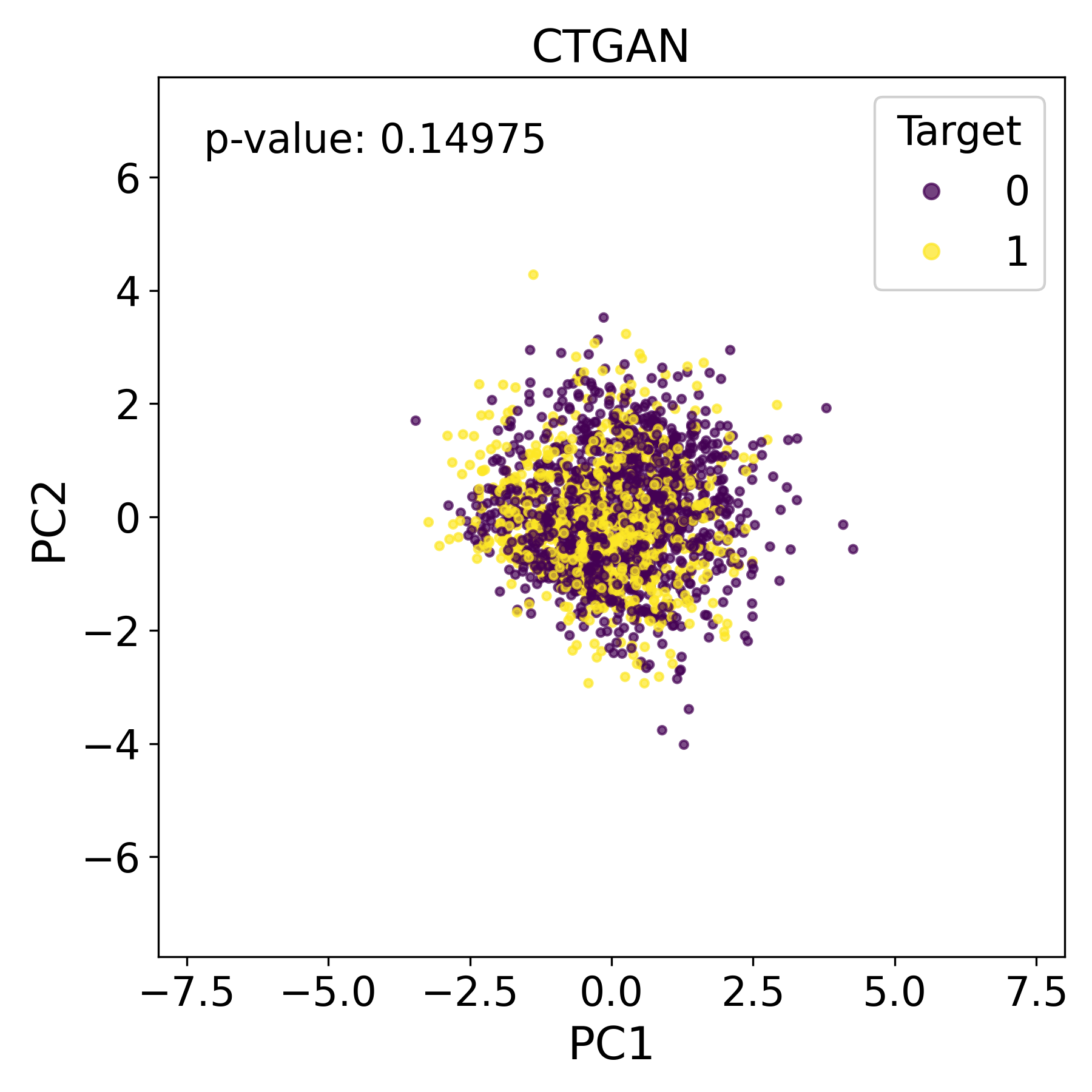}\quad
\includegraphics[width=.3\textwidth]{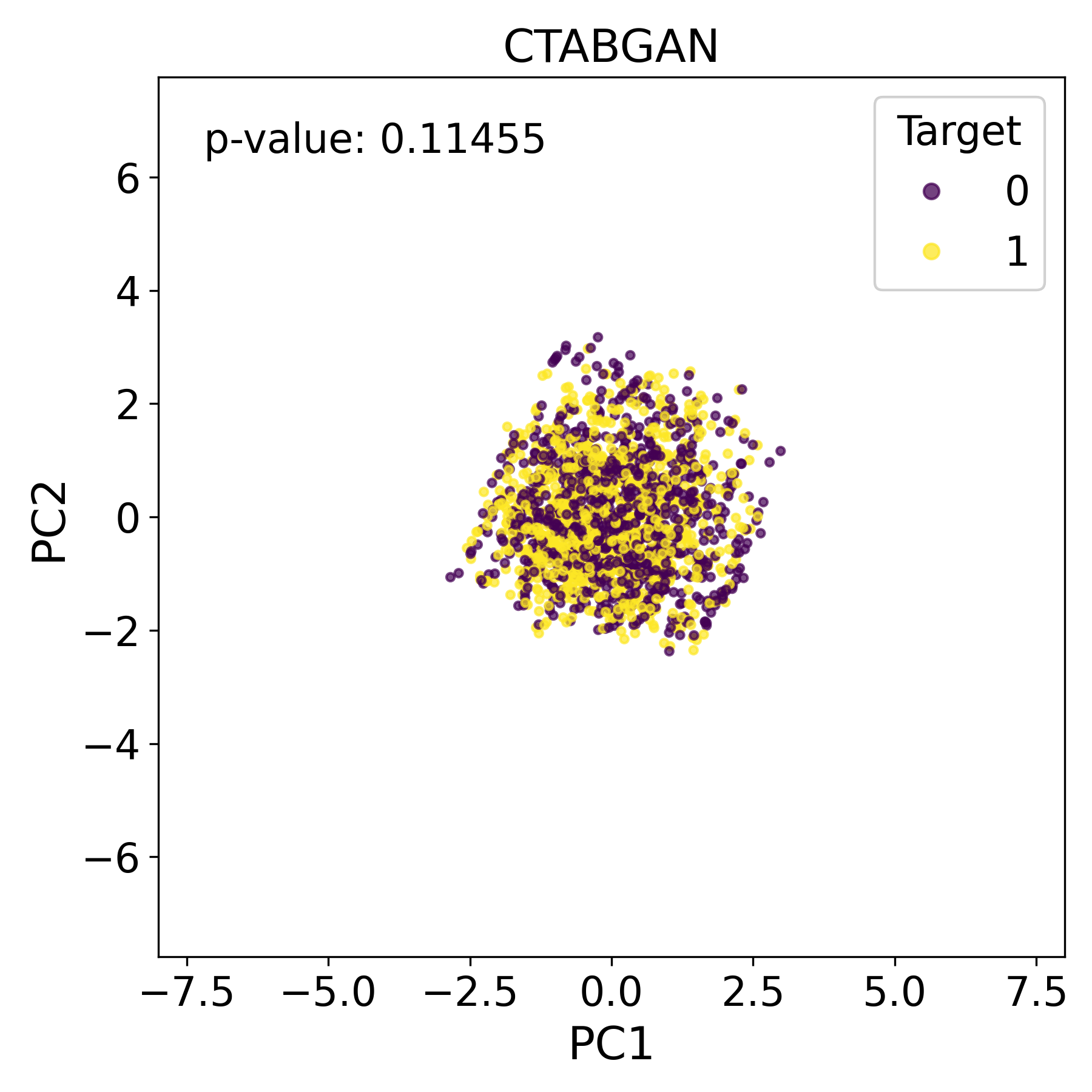}

\medskip

\includegraphics[width=.3\textwidth]{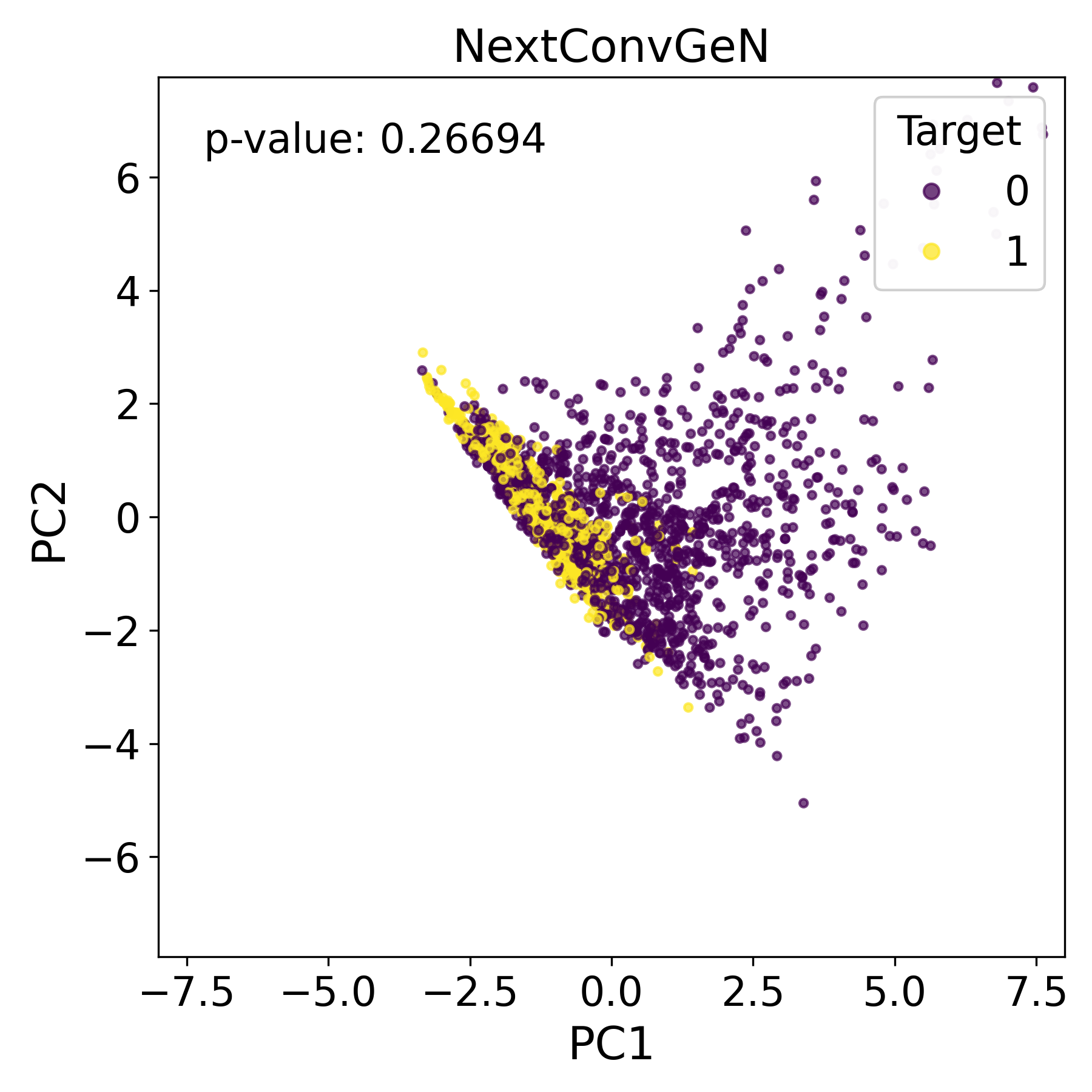}\quad
\includegraphics[width=.3\textwidth]{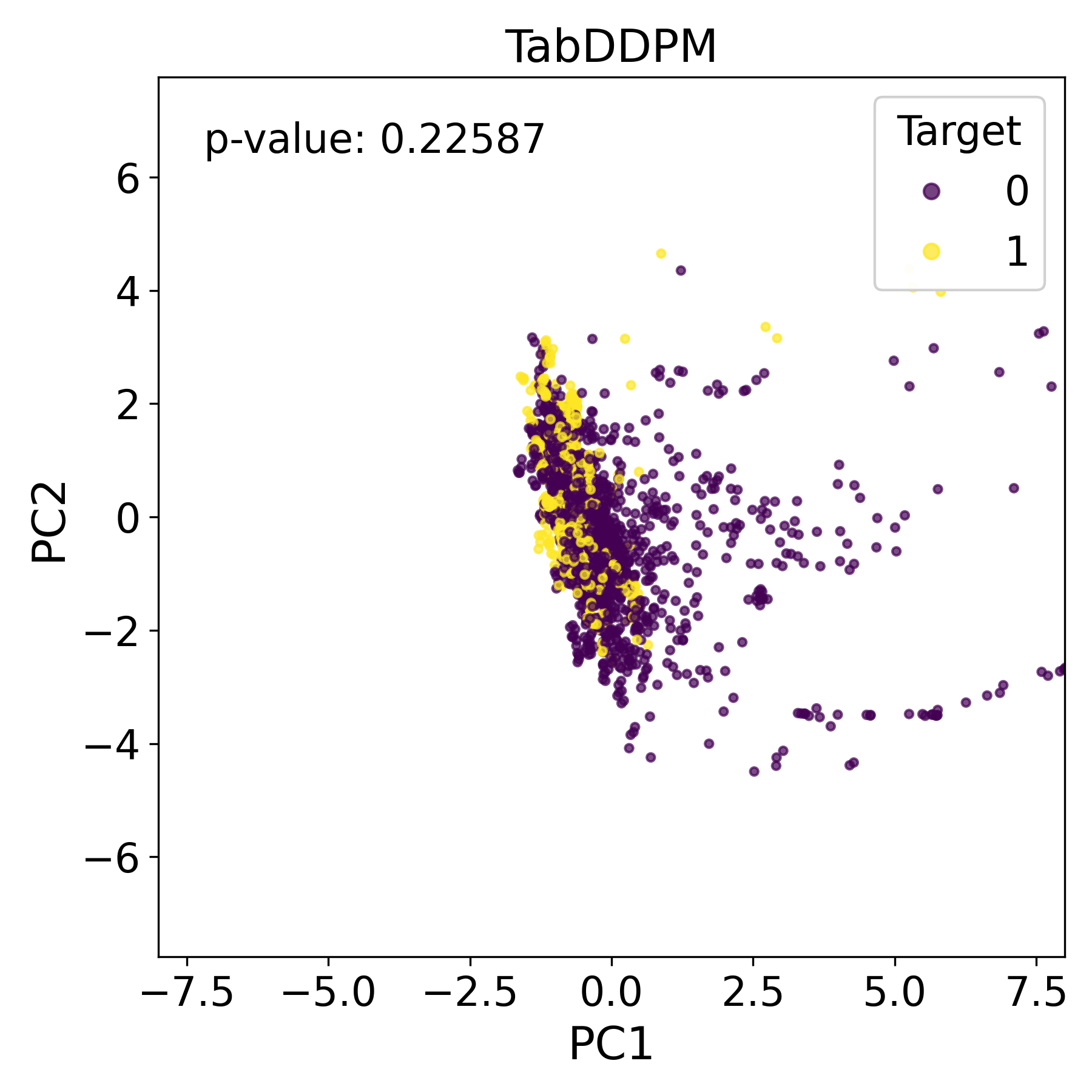}

\caption{PCA visualization of the first two principal components for Real data and synthetic data generated using CTGAN, CTABGAN, NextConvGeN, and TabDDPM models on the Indian liver patients dataset. The plot demonstrates that the synthetic data produced by the NextConvGeN and the TabDDPM model closely resembles the distribution of the real data, outperforming the other generative models in preserving the data structure.}
\label{IndianLiverPatients pca plots}
\end{figure}

\section{Ablation study}
To assess the effectiveness of the proposed NextConvGeN model, we conducted an ablation study to analyze the impact of key parameters, namely the \textit{fdc} and \textit{alpha\_clip} parameters. The study compares the performance of NextConvGeN with and without these parameters. Specifically, we generated synthetic samples from NextConvGeN for the datasets used for benchmarking in the manuscript, both with and without the \textit{fdc} parameter.

\begin{figure}[htp]
\centering
\includegraphics[width=.3\textwidth]{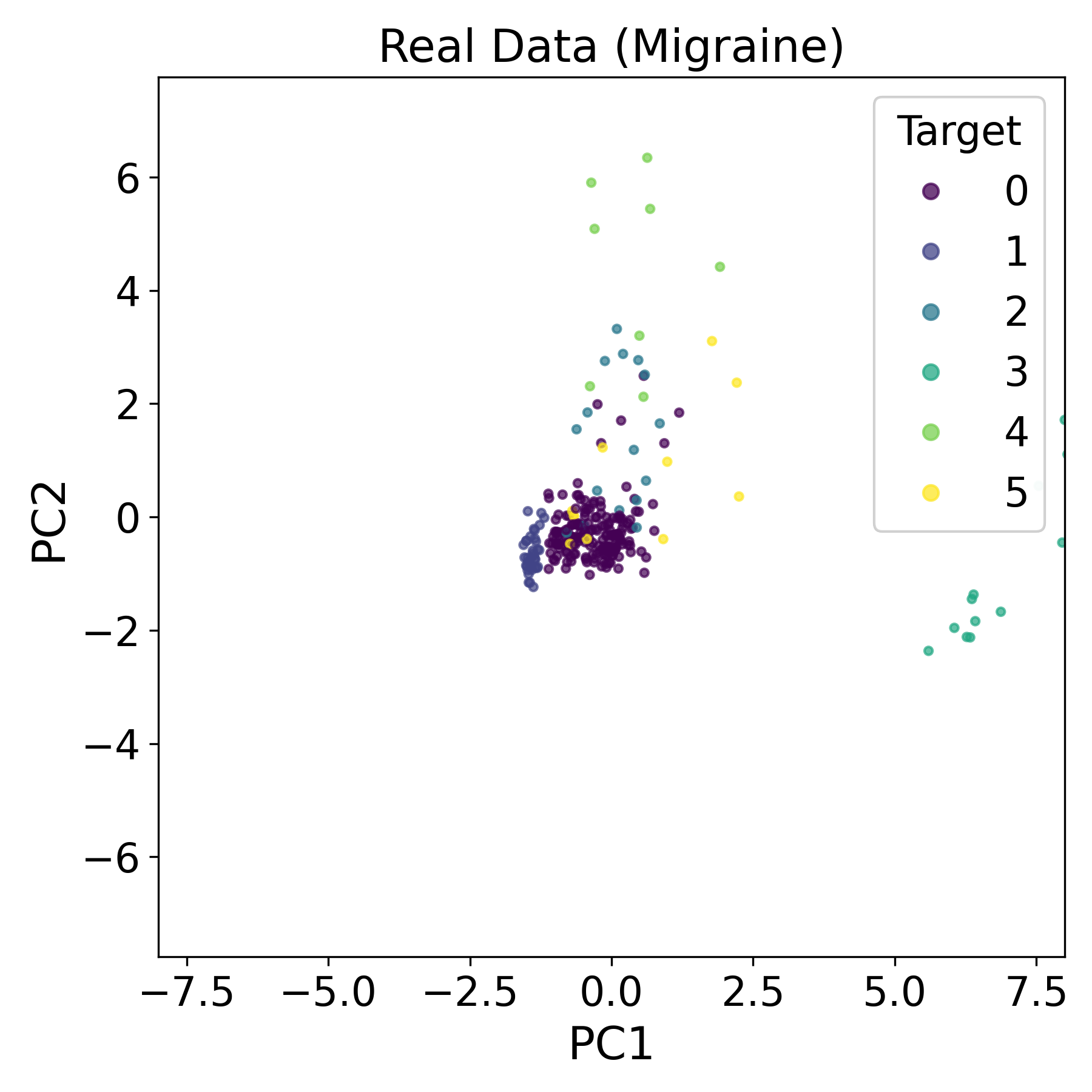}\quad
\includegraphics[width=.3\textwidth]{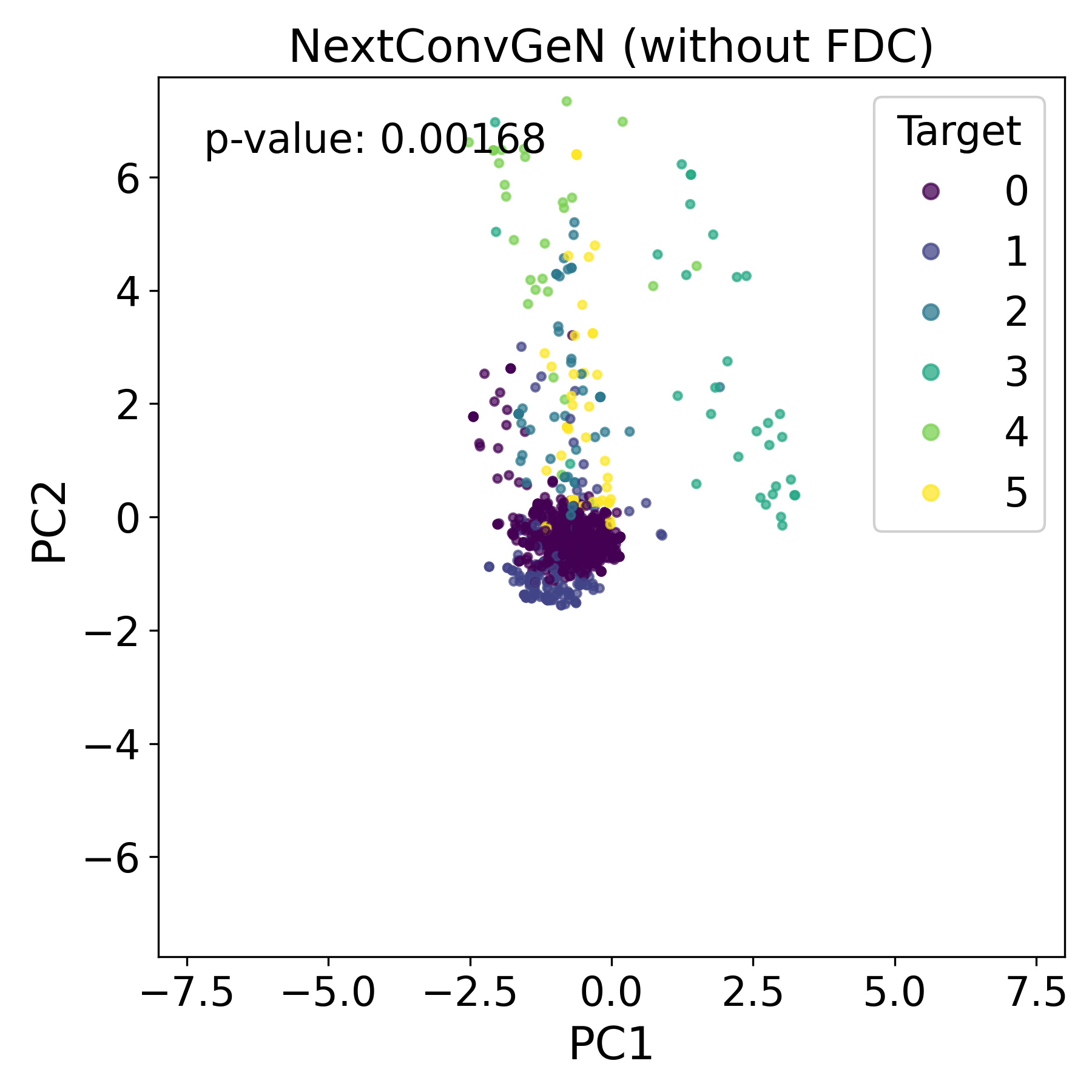}\quad
\includegraphics[width=.3\textwidth]{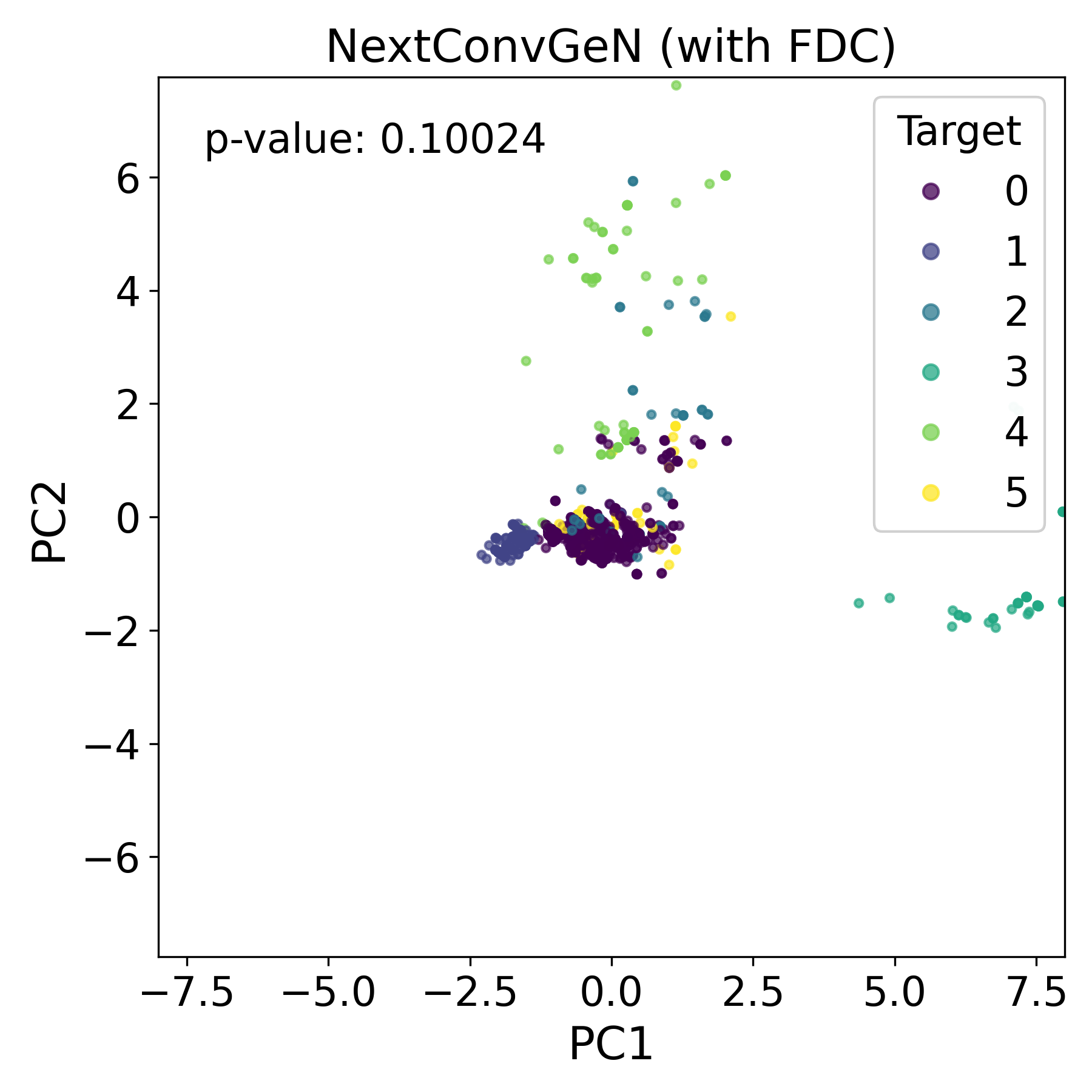}

\medskip

\includegraphics[width=.3\textwidth]{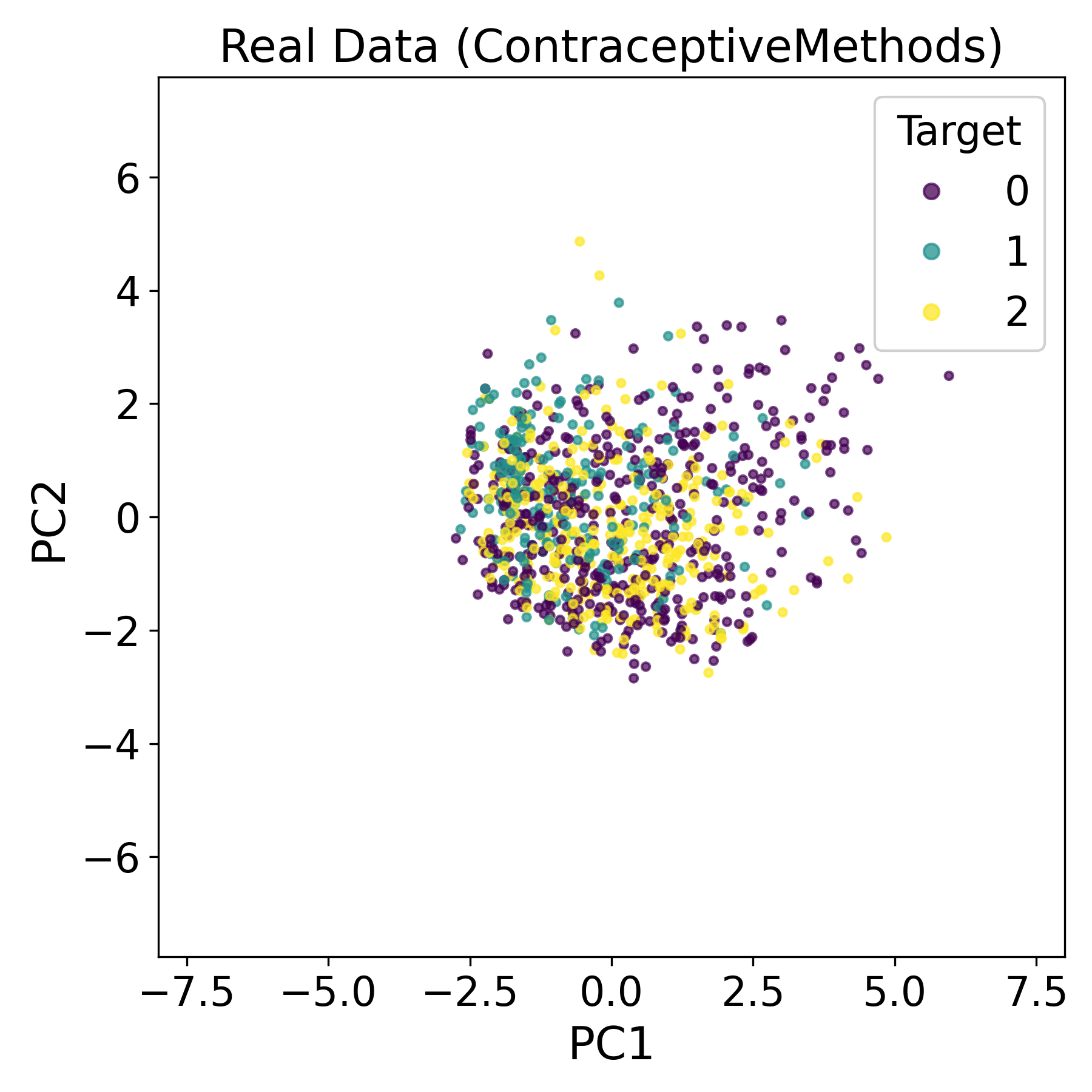}\quad
\includegraphics[width=.3\textwidth]{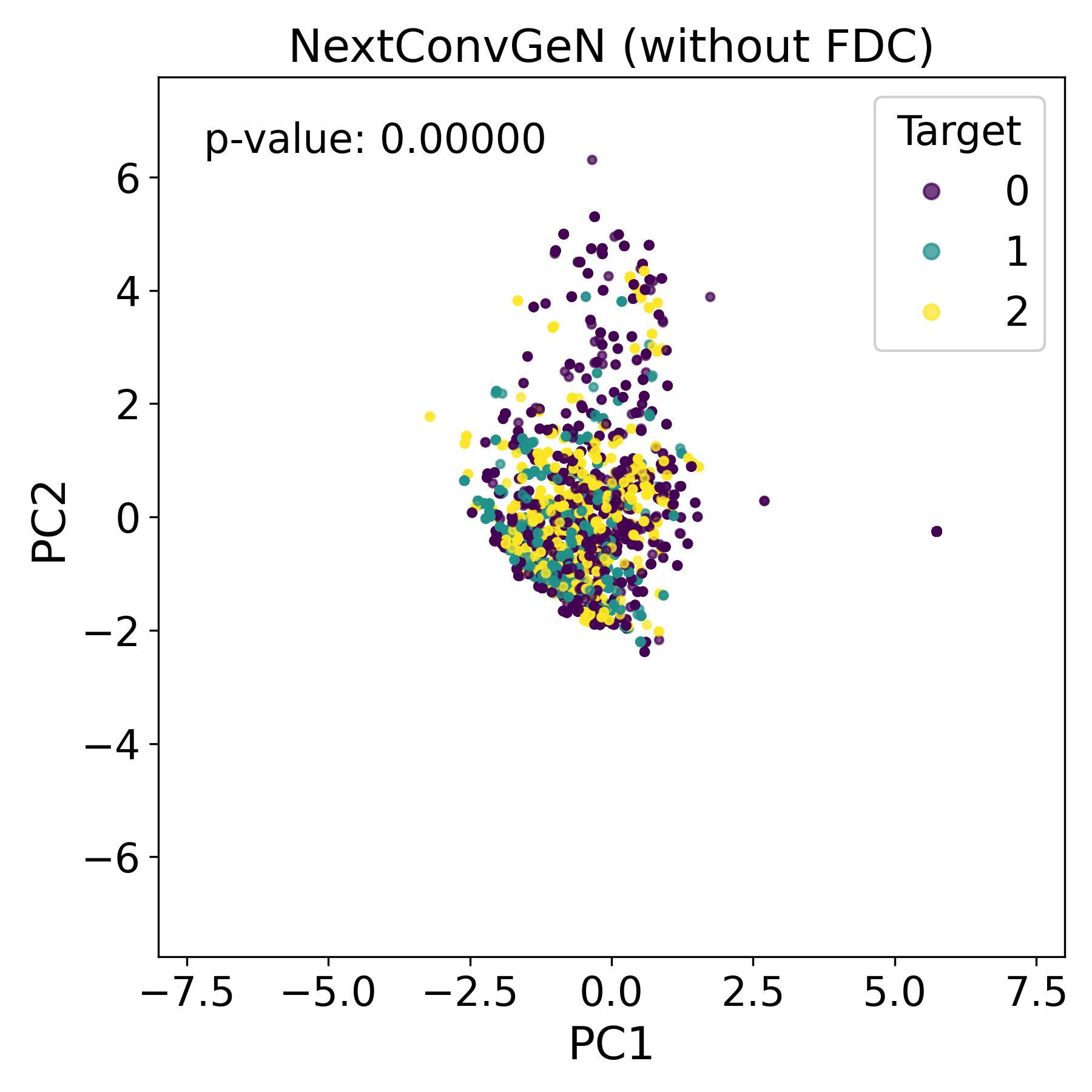}
\quad
\includegraphics[width=.3\textwidth]{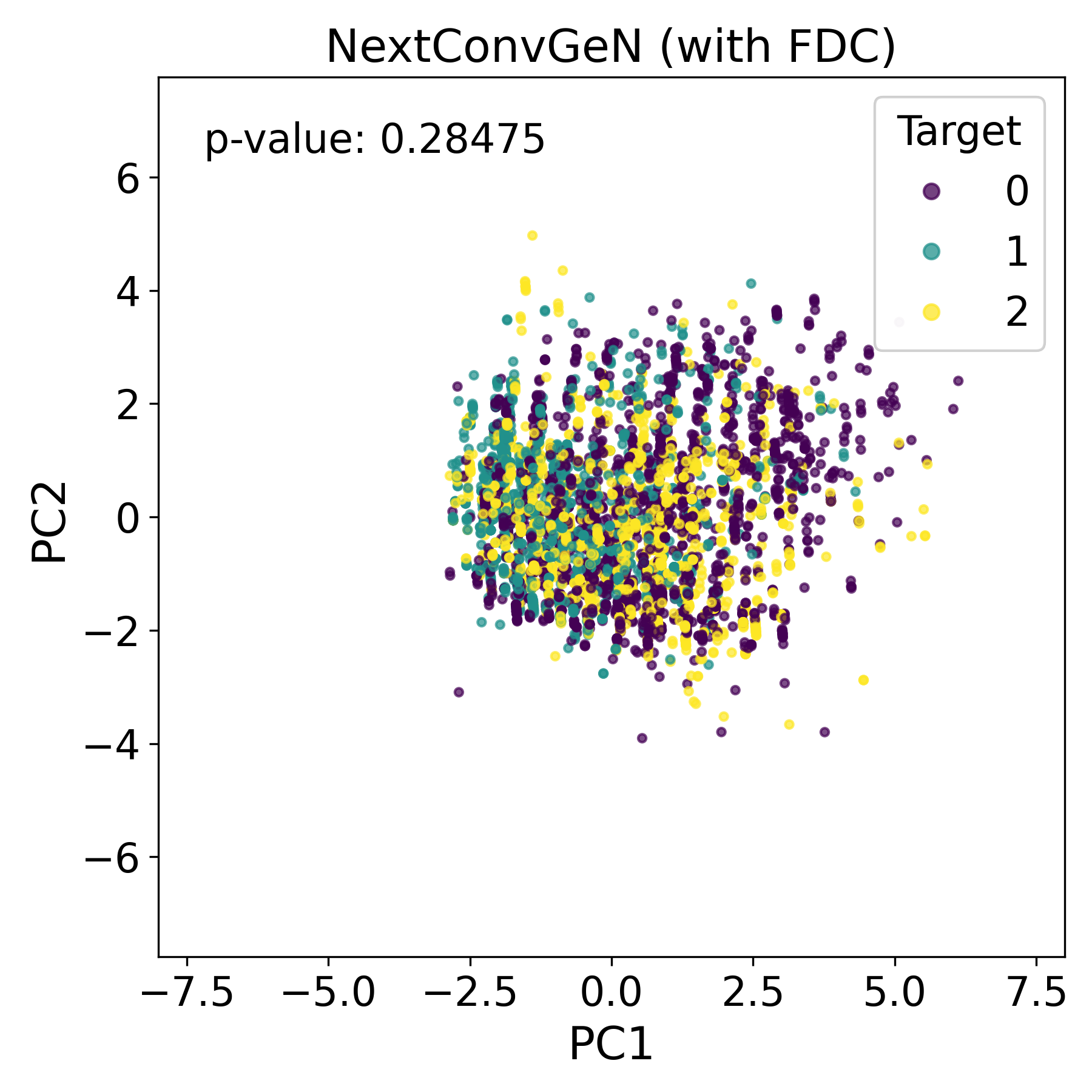}

\caption{Visualization of the first two PCA components for the real data, synthetic data generated by NextConvGeN without the \textit{fdc} parameter, and synthetic data generated by NextConvGeN with the FDC parameter, for the Migraine dataset (first row) and the Contraceptive dataset (second row). The plots demonstrate that the synthetic data generated with the \textit{fdc} parameter aligns more closely with the real data distribution, as evidenced by the visual overlap and p-values from the Peacock test exceeding $0.05$, indicating that the real and synthetic data are from the same distribution.}
\label{fdc ablation plots}
\end{figure}

The \textit{fdc} parameter facilitates neighbor search in a reduced feature space (fdc-reduced dimension), which is designed to account for the diverse feature types (continuous, nominal, and ordinal) present in the datasets. In the absence of this parameter, the neighbor search operates in the high-dimensional feature space. To evaluate the impact of this parameter, we extracted two principal components (PCA) from both the synthetic data and their corresponding real dataset. The reduced-dimensional representations were then compared using the 2D Kolmogorov–Smirnov test (Peacock test), alongside visual analysis of the synthetic and real data distributions in the reduced space shown in Figure \ref{fdc ablation plots}.

Our findings reveal that for datasets containing a higher proportion of categorical features (e.g., the Migraine and Contraceptive Methods datasets), the synthetic data generated by NextConvGeN with the \textit{fdc} parameter resulted in a p-value greater than 0.05. This outcome supports the alternative hypothesis that the real and synthetic data are drawn from the same distribution. These results highlight the role of the \textit{fdc} parameter in ensuring the effectiveness of the NextConvGeN model, particularly for datasets with diverse feature types.

Further, we conducted an ablation study to evaluate the impact of the \var{alpha\_clip} parameter introduced in the NextConvGeN model to avoid generating real samples as synthetic samples. The study compared the percentage of exact matches between real and synthetic samples (referred to as the exact match score) across various generative models, including NextConvGeN, both with and without the \var{alpha\_clip} parameter. The results, summarized in Table \ref{exact match scores}, demonstrate that the NextConvGeN model without the \var{alpha\_clip} parameter (i.e., \var{alpha\_clip} = 0) produces a significant number of exact copies in synthetic data for most datasets. However, when the \var{alpha\_clip} parameter is enabled (\var{alpha\_clip} = 0.351), the exact match score is effectively reduced to near zero. This highlights the critical role of the \var{alpha\_clip} parameter in controlling the similarity between synthetic data and the corresponding real data.

\begin{table*}[htbp]\scriptsize\caption{The table presents the percentage of exact copies in the synthetic dataset, resembling the real data, averaged over five runs with different random seeds across various benchmarking datasets. For the NextConvGeN model, results are provided both with and without the \var{alpha\_clip} parameter. The findings indicate that the absence of the \var{alpha\_clip} parameter results in a higher percentage of exact copies, while its inclusion minimizes these copies.}\label{exact match scores}\centering\tabularnewline\vspace{5pt}\begin{tabular}{@{\hskip4pt}c@{\hskip4pt}|@{\hskip4pt}c@{\hskip4pt}|@{\hskip4pt}c@{\hskip4pt}|@{\hskip4pt}c@{\hskip4pt}|@{\hskip4pt}c@{\hskip4pt}|@{\hskip4pt}c@{\hskip4pt}}\hline
\textbf{Dataset} & \textbf{CTGAN} & \textbf{CTAB-GAN} & \textbf{TabDDPM} & \textbf{NextConvGeN}&\textbf{NextConvGeN}\\
& & & & \textbf{(without alpha\_clip)} & \textbf{(with alpha\_clip)}\\
\hline
\hline
Heart failure & 0.0000 & 0.0000 & 0.0000 &  6.1244 & 0.0000 \\
Heart disease & 0.0000 & 0.0000 & 0.0000 &  0.0000 & 0.0000 \\
Lung cancer & 0.0000 & 0.0185 & 0.4792 &  3.1296 & 0.6296 \\
Indian liver patients & 0.0000 & 0.0000 & 0.0000 & 4.2020 & 0.0000 \\
Liver cirrhosis & 0.0000 & 0.0000 & 0.0000 & 0.0000 & 0.0000 \\
Contraceptive methods & 0.7182 & 0.0281 & 14.8205 & 0.6259  & 0.3932 \\
Pima Indian diabetes & 0.0000 & 0.0000 & 0.0000 & 8.8492 & 0.0000 \\
Obesity & 0.0000 & 0.0000 & 0.0000 & 1.4575 & 0.0000 \\
Migraine & 0.4563 & 0.0000 & 0.7757 & 3.5437 & 0.4259 \\
Stroke & 0.0000 & 0.0000 & 0.0000 & 0.0000 & 0.0000 \\
\hline\end{tabular}\end{table*}

Additionally, the TabDDPM model exhibited the highest exact match score ($>14\%$) for the Contraceptive Method dataset. For three datasets: Migraine, Lung Cancer, and Contraceptive Method, several models reported non-zero exact match scores. This behavior is attributed to the structure of these datasets, which predominantly consist of categorical features with only one continuous feature. In such cases, generative models must approximate values within the categories, with a higher chance of contributing to higher exact match scores.

\end{document}